\documentclass{article}
\usepackage{arxiv}
\usepackage[english]{babel}

\usepackage[utf8]{inputenc} % allow utf-8 input
\usepackage[T1]{fontenc}    % use 8-bit T1 fonts
\usepackage[colorlinks=true, allcolors=blue]{hyperref}
\newcommand{\email}[1]{\href{mailto:#1}{\tt{\nolinkurl{#1}}}}
\newcommand{\orcid}[1]{ORCID: \href{https://orcid.org/#1}{\tt{\nolinkurl{#1}}}}
\usepackage{authblk}

\usepackage{amsmath}
\usepackage{graphicx}
\usepackage{subfig}
\usepackage{float}
\usepackage[colorlinks=true, allcolors=blue]{hyperref}
\usepackage{array}
\usepackage{tabularx} % 加载tabularx宏包
\usepackage{mlmath}
\usepackage{longtable} % 使用longtable宏包
\usepackage[table]{xcolor} % 使用xcolor宏包，带table选项
\usepackage{multirow} % 引入multirow宏包

\usepackage{amsmath,amsfonts,amssymb,mathtools,amsthm}

\newtheorem*{theorem*}{Theorem}

\newtheorem*{lemma*}{Lemma}

\newtheorem*{property*}{Property}

\newtheorem{definition}{Definition}

\newtheorem*{assumption*}{Assumption}
\newtheorem*{prop*}{Proposition}

\newtheorem*{setting*}{Setting}

\title{Anchor function: a type of benchmark functions for studying language models}
\author[1,+]{Zhongwang Zhang}
\author[1,+]{Zhiwei Wang}
\author[1]{Junjie Yao}
\author[1]{Zhangchen Zhou}
\author[1]{Xiaolong Li}
\author[2,3]{Weinan E}
\author[1,*]{Zhi-Qin John Xu}
\affil[1]{Institute of Natural Sciences, School of Mathematical Sciences, MOE-LSC and Qing Yuan Research Institute, Shanghai Jiao Tong University}
\affil[2]{AI for Science Institute}
\affil[3]{Center for Machine Learning Research, School of Mathematical Sciences, Peking University}
\affil[+]{These authors contributed equally to this work}
\affil[*]{Corresponding author: \email{xuzhiqin@sjtu.edu.cn}}

\begin{document}
\maketitle
\begin{abstract}
Understanding transformer-based language models is becoming increasingly crucial, particularly as they play pivotal roles in advancing towards artificial general intelligence. However, language model research faces significant challenges, especially for academic research groups with constrained resources. These challenges include complex data structures, unknown target functions, high computational costs and memory requirements, and a lack of interpretability in the inference process, etc.

Drawing a parallel to the use of simple models in scientific research, we propose the concept of an anchor function. This is a type of benchmark function designed for studying language models in learning tasks that follow an ``anchor-key'' pattern. By utilizing the concept of an anchor function, we can construct a series of functions to simulate various language tasks. The anchor function plays a role analogous to that of mice in diabetes research, particularly suitable for academic research.

We demonstrate the utility of the anchor function with an example, revealing two basic operations by attention structures in language models: shifting tokens and broadcasting one token from one position to many positions. These operations are also commonly observed in large language models. The anchor function framework, therefore, opens up a series of valuable and accessible research questions for further exploration, especially for theoretical study.
% It is increasingly important to understand language models as they are playing critical roles in the way towards artificial general intelligence. The language model research is encountering significant challenges, particularly for academic research groups operating with constrained resources, such as complex data structure, Unknown target functions, large computational cost and memory, short of interpretability of the inference process etc. Similar to the constructing simple models in scientific research, such as the mouse model for studying diabetes, we propose anchor function, which is a type of benchmark functions for studying language models in learning tasks of``anchor-key'' pattern type. We use the concept of anchor function to construct a series of functions to simulate various language tasks. The anchor function serves as a role to language model as mice to studying diabetes, especially fitting for academic research with limited source. We use an anchor function example to reveal two basic operations by attention structures, i.e., shifting tokens and broadcasting one token at one position to many positions, which are also found to be common in large language models. The anchor function renders a series of high-value and accessible questions for further exploring.
\end{abstract}

\section{Introduction}
% Language models (LMs), large language models (LLMs) in particular, have shown a potential approach towards artificial general intelligence (AGI) \cite{sun2023survey}, such as ChatGPT \cite{openai2023gpt4} developed by OpenAI, Llama \cite{touvron2023llama} by meta, Claude by Anthropic. Trained by predicting next token in a very large dataset, such LLMs exhibit many surprising abilities and new interesting phenomena, like emergent abilities, in-context learning etc. In addition to good generalization of tasks seen during the training, LLMs can generalize to new tasks without training but with an appropriate prompt \cite{luo2023prompt}. Research in LMs can yield to various benefits, such as making LLMs better, more efficient and safer, understanding how language is processed by neural networks, providing insights into neuroscience etc. 

Language models (LMs), large-scale language models (LLMs) in particular, including prominent examples like ChatGPT by OpenAI \cite{openai2023gpt4}, Llama by Meta \cite{touvron2023llama}, Claude by Anthropic, and Gemini by Google DeepMind etc., are increasingly being recognized for their potential role in advancing artificial general intelligence (AGI) \cite{sun2023survey}. These transformer-based models \cite{vaswani2017attention}, trained through next-token prediction on extensive datasets, lead to the emergence of remarkable capabilities and novel phenomena such as in-context learning and debated emergent abilities \cite{schaeffer2023emergent}. Notably, LLMs demonstrate effective generalization on tasks encountered during training and an impressive ability to adapt to new tasks via apt prompting \cite{luo2023prompt}, without the need for additional training. The ongoing research in the field of language models is crucial for multiple reasons: enhancing the performance, efficiency, and safety of LLMs; gaining deeper insights into how neural networks process language; and contributing to our understanding of neuroscience and related domains.

% Research in LMs is facing substantive challenges, especially for research groups in academics with very limited resources. Some challenges are listed as follows.
Whether LLMs can understand language in the human sense is a heated debate \cite{mitchell2023debate}. Alternatively, in this work, we focus on studying how a transformer network can accomplish language tasks rather than discussing the ``understanding'' in LLMs. 
The language model research is encountering significant challenges, particularly for academic research groups operating with constrained resources. These challenges include, but are not limited to, the following:
\begin{itemize}
\item \textbf{Complex data structure:} Words can have multiple meanings depending on the context and one word can be broken down into several tokens, making it difficult to analyze the specific function of each token. For instance, the prompt ``calculate the sum of the first two numbers: 10, 20, 30'' involves multiple tokens to convey a mathematical operation, and the way these tokens interact to represent this intent can vary widely, such as ``add the first number and the second number: 10, 20, 30''.

\item \textbf{Training and testing boundarfy:} The intricate nature of language makes it hard to decisively determine the similarity between two sentences. With the vast size of language model training data, it's nearly impossible to completely separate training and test datasets or to ascertain if a particular example was part of the training data. This issue is evident in studies like the reversal curse of LLMs \cite{berglund2023reversal}, where the frequency of certain phrases in the training data is hard to verify. For example, one might argue that ``Who is Tom Cruise's mother?'' is more often seen than ``Who is Mary Lee Pfeiffer's son?'' in the training data set and it is hard to check this argument. 

\item \textbf{Unknown target functions:} Without a predefined target function, analyzing the training process, such as which aspects of the target function are learned first, becomes challenging.

\item \textbf{Large computational cost and memory:} Developing a well-trained language model typically requires extensive data and significant computational and memory demands, which are often beyond the reach of academic institutions.

\item \textbf{Data preparation for specific tasks:} Crafting a dataset for a particular task, like studying how an LLM learns sentiment classification, is complex and requires controlling numerous variables.

\item \textbf{Interpretability of the inference process:} Understanding how language models perform various tasks and interpreting their decision-making process are significant challenges.
\end{itemize}

Studying a natural phenomenon in scientific research often parallels the complexities encountered in language model research.
% Take, for instance, the simple act of an object falling freely is governed by a multitude of variables such as the object's mass, size, surface smoothness, gravitational force, friction, and temperature, among others. 
As Richard Feynman famously argued, ``What I cannot create, I do not understand'', and Geoffrey E. Hinton's paper, ``To recognize shapes, first learn to generate images'' \cite{hinton2007recognize}, both perspectives underline a fundamental principle in scientific inquiry: to truly comprehend a complex phenomenon, one should possess the capability to reproduce it. For example, mice are a valuable model for diabetes research due to their genetic and physiological similarities to humans, allowing for relevant gene function studies. Their short lifespan enables quicker observation of the disease process, enhancing research efficiency. Additionally, mice are cost-effective and easy to breed, reducing research costs. Importantly, their genes can be easily modified using genetic engineering, facilitating the study of specific genes in diabetes development. These advantages make mice an ideal choice for diabetes research.

% In addressing complex problems, simplification and focus on key factors are often employed. For instance, in studying free fall, one might primarily consider gravity, setting aside other variables. This methodology is also evident in medical research. Given the complexity of the human body and the ethical and financial considerations of human experimentation, diseases are often studied through more manageable animal models. 
This scientific approach to animal models is applicable to the study of language models. By designing careful and focused experiments, researchers can uncover significant characteristics of language models. For example, studies like Lake et al.'s \cite{lake2023human} demonstrate the potential of language models to learn systematic compositionality. \cite{garg2022can,akyurek2022learning,ahn2023Transformers,von2023Transformers} analyze the mechanism of transformers on various designed linear regression tasks. \cite{olsson2022context,chan2022data,reddy2023mechanistic} utilize the item-label sequence and \cite{lake2023human} designs a primitive composition dataset. \cite{rende2023does} generates the sequence from a Boltzmann distribution to train the transformer model.  Such targeted experimentation can effectively isolate and analyze specific capabilities of language models, contributing to a deeper understanding of their functionalities and limitations.

In language-related tasks, a common pattern is the identification of anchor words that indicate specific operations to be performed on other tokens within the sentence. For instance, in phrases like ``summation of 1, 16, 17'' and ``product of 1, 16, 17'', the anchor words ``summation'' and ``product'' respectively signal the mathematical operations to be applied to the key tokens 1, 16, and 17. 
% These key words serve as crucial indicators for the intended action, guiding the language model in processing the sentence and executing the appropriate calculation. Such patterns are essential in enabling language models to accurately interpret and respond to various prompts and instructions.
To better comprehend how neural networks manage the ``anchor-operation'' pattern, this paper introduces a concept referred to as the ``anchor function'' for studying language models. 
% An anchor function is essentially a tool designed to analyze how LMs identify and respond to specific cues or prompts within a sequence.
A basic example of an anchor function is the identity learning function. In this scenario, a designated number, which acts as an anchor prompt, appears only once in each input sequence. The output of this function is the number immediately following this specific anchor. For example, if we take ``3'' as the anchor, then for the sequence $(43,33,13,3,20,89,44,24,56)$, the output would be $20$. All other numbers in the sequence are inconsequential to the result.
This concept mirrors a task that finds out the name in the input sequence, for example, the output of ``He is from China, his name is Mike, he likes reading.'' is ``Mike'' and the anchor is ``name is''.

Furthermore, one could design a task involving two different anchors, each directing the network to execute a distinct operation. This kind of task is reflective of composite tasks in language models, where multiple prompts or cues within a sequence guide the model to perform various operations, demonstrating the model's ability to handle complex instructions. The anchor function qualifies as an effective type of benchmark function for studying language models, offering, but are not limited to, the following several advantages:

\begin{itemize}
\item The anchor function can simulate a range of language tasks, demonstrating its versatility in reproducing various linguistic scenarios.
\item It is cost-effective in terms of computation and memory, making it a feasible option for academic research groups with limited resources.
\item With a well-defined target function, the anchor function provides a clear and straightforward objective for analysis.
\item It allows for a distinct separation between training data and test data, essential for accurate model evaluation.
\item Experiments with the anchor function can be conducted under strict control of variables, ensuring precise and reliable outcomes.
\item This function is instrumental in studying both the generalization within a single task and the transfer of learning from known tasks to new, unseen tasks.
\item Fully-connected networks hardly learn anchor functions, in contrast to language models, which sheds light on the differing capabilities of various network architectures.
\end{itemize}

These attributes make the anchor function a robust and insightful tool for probing the complexities and capabilities of language models.

In this work, we will showcase a range of anchor functions that effectively simulate various language tasks. These functions serve as tools to investigate the operational mechanisms of transformer networks in executing these tasks. Specifically, in the context of the identity learning anchor function, we have identified that two fundamental operations, executed by the attention structure, are critical. These operations involve shifting and broadcasting tokens. Our findings also reveal that these attention processes are also present in LLMs, such as Llama. We further illustrate that these operations can be achieved by manipulating the orientation of weight vectors within a high-dimensional space. We also utilize frequency principle \cite{xu2019training,xu2019frequency,rahaman2018spectral,xu2022overview} and condensation \cite{luo2019theory,zhou2021towards} of existing deep learning theory to understand the training of transformer networks. Building on the anchor functions, we outline a series of intriguing and significant issues that can be further explored in transformer networks using the anchor function framework. This approach opens up new avenues for understanding and enhancing the capabilities of transformer-based models, especially for theoretical study.

\section{Related works}

\cite{voita2019analyzing} evaluate the contribution of individual attention heads in the encoder to the overall performance of the model and propose a method to prune the attention heads. \cite{clark2019does} analyze the attention mechanism of the Bert model and find that its attention head shows some pattern like specific positional offsets or focuses on delimiter tokens. \cite{kovaleva2019revealing} analyze Bert's encoding capabilities and attention matrix, propose the idea that Bert is over-parameterized, and improve Bert's performance by disabling certain attention layers. 

\cite{kaplan2020scaling} study the empirical scaling law of language model performance on cross-entropy loss.  \cite{ma2022self} prove that self-attention is a natural structure for dealing with sequence-to-sequence problems. \cite{jiang2023approximation} examine the ability of the transformer model to approximate sequential relationships, establish a universal approximation theory in the hypothesis space of the transformer, and provide an explicit approximation rate estimate. 

\cite{wang2023label} investigate the working mechanism of in-context learning (ICL) through an information flow lens. They demonstrate a mechanism by which ICL aggregates label descriptions into label words at shallow layers and uses label words as the reference for the final output at deeper layers. 

\cite{santoro2016meta} introduce a memory-augmented neural network to solve a meta-learning task. In such a task, each data episode has the same meta structure, but a different detail rule. \cite{lake2023human} introduce a similar meta-learning for compositionality (MLC)  to verify that the transformer network can achieve the systematic generalization of compositionality. In MLC, each sequence in the training or test data contains a different grammar generated by the same meta-grammar. As the training converges, the standard encoder-decoder transformer can predict the output of test data entirely dependent on the grammar in the input sequence in the given test data.
% This method allows machines to achieve human-level systematic generalization of compositionality.

% \cite{olsson2022context} shows that to predict $y$ in a sequence $\ldots,x,y,\ldots,x$ for $x,y$ not seen during the training, a attention-based neural network will compute through an induction head. 
% \cite{olsson2022context} shows that the attention structure of a transformer neural network in learning the task ``$\ldots,[A],[B],\ldots,[A]\rightarrow [B]$'' often exhibits a shift operation, which is termed ``induction head''. This task is exactly the identity learning anchor function defined in this paper and the shift operation by attention is also validated in our experiments.

\cite{elhage2021transformer} notice that in a two-layer transformer neural network, some heads in the second layer attention matrix can notice the similarity between the current token and the previous tokens in the input sentence, then, the network would output the next token of the previous similar token, thereby completing tasks like ``$\ldots,[A],[B],\ldots,[A]\rightarrow [B]$''. Such a head that pays attention to previous similar tokens is called an ``induction head''. \cite{olsson2022context} point out that induction heads may be the primary mechanism for the transformer model to achieve in-context learning.

% a transformer-based neural network will compute through a structure named ``the induction head''. And the minimal structure is a two-layer attention-only model, which is similar to the anchor function of identity task in this paper. 

\cite{chan2022data} utilize the Omniglot dataset \cite{lake2019omniglot} to show that the data distribution can affect the trade-off between in-weight learning and in-context learning.  
% including burstiness and within-class variation in the ``item-label'' experiments.

\cite{reddy2023mechanistic} reproduce the phenomenon that the emergence of the induction attention heads associates with the emergence of the in-context learning \cite{olsson2022context} through designed experiments with input sequence consisting of several ``item-label'' pairs \cite{chan2022data}. \cite{reddy2023mechanistic} construct a simple model to show the induction head can lead to in-context learning. Note that this ``item-label'' sequence task is similar to the anchor function of the classification task in this paper.

% @article{vinyals2016matching,
%   title={Matching networks for one shot learning},
%   author={Vinyals, Oriol and Blundell, Charles and Lillicrap, Timothy and Wierstra, Daan and others},
%   journal={Advances in neural information processing systems},
%   volume={29},
%   year={2016}
% }

% @inproceedings{santoro2016meta,
%   title={Meta-learning with memory-augmented neural networks},
%   author={Santoro, Adam and Bartunov, Sergey and Botvinick, Matthew and Wierstra, Daan and Lillicrap, Timothy},
%   booktitle={International conference on machine learning},
%   pages={1842--1850},
%   year={2016},
%   organization={PMLR}
% }

% @article{chan2022data,
%   title={Data distributional properties drive emergent in-context learning in transformers},
%   author={Chan, Stephanie and Santoro, Adam and Lampinen, Andrew and Wang, Jane and Singh, Aaditya and Richemond, Pierre and McClelland, James and Hill, Felix},
%   journal={Advances in Neural Information Processing Systems},
%   volume={35},
%   pages={18878--18891},
%   year={2022}
% }

% @inproceedings{lake2018generalization,
%   title={Generalization without systematicity: On the compositional skills of sequence-to-sequence recurrent networks},
%   author={Lake, Brenden and Baroni, Marco},
%   booktitle={International conference on machine learning},
%   pages={2873--2882},
%   year={2018},
%   organization={PMLR}
% }

\section{Definitions}

In this section, we define the anchor function. We also propose a method for partitioning the dataset and define generalization on data and tasks.
    
\subsection{Anchor function}
We will give some examples of anchor functions, followed by a general definition.

% \textcolor{red}{Suppose we have a sentence $(\vx_1,\ldots, \vx_l) \in \sR^{l\times d}$. An anchor set $A$ consists of several special tokens, serving as various types of prompts with  designated functions, for example, for $\bar{A}\subset A$ that exists in the input sequence,  $f_{\bar{A}}$, termed as an anchor function, maps specific phrases $G=\{\vx_{j}|j\in I_{\bar{A}}\}$ in this sentence to the target output $f_{\bar{A}}(\vx_1,\ldots,\vx_l)$. The following are some specific examples of anchor functions.}

\subsubsection{One-anchor function}
Consider one anchor function for identity learning $f(X):\sR^{n\times d}\rightarrow \sR^{d}$, where $n$ is the number of tokens, and $d$ is the dimension of each token, $X=(\vx_1,\vx_2,\ldots,\vx_n)$ and $\vx_i \in \sR^{d}$. In each $X$, one and only one of $\vx_1,\vx_2,\ldots,\vx_{n-1}$ has to equal to a designated prompt anchor $\va\in\sR^{d}$, and 
\begin{equation}
    f(\vx_1,\ldots,\vx_n)=\vx_{i+1}, {\rm \quad where \quad} \vx_i=\va. \label{eq: identitt}
\end{equation}
For example, when $d=1$ and $a=3$, $f(12,33,14,3,42,22,32,20,28)=42$.
To use neural network to learn this function, each token is represented by a one-hot vector, i.e., $\vx\in \{0,1\}^{d}$, where $d$ is the vocabulary size. 

The prompt anchor indicates a task or an operation for its next token. We can generalize the identity function to an arbitrary function, i.e., 
\begin{equation}
    f(\vx_1,\ldots,\vx_n)=g(\vx_{i+1}), {\rm \quad where \quad} \vx_i=\va,
\end{equation}
where $\vx_i$ is called an anchor item and $\vx_{i+1}$ is called a key item. 
This function can be further reformulated as 
\begin{equation}
    f(\vx_1,\ldots,\vx_n)=\sum_{j=1}^{n-1}\frac{g(\vx_{j+1})}{\norm{\vx_{j}-\va}}/\sum_{j=1}^{n-1}\frac{1}{\norm{\vx_{j}-\va}}.
\end{equation} 
% The prompt anchor can be placed before the token that is to be computed, that is, 
% \begin{equation}
%     f(\vx_1,\ldots,\vx_l)=\sum_{j=2}^{l}\frac{g(\vx_{j-1})}{\norm{\vx_{j}-\va}}/\sum_{j=1}^{l-1}\frac{1}{\norm{\vx_{j}-\va}}.
% \end{equation} 
In general, a one-anchor function satisfies the following form
\begin{equation}
    f(\vx_1,\ldots,\vx_n)=\sum_{j=2}^{n}\frac{g_j(\{\vx_{k}|k\in I_j\})}{\norm{\vx_{j}-\va}}/\sum_{j=1}^{n-1}\frac{1}{\norm{\vx_{j}-\va}},
\end{equation} 
where $g_j$ is a function depending on index $j$, and $I_j$ is a set of indexes depending of $j$.

\subsubsection{Two-anchor composite function}
Consider one two-anchor composite function $f(X):\sR^{n\times d}\rightarrow \sR^{d}$. The anchor set consists of a list of different tokens, i.e., $A=\{\va_1,\va_2,\cdots,\va_J\}$, and each token $a_k\in A$ corresponds to function $g(\vx;a_k)$. In each input sequence $X$, one and only one pair of two consecutive elements belong to $A$, such as  $\vx_i,\vx_{i+1}\in A$. Then,
\begin{equation}
    f(\vx_1,\ldots,\vx_n)=g(g(\vx_{i-1};\vx_{i});\vx_{i+1}), {\rm \quad where \quad} \vx_i,\vx_{i+1}\in A.
\end{equation}
%ADD: MORE GENERAL FORM
%\subsection{Multi-composition anchor function}
%Consider one multi-composition anchor function $f(X):\sR^{l\times d}\rightarrow \sR^{d}$. The anchor set consists of $p$ different tokens, i.e., $\Omega=\{\va_1,\va_2,\ldots,\va_p\}$, and token $a_k\in\Omega$ corresponds to function $g(\vx;a_k)$. In each $X$, one and only one group of $m$ consecutive elements belong to $\Omega$, such as  $\vx_i,\vx_{i+1},\ldots,\vx_{i+m-1}\in\Omega$. Then,
%\begin{equation}
%    f(\vx_1,\ldots,\vx_l)=g(\ldots g(g(\vx_{i+2};\vx_{i});\vx_{i+1})\ldots;\vx_{i+m-1}), {\rm \quad where \quad} \vx_i,\vx_{i+1},\ldots,\vx_{i+m-1}\in\Omega.
%\end{equation}
%ADD: MORE GENERAL FORM

\subsubsection{Forward-backward recitation anchor function}
    Consider one forward-backward recitation anchor function $f(X):\sR^{n\times d}\rightarrow \sR^{d}$,

    $$ f(\vx_1,\ldots,\vx_n)=\left\{
        \begin{aligned}
        \vx_{i+1} \quad & \mathrm{if\ } \vx_i=\va \mathrm{\ and\ } \vx_n=\vx_{i-1} {\ \mathrm{(forward\ task)}},\\
        \vx_{i-1} \quad & \mathrm{if\ } \vx_i=\va \mathrm{\ and\ } \vx_n=\vx_{i+1} {\ \mathrm{(backward\ task)}}.
        \end{aligned}
        \right.
    $$
This task also has an analytical formulation, i.e.,
    \begin{equation}
        f(\vx_1,\ldots,\vx_n)=-\vx_n + \sum_{j=2}^{n-1}\frac{\vx_{j+1}+\vx_{j-1}}{\norm{\vx_{j}-\va}}/\sum_{j=2}^{n-1}\frac{1}{\norm{\vx_{j}-\va}}.
    \end{equation} 
This anchor function simulates the recitation of a previous sentence or a following sentence with the hint of the current sentence.
% The actual scenario corresponding to this task is that the corpus of the LM already contains long sentences composed of two or more single sentences, for example, a poem. At this time, the LM needs to provide the previous sentence content or the next sentence content of the sentence fragment given by the user.

\subsubsection{General anchor function}
A general anchor function with an anchor set $A$ can be formulated in the following form:
\begin{equation}
	f(\vx_1,\ldots,\vx_n) = g(\vx_1,\ldots,\vx_n;\{\vx_1,\ldots,\vx_n\} \cap A),
\end{equation}
where $g$ is a function of a subset of $\{\vx_1,\ldots,\vx_n\}$, where $g$ depends on the anchor terms $\{\vx_1,\ldots,\vx_n\}\cap A$, and the subset items are called key items.
	
\subsection{Training and test data}
% Due to the complexity of language, it is often difficult to separate training and test data well, which is neither trivial for anchor functions. For example, consider the identity learning anchor function. If the number after the anchor is taken from $[10,50]$ in the training dataset while $[51,60]$ in the test dataset, the neural network can never learn the test dataset. The reason is that the neural network never needs to differentiate those tokens in $[51,60]$. 

Due to the complexity of natural language, cleanly separating training and test data is often challenging. The idealized anchor function appears to be capable of accomplishing the aforementioned task. However, a straightforward division based on data ranges proves to be impractical. To illustrate, consider a scenario where the range of key items in the training set is denoted as $[a, b]$,  while in the test dataset, it is represented as $[b+1, c]$. The encoding of data within the interval $[b+1, c]$ is not learned during the neural network training process. As a result, the neural network fails to produce the key item output for the test dataset.

\begin{figure}[ht]
    \centering
    \includegraphics[width=0.8\linewidth]{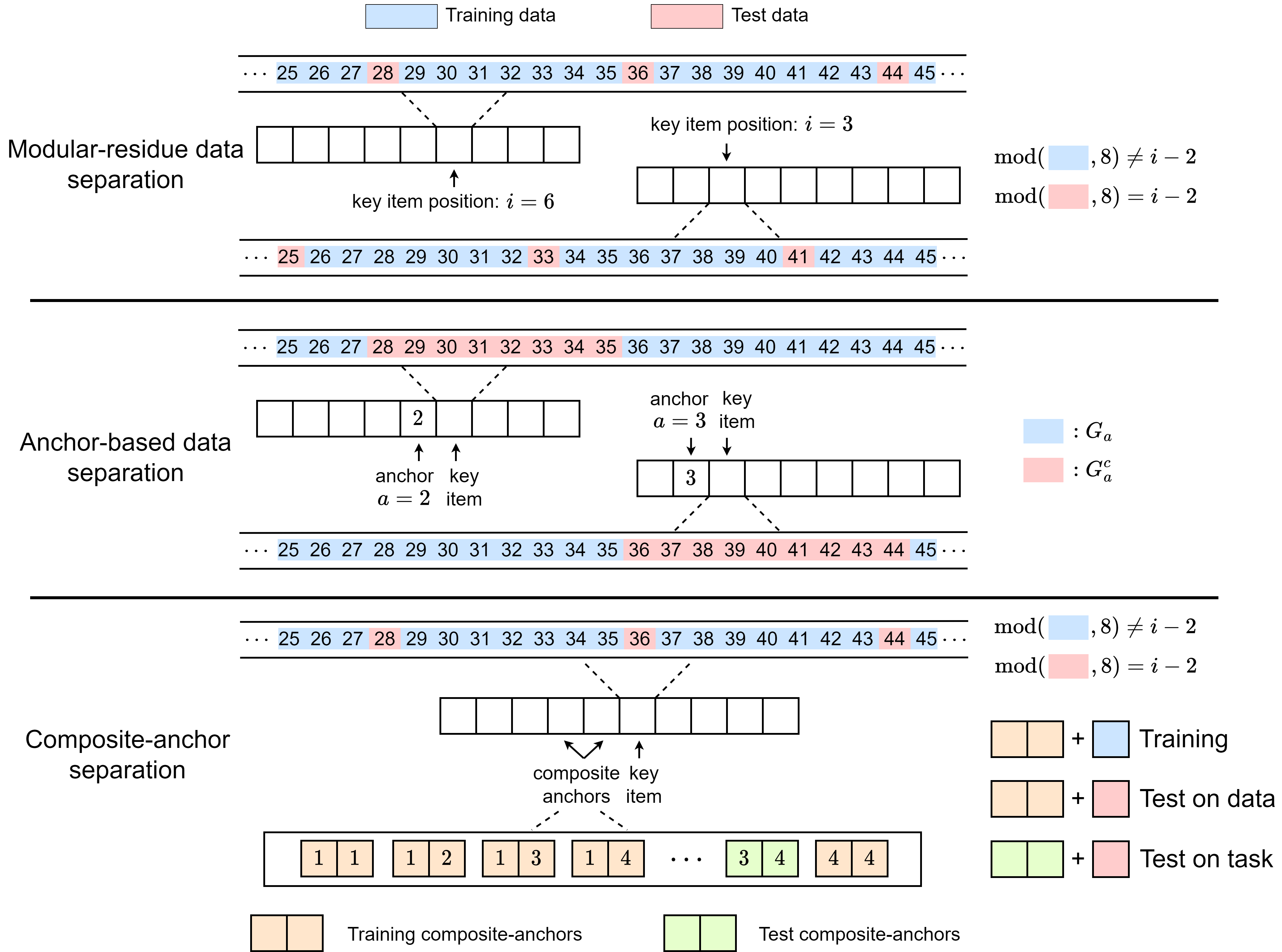}
    \caption{Schematic diagram of three ways to split the training set and the test set.}
    \label{fig:dataset}
\end{figure}

In this section, we propose three methods to separate training and test data, as shown in Fig. \ref{fig:dataset}.

\subsubsection{Modular-residue data separation}
Consider a task with an input sequence of length $n$. Define $\Gamma_i$ as a set depending on the position~$i$. For example, $\Gamma_i = \{i-2\}$ for the case where key items can not be placed in the first position. Modular-residue data separation is as follows.

For an input sequence of the training dataset, if the token of the $i$-th position is a key item, then, a number $x$ can be placed in the $i$-th position of such input sequence only when ${\rm mod}(x,n-1)\notin \Gamma_i$.

For an input sequence of the test dataset, if the token of the $i$-th position is a key item, then, a number $x$ can be placed in the $i$-th position of such input sequence only when ${\rm mod}(x,n-1)\in \Gamma_i$.

% We propose the following method to differentiate training and test data. Suppose the computed token is in the $i$th position. If token $x$ satisfies ${\rm mod}(x,l)\in \Gamma_i$, then, the pattern $x$ only exists in the sequence of test data, where ${\rm mod}(x,l)$ represents the operation of taking the modulus of $x$ with respect to $l$, $\Gamma_i$ is a given set depending on $i$, such as $\Gamma_i={i}$. Such separation makes sure each token is shown in the output during the training. The test data and the training data do not overlap.

\subsubsection{Anchor-based data separation}
Consider a problem with multiple anchors. Denote the anchor set as $A=\{a_i\}_{j}^{J}$. Take an example that in each input sequence, one and only one token belongs to the anchor set. Anchor-based data separation is as follows.

In the training data set, key items of anchor $a_i$ belong to a set denoted as $G_{a_i}$ and its output token belongs to a set denoted as $G^{o}_{a_i}$. Denote the set of all tokens as $I$ and the set of all possible output tokens as $I^{o}$. These sets satisfy the following requirements:  $G_{a_i}\subsetneq I$ and $G^{o}_{a_i}\subsetneq I^{o}$,  $I=\cup G_{a_i}$ and $I^{o}=\cup G^{o}_{a_i}$. The key items for $a_i$ in a test sequence should belong to $G_{a_i}^c=I\backslash G_{a_i}$.

\subsubsection{Composite-anchor separation}
Consider an anchor set with multiple anchors. In each input sequence, there exists a composition of certain anchors. The compositions of anchors in the training set and in the test set do not overlap.

\subsection{Generalization}

% \begin{figure}
%     \centering
%     \includegraphics[width=1\linewidth]{pic/two_kind_of_generalization.png}
%     \caption{Task generalization and data generalization.}
%     \label{fig:generalization}
% \end{figure}

% In an anchor function, each anchor is represents a task. In each task, there are many data points. Therefore, we can classify generalization into two types, i.e., generalization over data and generalization over tasks.
% % \subsection{Generalization over data}

% For an anchor function, the input sequence contains $K$ consecutive tokens as the anchor. Each token is taken from the given anchor set and indicates an operation. 
% \begin{definition} [generalization error over data] For a dataset, in which the combination of the anchor tokens in each sequence has been used in the training data, then, the error of this dataset is generalization error over data.
% \end{definition}

% \begin{definition} [generalization error over tasks] For a dataset, in which the combination of the anchor tokens in each sequence has never been used in the training data, then, the error of this dataset is generalization error over tasks.
% \end{definition}

The division of the dataset naturally leads to two concepts of generalization: generalization on data and generalization on tasks. The first type describes the model's ability to generalize to seen tasks and unseen data, while the second type pertains to its ability to generalize to unseen tasks.

\textbf{Generalization on data.} Generalization on data primarily relies on the test set, which is divided according to key items. In this test data, the anchor function (i.e., the task) has already been encountered in the training set. This form of generalization is based on data separation, such as the case generated through modular-residue and anchor-based data separation. 
% Additionally, the data composition of the ``test on data'' in composite-anchor separation is also designed to explore the generalization on data.

\textbf{Generalization on task.} Generalization on task primarily depends on the test set, divided based on the anchors. Although some anchor combinations in the test data are new (not present in the training set), the model can infer unseen composite anchor functions using the existing composite anchors from the training set. 
% This type of generalization is primarily grounded in the data composition of the ``test on task'' in composite-anchor separation.

\subsection{Loss function}
    In this work, we use a transformer architecture, which takes an input sequence of length \(n=9\), subsequently yielding an output matrix \(X^{(\mathrm{out})} \in \mathbb{R}^{n \times d}\), where \(d\) is the size of the dictionary. For the $i$-th input sequence, the loss function is the cross-entropy only for the output of the last token after softmax (denoted as \(\mathbf{x}^{(\mathrm{out}, i)}_{n}\in\sR^{d}\)). The total loss over the entire training dataset is computed as:
    % \[
    % l_i = -\sum_{c=1}^{d} y_{i,c} \log(\mathbf{x}^{(out)}_{i,l,c}),
    % \]

    $$\mathcal{L} = -\frac{1}{N} \sum_{i=1}^{N}\sum_{c=1}^{d} \vy^{(i)}_c \log(\mathbf{x}^{(\mathrm{out}, i)}_{n,c}),$$
    where \(\vy^{(i)}\) is a one-hot label for the $i$-th sequence and \(N\) represents the size of the training dataset.

\section{Tasks and results}

\begin{figure}[h]
        \centering
        \includegraphics[width=\linewidth]{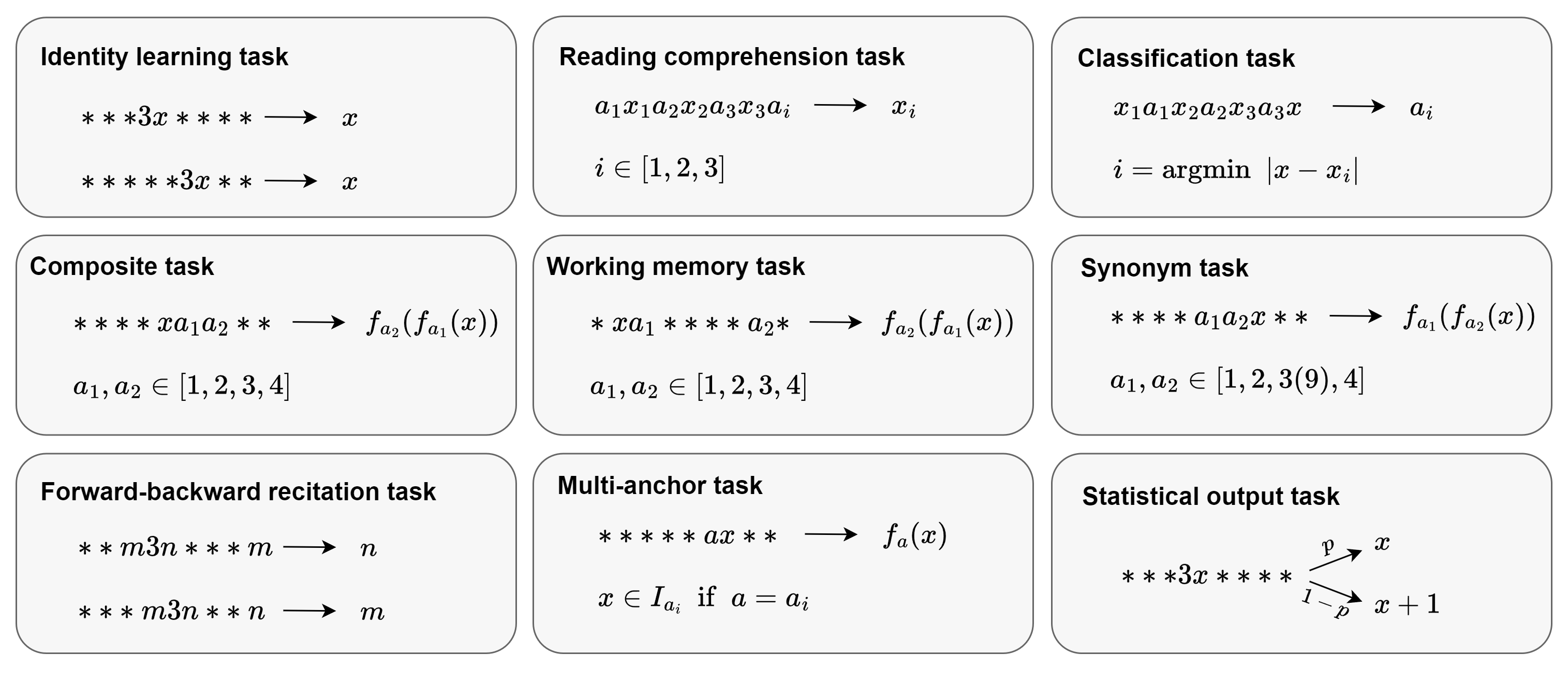}
        \caption{Task examples of anchor functions.}
        \label{fig: tasks}
    \end{figure}

\begin{figure}
    \centering
    \subfloat[loss]{\includegraphics[width=\linewidth]{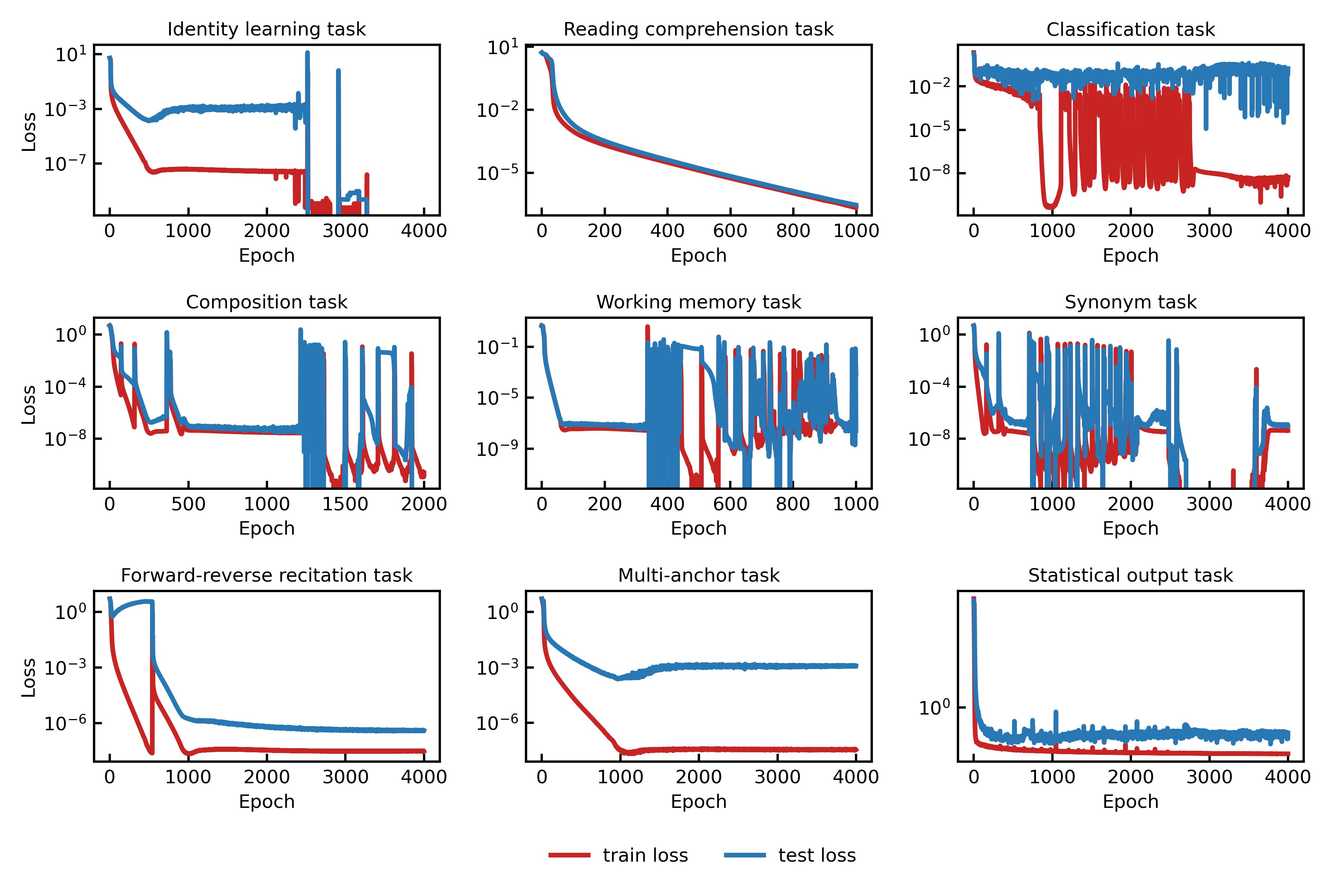}}
    \newline
    \subfloat[accuracy]{\includegraphics[width=\linewidth]{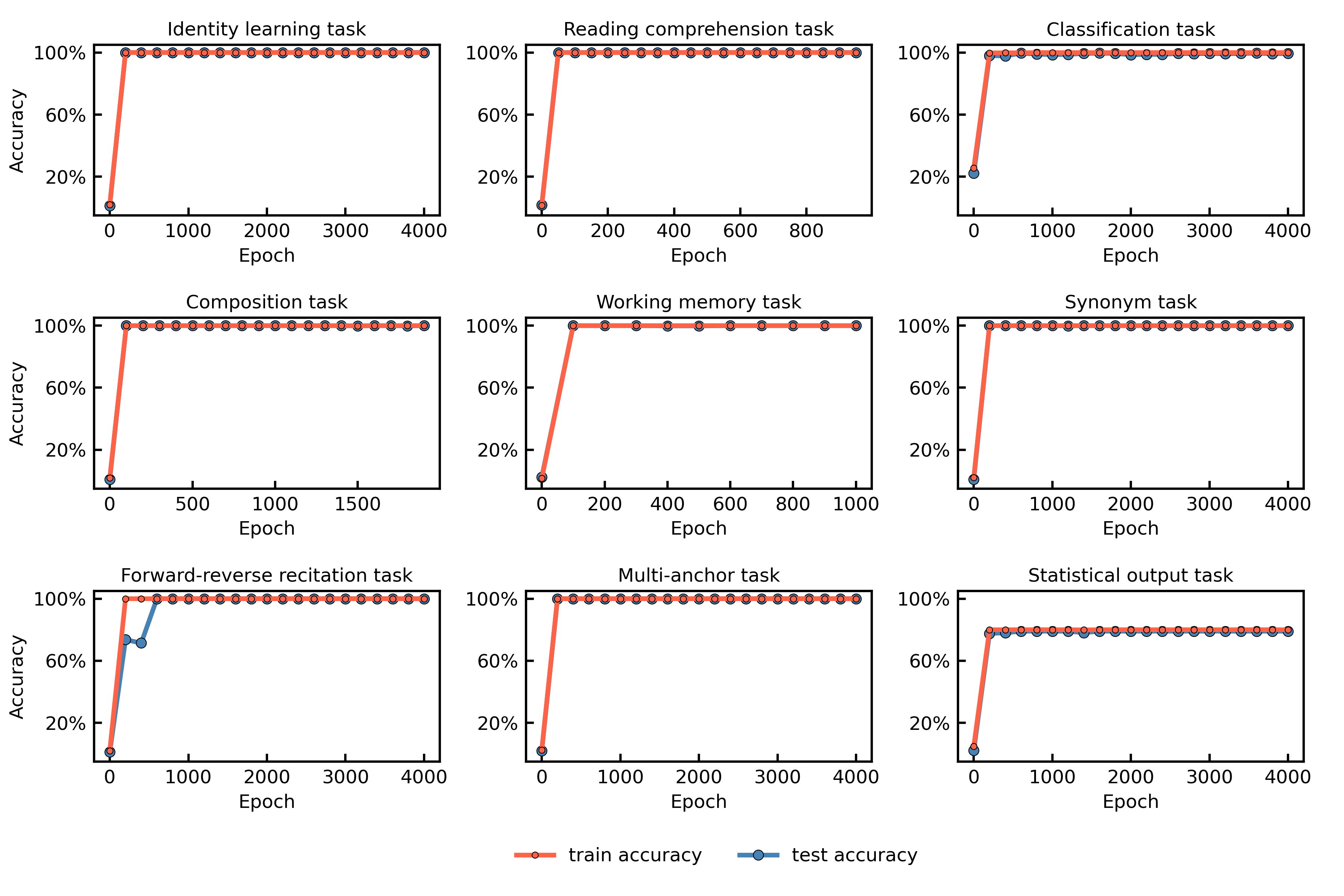}}
    \caption{Loss and accuracy of each task.}
    \label{fig:loss and acc}
\end{figure}

In this section, we apply transformer models in various tasks to simulate real-world scenarios and delve into certain mechanisms underlying them. Without specification, we employ a 4-layer decoder-only transformer network to learn tasks. Each decoder layer comprises four heads. Please refer to the appendix for detailed settings. Nine tasks are presented in Fig. \ref{fig: tasks} and the loss and accuracy for each task during the training are summarized in Fig. \ref{fig:loss and acc}. A multi-layer transformer neural network can learn and generalize all listed examples.

    \subsection{Identity learning task}

    We begin by examining the identity learning function, which simulates a task to find out the name in the input sequence. For example, the output of ``He is from China, his name is Mike, he likes reading.'' is ``Mike'' and the anchor is ``name is''.
    % A practical scenario for this task can be as follows: ``I am from China, my name is John, and I like to read books. I am called John''. Here, the anchor item is ``name'', and the expected output is the word ``John''; other parts of the sentence do not affect the output. 
    
    The identity learning function is defined in Eq. (\ref{eq: identitt}), where the anchor item serving as a prompt occurs only once in the input sequence. The function's output is the digit following that specific anchor. In our experiments, we use ``3'' as the anchor term. Meanwhile, we study the mechanism by which the transformer model has good generalization ability on the identity learning task in Section \ref{sec:mechanism}.

    % \subsubsection{Comparison of Different Models}

\begin{figure}[ht]
        \centering
        \includegraphics[width=0.45\linewidth]{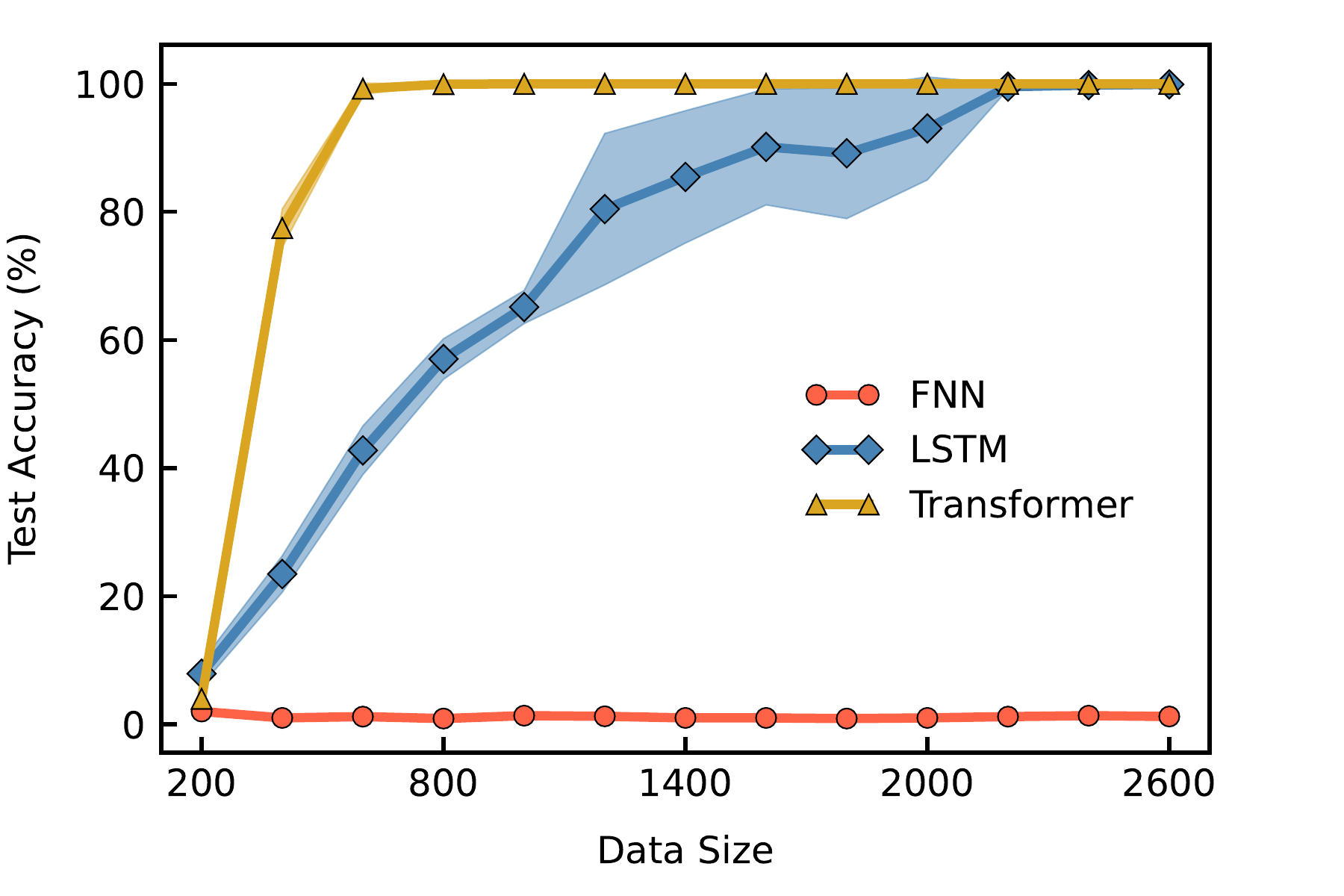}
        \caption{Performance comparison of FNN, LSTM, and transformer. For each experimental setting, we used 10 random seeds for the experiment. Each experiment was trained for 4000 epochs. The scatter points are the average of the ten experimental results, and the shaded parts are the maximum and minimum values.}
        \label{fig:sim_task_com}
    \end{figure}

    We compared the generalization capabilities of fully-connected neural networks (FNNs), LSTMs, and transformers in the identity learning task. The respective parameter counts for these models are 416M, 5.3M, and 5.6M. Please refer to the appendix for detailed setups for the three models. As shown in Fig.~\ref{fig:sim_task_com}, the transformer demonstrates excellent generalization performance with just 600 training data points, while LSTM requires around 2200 training data points to achieve a similar level of generalization. DNN, on the other hand, fails at generalization.

    % \begin{figure}
    %     \centering
    %     % \subfloat[]{\includegraphics[width=0.3\linewidth]{pic/reading_comprehension.png}}
    %     % \subfloat[]{\includegraphics[width=0.3\linewidth]{pic/classification.png}}
    %     \includegraphics[width=0.5\linewidth]{pic/classification_emb.png}
    %     \caption{The t-SNE distribution of vectors after passing through the embedding layer for numbers from 20 to 100.}
    %     \label{fig:classification}
    % \end{figure}

    \subsection{Reading comprehension task}
    
    In daily reading comprehension, one usually identifies the anchors in the question first and then searches the similar anchors in the given text. The answer can often be found in the content near the anchor. We set the following task to simulate this reading comprehension task. 
    % people usually consult the original text based on the anchor points in the question, search for words in the original text that are the same or very similar to the anchor points, and conduct answer near that word. In this particular task, 
    We define distinct anchor values $a_i \in \{1,2,\cdots,8\}$, and item values $x_i \in [20,100]$. The input sequence follows the pattern $(a_1, x_1, a_2, x_2, a_3, x_3, a_4, x_4, a)$, where $a_i$ is an anchor, $a_i\neq a_j$ for $i \neq j$. The output label is $x_k$ where $k$ satisfies $a_k = a$. In this task, the division of the training set and test set is done through the ``Modular-residue data separation'' method. We use a total of 1000 data points for training, with a batch size of 50, and train for 1000 epochs.
    
    \subsection{Classification task}
    
    The classification task is a variation of the reading comprehension task, focusing on content that is similar rather than identical to the anchors. In this task, we define distinct anchor values $a_i \in \{1,2,\cdots,8\}$, and item values $x_i \in [20,100]$. The input sequence follows the pattern $(x_1, a_1, x_2, a_2, x_3, a_3, x_4, a_4, x_5)$, where $a_i$ is an anchor, $a_i\neq a_j$ for $i \neq j$. The output label is $a_k$ where $k = \text{argmin}_{i \in \{1,2,3,4\}} |x_i - x_5|$. In this task, the division of the training set and test set is done through the ``Modular-residue data separation'' method. We use a total of 30000 data points for training, with a batch size of 50, and train for 4000 epochs.
    % we maintain the conditions $a_i \in [8]$ and $a_i \neq a_j$ for all distinct $i$ and $j$, along with $x_i \in [20,100]$. The input sequence takes the form $(x_1, a_1, x_2, a_2, x_3, a_3, x_4, a_4, x_5)$, with the target denoted as $a_k$, where $|x_k - x_5|= \text{min}_{i \in [4]} |x_i - x_5|$.
    
\begin{figure}[ht]
        \centering
        \includegraphics[width=0.9\linewidth]{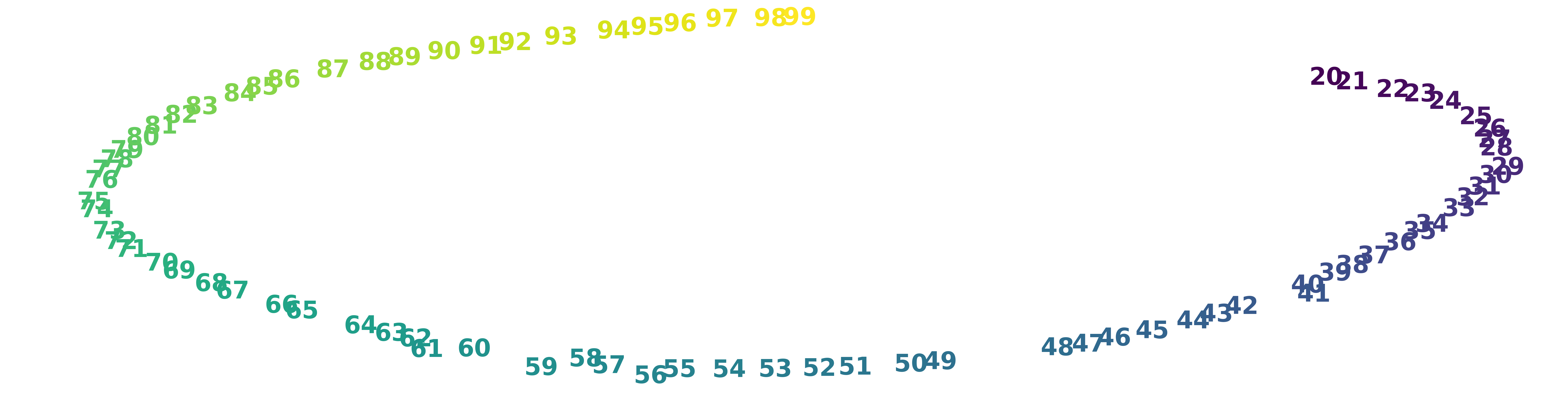}
        \caption{The t-SNE distribution of token embeddings after training for numbers from 20 to 99.}
        \label{fig:classification}
    \end{figure}

    As shown in Fig. \ref{fig:classification}, we visualize the token embedding by t-SNE \cite{van2008visualizing} after training. Obviously, the embedding captures the order of tokens, which is crucial in this classification task.
    % Simultaneously, we illustrate the t-SNE distribution of vectors post-embedding layer processing for numbers ranging from 20 to 100 in Fig. \ref{fig:classification}. This t-SNE distribution visually captures the learned ordinal relationships between the numbers, a crucial aspect of encoding magnitude relationships in the classification task.

    \subsection{Composite task}
        The composite task is a natural extension of one-anchor tasks. In each input sequence, there are always two anchors to indicate a composite function. A fundamental question arises: within the composite task, can the model learn the function of each individual anchor that constitutes the composite function? In this task, we separate the data by both modular-residue data separation and composite-anchor separation. Therefore, we can study the generalization of transformer on data and tasks. We also study how model size affects the performance.
        
        % Additionally, how does the model perform on composite tasks corresponding to anchor combinations not encountered in the training set, i.e., the generalization on task? In this section, we delve into the generalization capability of composite functions in handling both data and tasks. Meanwhile, we investigate the influence of the model size on the generalization ability. 
        
        Initially, we define four anchors, each representing a specific operation. These anchors are then paired up to create 16 distinct composite anchor functions. 15  pairs are selected to construct the training set.  The model's generalization ability is tested across all 16 operations. Notably, the model's generalization on the 15 composite operations used for training reflects its ability to generalize on data, while its generalization on the composite operations absent from the training set reflects its ability to generalize on tasks.

        \begin{figure}
            \centering
            \includegraphics[width=0.8\linewidth]{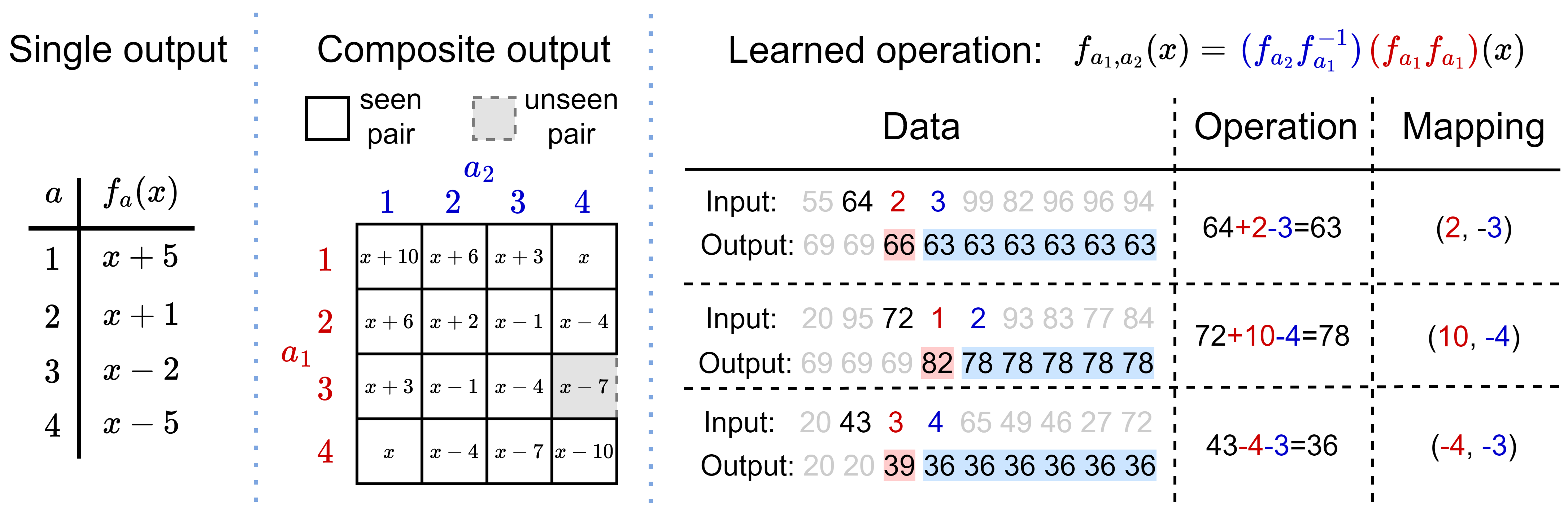}
            \caption{Single output: Function mapping of one anchor. Composite output: The two-anchor composite function mapping obtained after the composition of two one-anchor functions. Learned operation: The mapping operation of the composite function actually learned by the model. The examples shown represent the actual outputs of the model for the given inputs, and the corresponding operations on the examples support the learned operation of the model.}
            \label{fig:composition_drawio}
        \end{figure}

        As shown in the left in Fig.~\ref{fig:composition_drawio}, we denote $f_a$ for the function of anchor $a$ and combine each single anchor to obtain the two-anchor composite function $f_{a_1,a_2}$ of the anchor pair $\{a_1, a_2\}$. The gray anchor pair in the ``Composite output'' part does not appear in the training set. In the right in Fig. \ref{fig:composition_drawio}, we illustrate the model with 10 layers after training via three input examples. We found the network learns a specific composite function. For the first anchor $a_1$, network outputs $(f_{a_1}f_{a_1})(x)$, where $x$ is the key item. For the second anchor, network outputs $(f_{a_2}f_{a_1}^{-1})(f_{a_1}f_{a_1})(x)$, which cancels one operation of $f_{a_1}$.

        We found model size can significantly affect the output pattern after training. In the case of the model with 10 layers, as shown in the right in Fig. \ref{fig:composition_drawio}, after the input of the second anchor, the network outputs the correct answer and keeps broadcasting the correct answer for the subsequent input tokens. To characterize the output pattern for a specific composite of anchors, we define the mapping sequence for an input as follows.

        \begin{definition} [mapping sequence]
        Consider a model $f_{\vtheta}(\cdot)$ and an input sequence \(\vx = (x_1, x_2, \ldots, x_n)\),  where \(i\) denotes the index of the key item, and \(i+1\) and \(i+2\) represent the anchor indexes. The model output is \(f_{\vtheta}(x) = (f_{\vtheta}(x_1), \ldots, f_{\vtheta}(x_n))\). Then, define the sequence \((f_{\vtheta}(x_{i+1})-x_i, f_{\vtheta}(x_{i+2})-f_{\vtheta}(x_{i+1}), \ldots, f_{\vtheta}(x_j)-f_{\vtheta}(x_{j-1}))\) as a mapping sequence of the model \(f_{\vtheta}\), where \(f_{\vtheta}(x_k) = f_{\vtheta}(x_{k-1})\) for any \(k>j\) and \(f_{\vtheta}(x_j) \neq f_{\vtheta}(x_{j-1})\).
        \end{definition}

\begin{figure}
            \centering
            \subfloat[]{\includegraphics[height=0.313\linewidth]{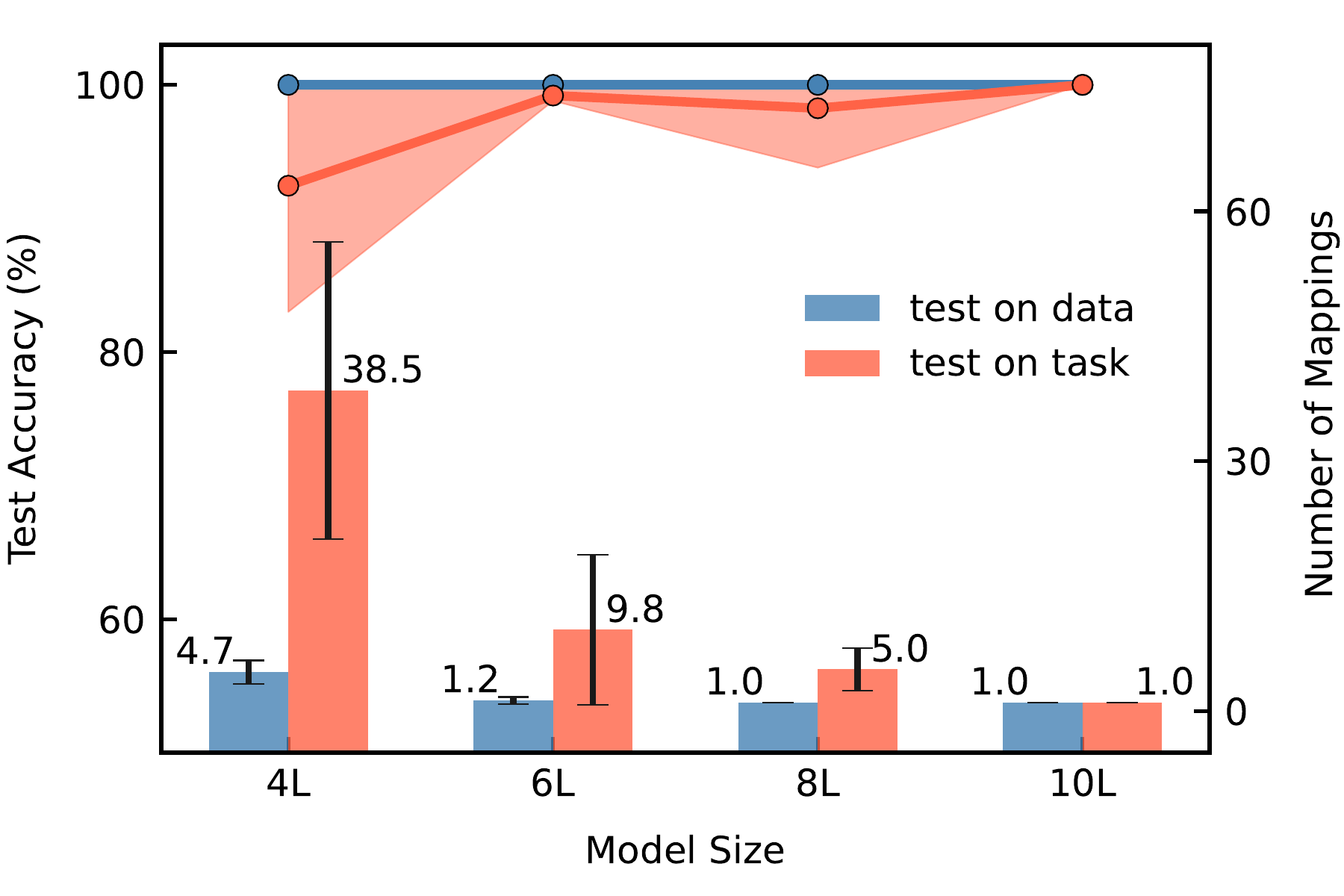}}
            \subfloat[]{\includegraphics[height=0.3\linewidth]{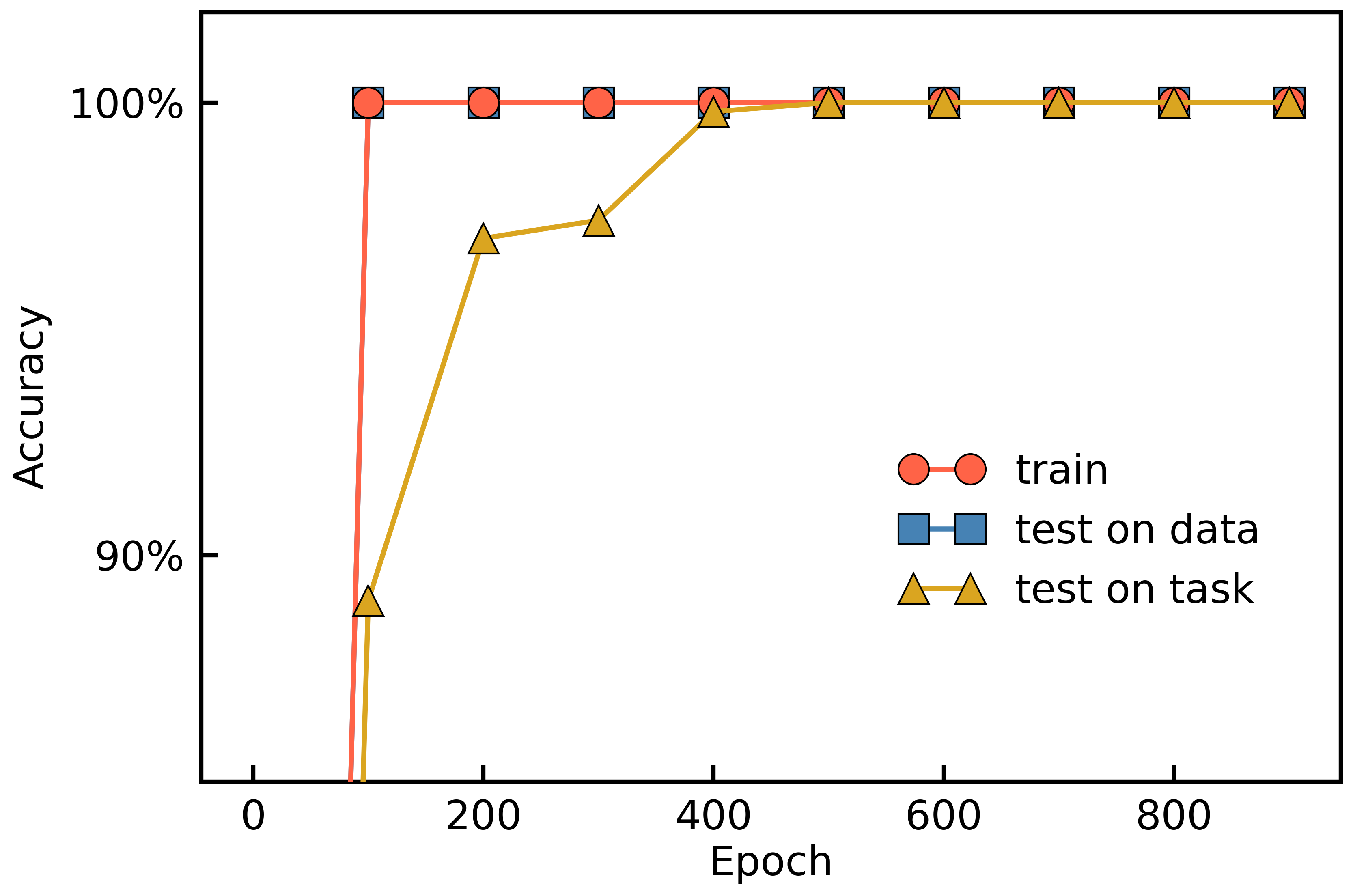}}
            \caption{(a) The relationship between test accuracy (left ordinate, curve) and the number of mapped sequences (right ordinate, bar) with the model size, where ``L'' represents the layer. The blue color reflects the generalization ability on data, while the red color reflects the generalization on task. Each configuration is tested five times independently, and the mean is taken. The shaded area and error bars respectively represent the variance in test accuracy and the number of mapped sequences. (b) The training accuracy and test accuracy of the composite task. The convergence speed of the model's generalization on the data is much faster than its convergence speed on the task. }
            \label{fig:composition}
        \end{figure}

        As shown in Fig. \ref{fig:composition}(a), for the case of the model of 10 layers, there is only one mapping sequence for each composite of anchors. However, when the model size is smaller, it shows several different mapping sequences for each composite of anchors. This result depending on the model size is an interesting phenomenon for further study. In addition, the test accuracy on task is usually smaller than the test accuracy on data (always achieves $100\%$) after training. This suggests the generalization on task is more difficult. To verify this, we display a learning process in Fig. \ref{fig:composition}(b). The training accuracy and the test accuracy on data achieve $100\%$ much earlier than the test accuracy on task.

        % % 需要有一段关于叠词作用的说明
        % \textbf{There needs to be an explanation here about the role of overlapping words.}

        % \subsection{Anchor length}
        
        % \begin{figure}
        %     \centering
        %     \includegraphics[width=0.5\linewidth]{pic/anchorlength_task.pdf}
        %     \caption{Enter Caption}
        %     \label{fig:anchor_length}
        % \end{figure}

    \subsection{Working memory task}

        The composite task demonstrates the model's ability to learn a single anchor operation from composite anchors. The second anchor processes the output from the first anchor instead of directly processing the key item. This sequential processing necessitates working memory capabilities in the model if there are some irrelevant items between the two anchors. 
        
        % , enabling it to use the first anchor's output as input for the second anchor. To further examine the working memory capabilities, we separate the two anchors in the input sequence and observe how information is transmitted when the output shifts from the first to the second anchor. We have simplified the experimental setup for this task. 
        
\begin{figure}
            \centering
            \includegraphics[width=0.8\linewidth]{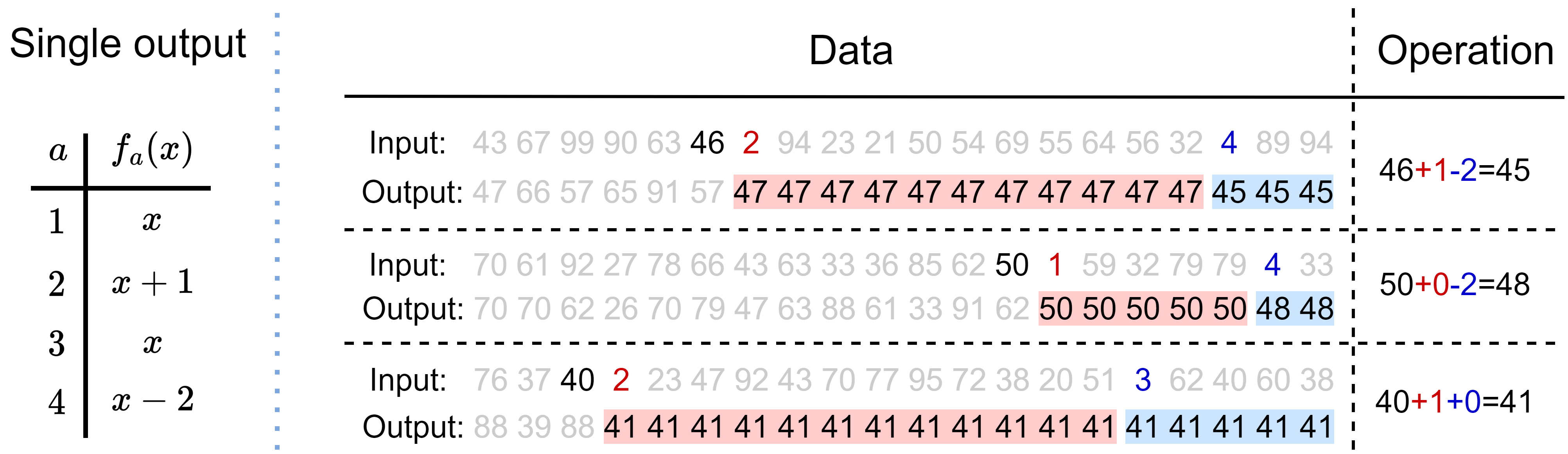}
            % \subfloat[]{\includegraphics[width=0.28\linewidth]{pic/working_memory.png}}
            \caption{Single output: Mapping function for one anchor. Where $a_1 \in \{1, 2\}, a_2 \in \{3, 4\}$. Data: Input sequence and its corresponding output from the model. The red shadow represents the working memory for $a_1$, and the blue part corresponds to the working memory for $a_2$. Operation: The learned mapping for the anchors by the model.}
            \label{fig:workong_memory}
        \end{figure}

        Specifically, we let the first anchor \(a_1 \in \{2, 3\}\) and the second anchor \(a_2 \in \{4, 5\}\), with their corresponding mappings illustrated in Fig. \ref{fig:workong_memory}. Notably, the model observes only the final output and not the intermediate computations. Therefore, as long as the final output of two composite anchors is the same as the given labels, the network can learn a different function for a single anchor operation, compared with our given operation for the anchor.
    
         As depicted in Fig. \ref{fig:workong_memory}, the first anchor's output is the intermediate result via the operation of the first anchor on the key item. The model memorizes this intermediate result by keeping out this value for the input of irrelevant tokens. Once the model sees the second anchor, the intermediate result is no longer useful and the model outputs the final correct answer.

    \subsection{Synonym task}

\begin{figure}
            \centering
            \subfloat[]{\includegraphics[width=0.71\linewidth]{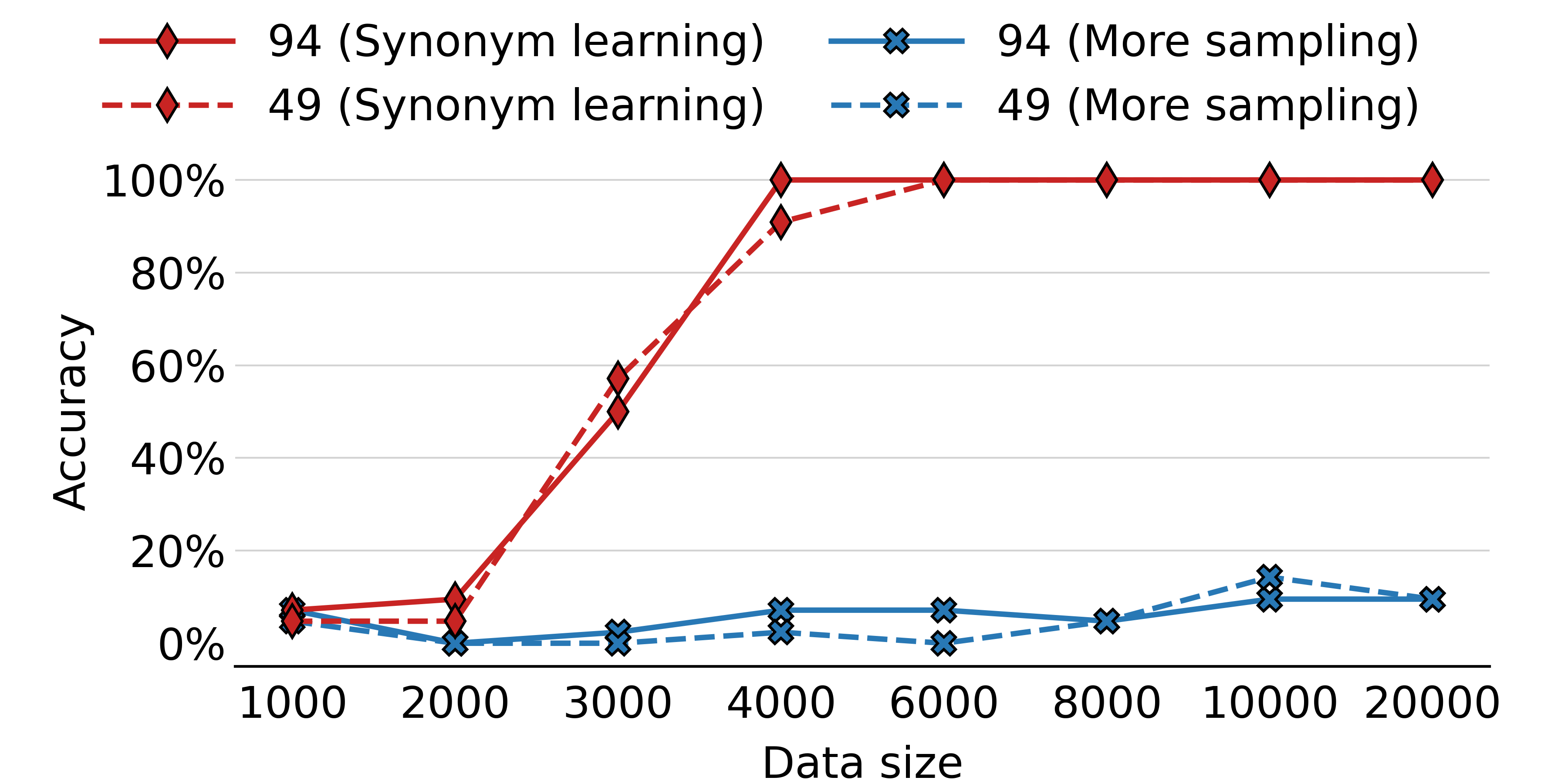}}
            \subfloat[]{\includegraphics[width=0.26\linewidth]{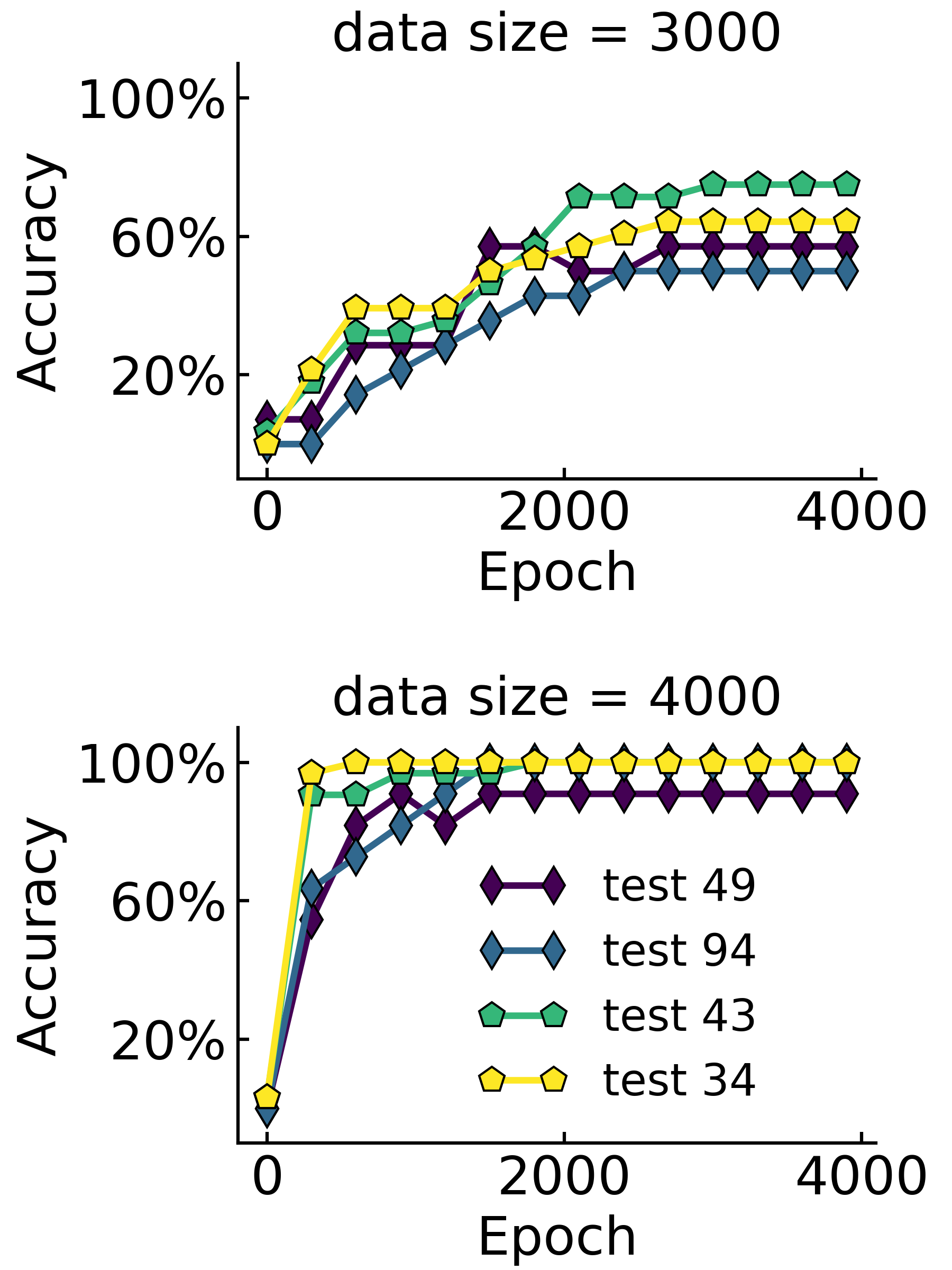}}
            \caption{(a) Compare the impact of two types of data enhancement methods on task generalization. These types of data enhancement methods include adding synonyms (red) and sampling more data with the original composite anchors (blue). (b) The test accuracy of transformer predictions on data with unseen anchors in the synonym task. The composite anchors (4,9) and (9,4) are unseen anchors. training data size for the upper and lower cases are 3000 and 4000, respectively.}
            \label{fig: compare near synonym learning}
        \end{figure}

        The synonym task is an extension of the composite task, aimed at studying the capacity of language models to learn the synonymous relationship between two anchors. The composite anchors in the original dataset are taken from: 
        \{(1,9), (2,9), (9,1), (9,2), (1,2), (1,4), (2,1), (4,1), (2,4), (4,2)\}. The number of the training data set for each composite anchor is the same. However, the transformer network struggles with learning composite anchors such as (4,9) and (9,4) under the above setup for different total training data sizes, as shown by the blue curves in Fig. \ref{fig: compare near synonym learning}(a).
        
        We set an anchor ``3'' that performs the same function as anchor ``9''. To see whether transformer networks can learn the function of anchors (4,9) and (9,4) through the composite anchors (4,3) and (3,4), we add the data with synonym anchor ``3''   to the training set. Specifically, we integrate composite anchors \{(1,3), (3,1), (2,3), (3,2), (4,3), (3,4)\} into the original dataset. In the case with a total number of training data as $M$, the number of training data for composite anchors that contain ``9'' (such as (1,9)) is fixed as 100, while for other composite anchors is the $(M-400)/12$. As shown by the red curves in  Fig. \ref{fig: compare near synonym learning}(a), the network can accurately predict the output of  (4,9) and (9,4) as long as the data size is larger than a certain value. In  Fig. \ref{fig: compare near synonym learning}(b), we display the learning process for cases of two data sizes. Obviously, the learning of (4,9) and (9,4) is slightly later than (4,3) and (3,4). These results demonstrate that introducing synonyms significantly impacts the transformer's task generalization.

    \subsection{Forward-backward recitation task}
    
        % When we ask an LLM for the next part of a poem, its accuracy in answering is often higher than that in answering the previous part of the poem. This task will reflect this asymmetry of the LLM, and explore the impact of loss spike on accuracy. The proportion of forward tasks and reverse tasks in the data set we constructed accounts for 50\% each. Judging from the training results shown in Fig.~\ref{fig: context learning}, no matter from the beginning or the end of the training process, the accuracy of the neural network on the forward task is always higher than the reverse problem. And as the amount of data increases, the reverse problem becomes more likely to be overfitted.

\begin{figure}
            \centering
            \includegraphics[width=0.7\linewidth]{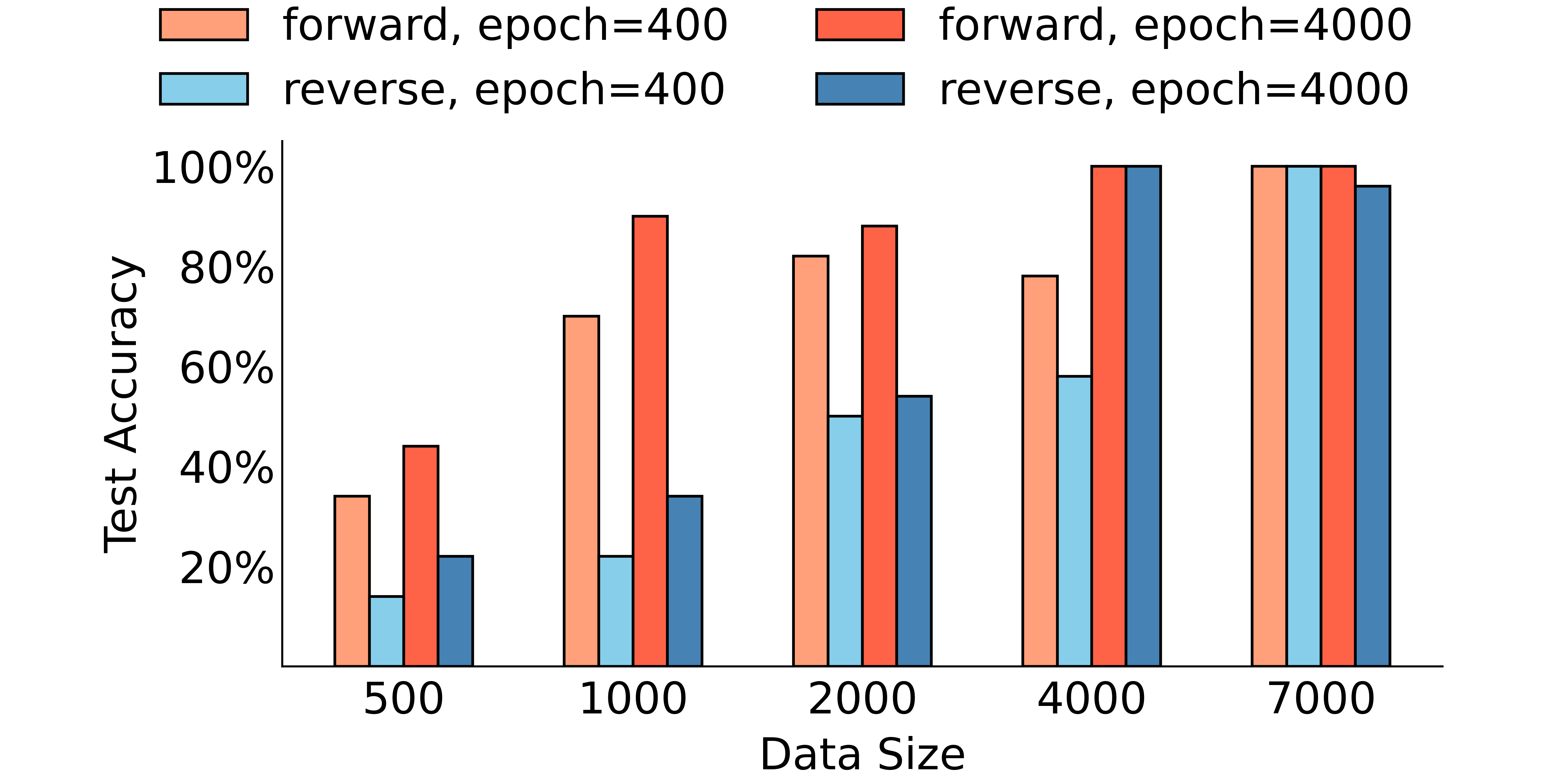}
            % \subfloat[]{\includegraphics[width=0.33\linewidth]{pic/loss_spike_forward_backward.png}}
            \caption{Accuracy of forward prediction and backward prediction in forward-backward recitation task changes with training data size. The figures show the accuracy of the test data when training to 400 epochs and 4000 epochs respectively. }
            % (b) The training process of the forward-backward recitation task under different amounts of training data. The solid red line is the training error. The dotted solid lines are the accuracy rates of the forward task and the backward task respectively.}
            \label{fig: context learning}
        \end{figure}

        Usually, one finds that reciting the next line of a poem (forward task) is much easier than reciting the previous one (backward task). Empirically, LLM is also found to be easier for forward tasks than backward tasks. For example, we use GPT4 to test the forward-backward recitation task with 20 poems. In the task of forward recitation, GPT4's accuracy is 95\%, while the accuracy of the backward recitation is only 5\%.
        % When asking an LLM for the subsequent line of a poem (forward task), it generally exhibits higher accuracy compared to responding to the previous line (backward task). 
        Based on this observation, we design an idealized forward-backward recitation task to study this asymmetry of LLMs.
        % , highlighting LLMs' asymmetry and examining the effect of loss spike on their generalization capacity. 
        An input sequence follows the pattern ``($*$,$*$,$m$,$3$,$n$,$*$,$*$,$*$,$n$)'' or ``($*$,$*$,$m$,$3$,$n$,$*$,$*$,$*$,$m$)'', where ``3'' is the anchor, and others are tokens randomly selected from $[20,100]$, ``$*$'' indicates that the token in this position is irrelevant, and ``$m$,$3$,$n$'' can be any position. The output of ``($*$,$*$,$m$,$3$,$n$,$*$,$*$,$*$,$n$)'' is ``$m$'', while the output of ``($*$,$*$,$m$,$3$,$n$,$*$,$*$,$*$,$m$)'' is ``$n$''. As shown in Fig. \ref{fig: context learning}, the transformer network's generalization ability in forward tasks is consistently superior to that in backward tasks.

    \subsection{Statistical output task}

        Statistical output is a common scenario, i.e., the output of a sequence follows a probability distribution. To simulate this scenario, we design a task based on the identity learning anchor function. 
        % as follows: We first build dataset \(S\) similarly to the identity learning task and make five copies of it. The label of the last copy is altered to \(x+1\). These five copies make up the fairness learning dataset. Fig.~\ref{fig: fairness learning}(a) shows that the transformer is inclined to produce labels that are more frequently represented.

 We examined two types of dataset construction: 
 \begin{itemize}
     \item The dataset \(S\) is built in the same way as the identity learning task and replicated five times, modifying the label of the final copy to \(x+1\). These five datasets constitute the first type of fairness learning dataset.

     \item the dataset \(S\) is built in the same way as the identity learning task and randomly changed the labels of 20\% of the data to \(x+1\). The newly obtained dataset is called the second type of fairness learning dataset.
 \end{itemize}
 
\begin{figure}[ht]\centering
            \subfloat[]{\includegraphics[width=0.33\textwidth]{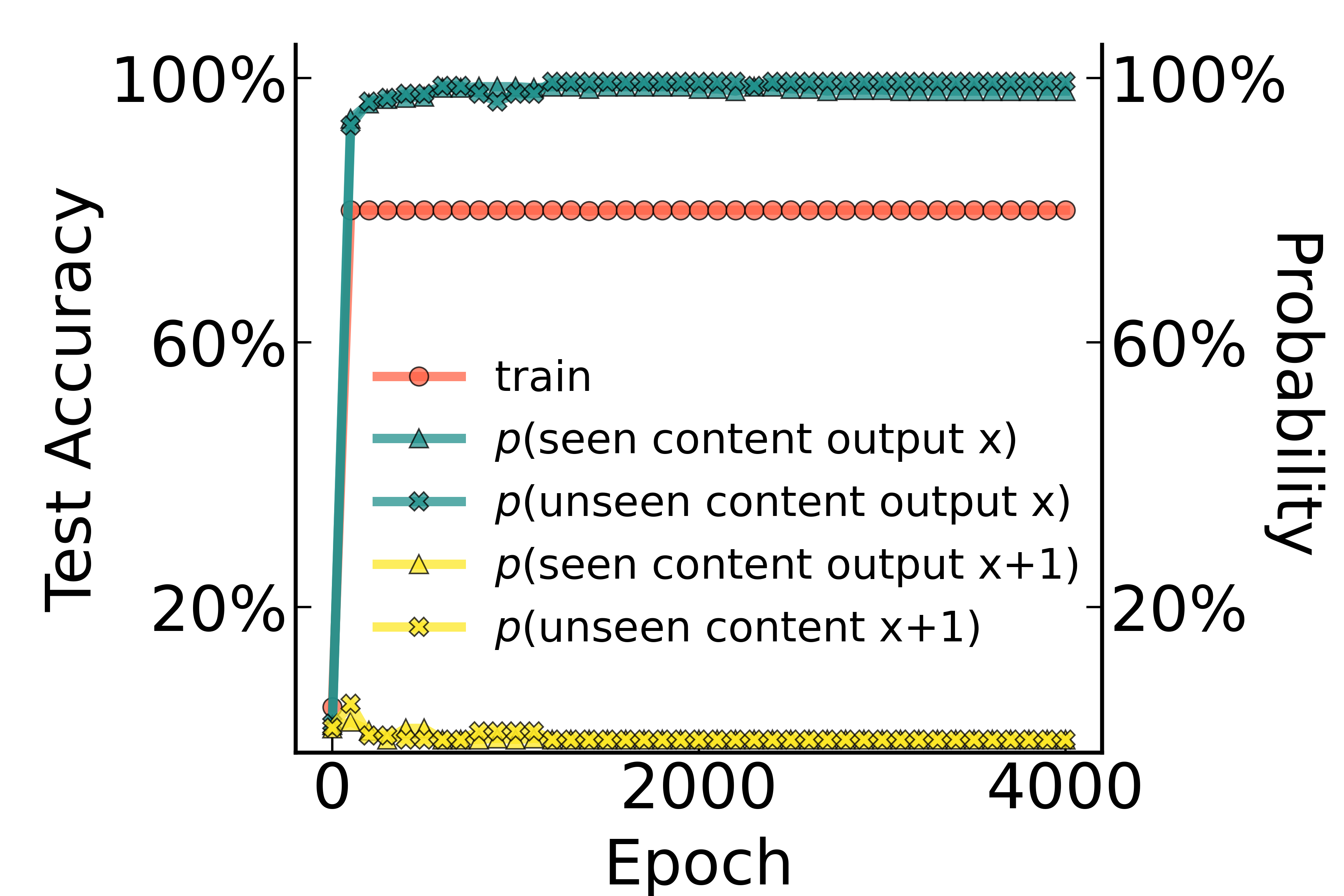}}
            \subfloat[]{\includegraphics[width=0.33\textwidth]{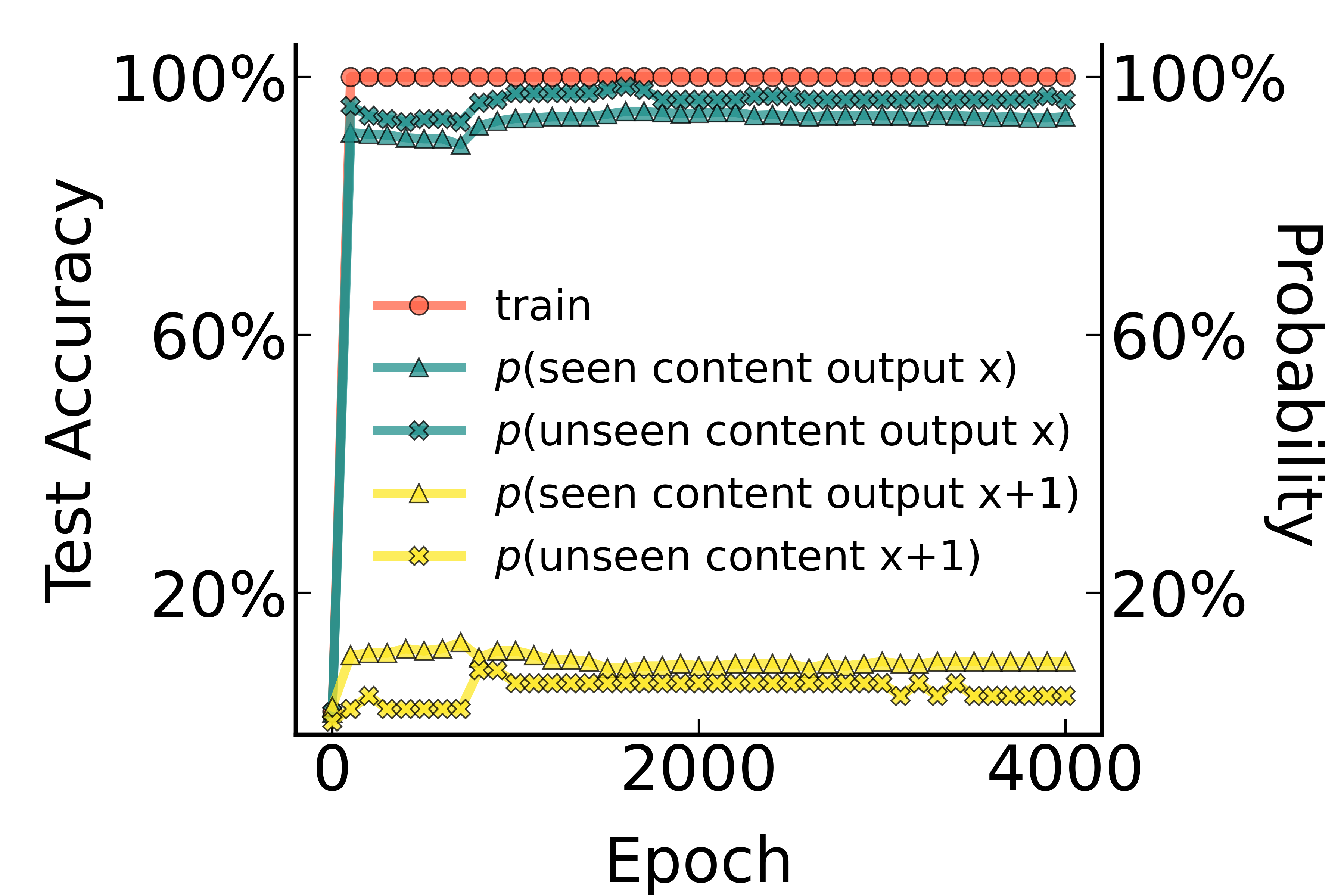}}
            \subfloat[]{\includegraphics[width=0.33\textwidth]{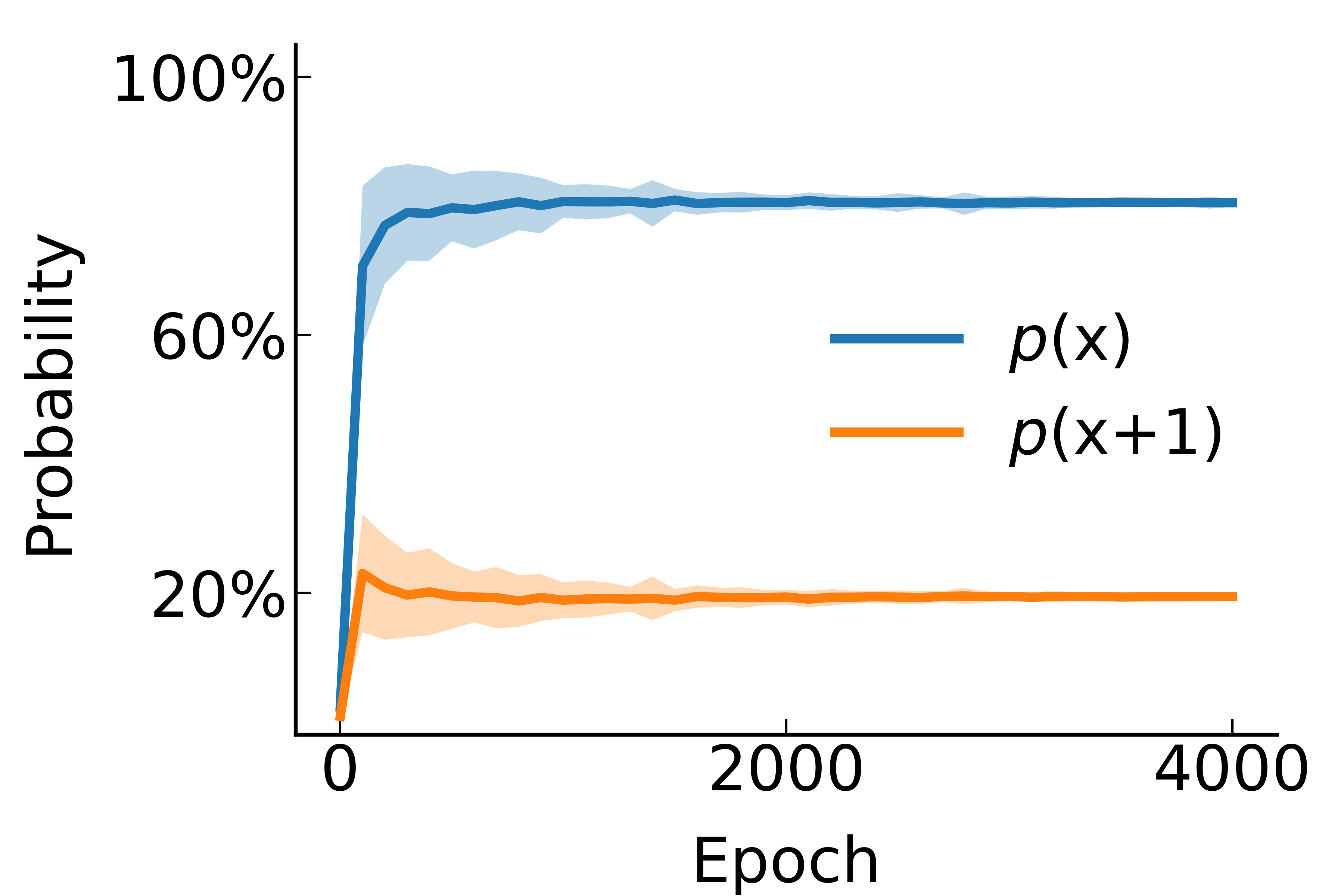}}
            \caption{(a, b) The accuracy of transformer predictions in the statistical output task. The transformer is trained with the first type of dataset (a) and the second type of dataset (b). The training data are composed of 80\% ``$3x$ to $x$'' and 20\% ``$3x$ to $x+1$'' sequences. The total data size of the training set is 9000. Then test the probability of the transformer output $x$ and output $x+1$ in the newly generated train data set (seen content) and test data set (unseen content), which are distinguished according to whether the remainder of modulo 8 is 0. (c) The probability of token ``x'' or ``x+1'' after the projection layer.}
            \label{fig: fairness learning}
        \end{figure}

   As shown in Fig.~\ref{fig: fairness learning}(a), for the first type of data, the training accuracy can only achieve $80\%$ (red). This is because we only consider the output with the largest probability, while the output for an input is statistical in the training dataset. For the data with label $x$, the network can achieve $100\%$ while $0\%$ for the data with label $x+1$. Results are similar for the second type of data as shown in Fig.~\ref{fig: fairness learning}(b). 

   We next examine if the network learns the probability distribution of output. For input, we can record the probability of output $x$ and $x+1$, respectively, i.e., softmax of the output layer.
   As shown in Fig.~\ref{fig: fairness learning}(c), for the first type of data, as the epoch increases, the probability of output $x$ increases towards $80\%$ while $x+1$ towards $20\%$, both of which satisfy the experimental setting.

   % both types of datasets, the transformer networks indeed tend to output labels with a larger proportion. Furthermore, we observe the probability distribution of each data being output as \(x\) or \(x+1\) after passing through the transformer's projection layer. We find that the mean of probability distribution matches the probability distribution of the label. Moreover, as training progresses, the variance of this probability gradually decreases.

    \subsection{Multi-anchor task}
                
        % 3x to x   4x to x  不同区间
        % The multi-anchor task focuses on the generalization capability of multiple different anchors across different intervals. As shown in Fig.~\ref{fig: tasks}, we used 10 different anchor $a\in [1,2,3,4,5,6,7,8,9,10]$. And define $f_a(x)=x+a$, and $I_a=[20, 100]\backslash[20+8*a, 20+8*(a+i)-1]$. Our focus is on whether the transformer's prediction ability can be generalized to statements where the anchor is $a$ and $x$ is taken from $I_a^c$ after training. 

        In the case of modular-residue data separation, it only ensures different positions for the pair of anchor and key items i.e., (\(a,x\)), in the training and test sets. One might argue the network only learns shift invariation but not the operation. To verify the learning of operation, we design the following multi-anchor task.

        We utilize ten distinct anchors \(a \in \{1,2,3,4,5,6,7,8,9,10\}\) and define the anchor function \(f_a(x) = x + a\), where $x$ is the key item next to the anchor. For the test dataset, the key item for anchor $a$ belongs to $G_{a}^{c}=[12 + 8a, 12 + 8(a+1)-1]$, and 
        for the training set,  the key item for anchor $a$ belongs to $G_{a}=[20, 100]\backslash G_{a}^{c}$, i.e., anchor-based data separation.
        % , along with \(G_a = [20, 100] \cap [20 + 8a, 20 + 8(a+i)-1]^C\). 
        Our focus is on whether, after training, the transformer's predictive ability can generalize to sentences where the pair of anchor and key item is unseen in the training. 
        % The primary challenge in this task arises because the test set's anchor-key item combinations, namely (\(a,x\)), have never occurred in the training set. This contrasts with previous tasks, which only ensured different positions for (\(a,x\)) in the training and test sets.

\begin{figure}[h]
            \centering
            \includegraphics[width=0.7\linewidth]{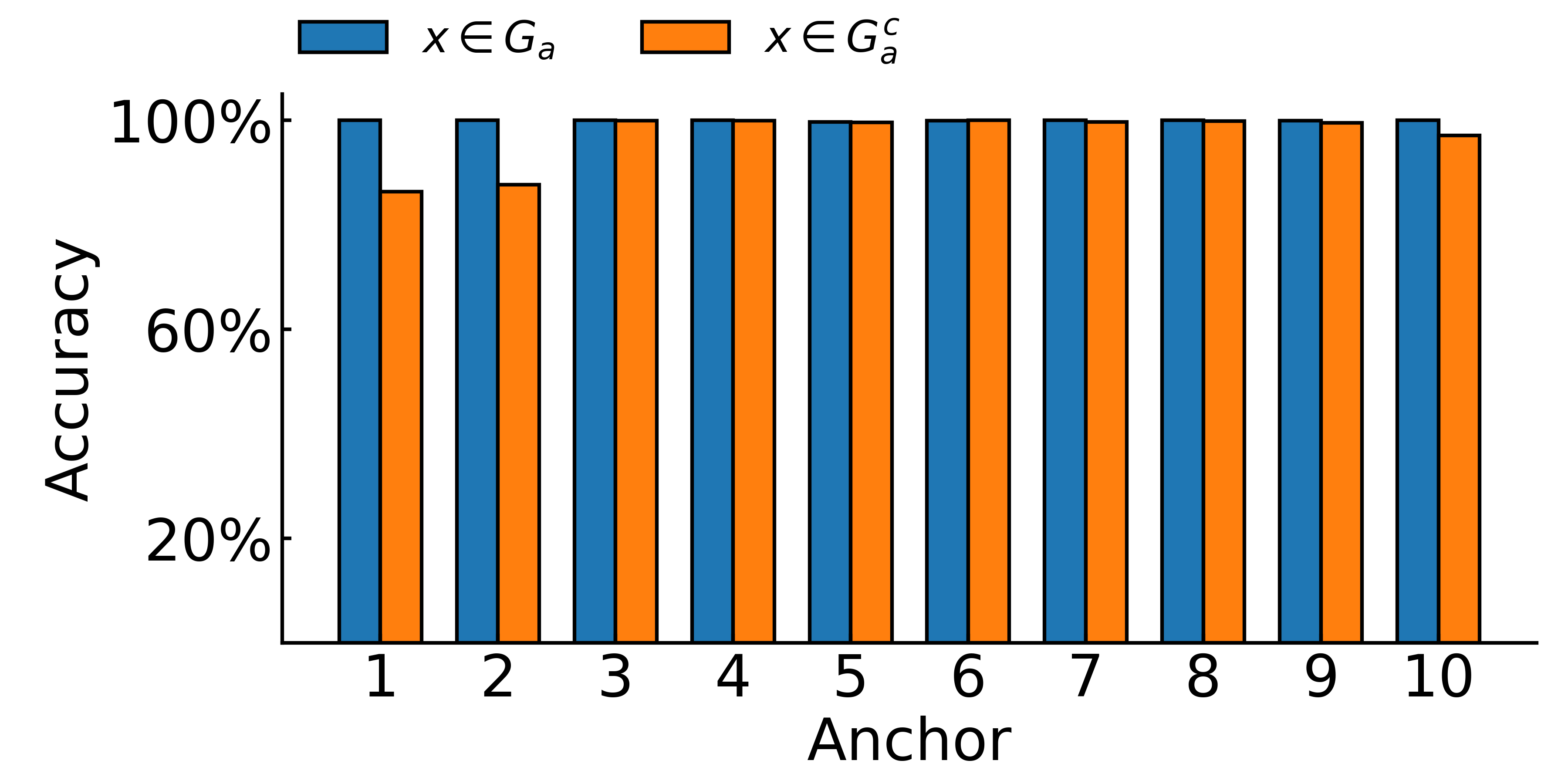}
            \caption{Accuracy of multi-anchor task after training. We examined the predictive ability of transformer on data sets constructed with different anchors. The test set is $G_{a}^{c}=[12 + 8a, 12 + 8(a+1)-1]$, and the training set is $G_{a}=[20, 100]\backslash G_{a}^{c}$.}
            \label{fig: multianchor-multiinterval acc}
        \end{figure}

        As shown in Fig. \ref{fig: multianchor-multiinterval acc}, in most cases, the model exhibits good generalization capabilities on anchor-based data, i.e., the network learns the operation rather than shift-invariance. For the cases near the boundary (anchor ``1'', ``2'', ``10''), due to the construction method, the frequency of their output tokens is smaller than other anchors, which may be the reason why the learning boundary anchors are more difficult.
        % Additionally, it is noted that for cases where \(x\) is near the boundary, such as anchor \(a=1\), \(x \in G_1=[20, 27]\), the model's generalization is worse compared to other cases. This may be due to these data points only appearing in the input data and not in the output data. For instance, there is no case in the training data where \(x + a = 21\).

\section{Mechanism study of the identity learning task} \label{sec:mechanism}

In this section, we first provide a brief explanation of the mechanism of the identity learning task by identifying two key operations in a two-layer model, i.e., shift and broadcast. We then conduct a series of ablation experiments to explore the role of each structure in the transformer and find a very simplified two-layer model that can accomplish the identity learning task. The operations of shift and broadcast also play key roles in this simplified model. Finally, we show these two operations are also common in LLMs and have a short discussion. A very detailed discussion of the mechanism for the identity learning task can be found in Appendix \ref{sec:mechanism}.

% We then conduct a detailed analysis of this mechanism using a simplified model, followed by a discussion on the generality of this attention-based information transfer mechanism. Due to space constraints, the detailed analysis of the two-layer model mechanism can be found in the appendix.

\subsection{A brief explanation of a two-layer model}

\begin{figure}[ht]
    \centering
    \subfloat[first layer, shift operation]{\includegraphics[width=0.4\linewidth]{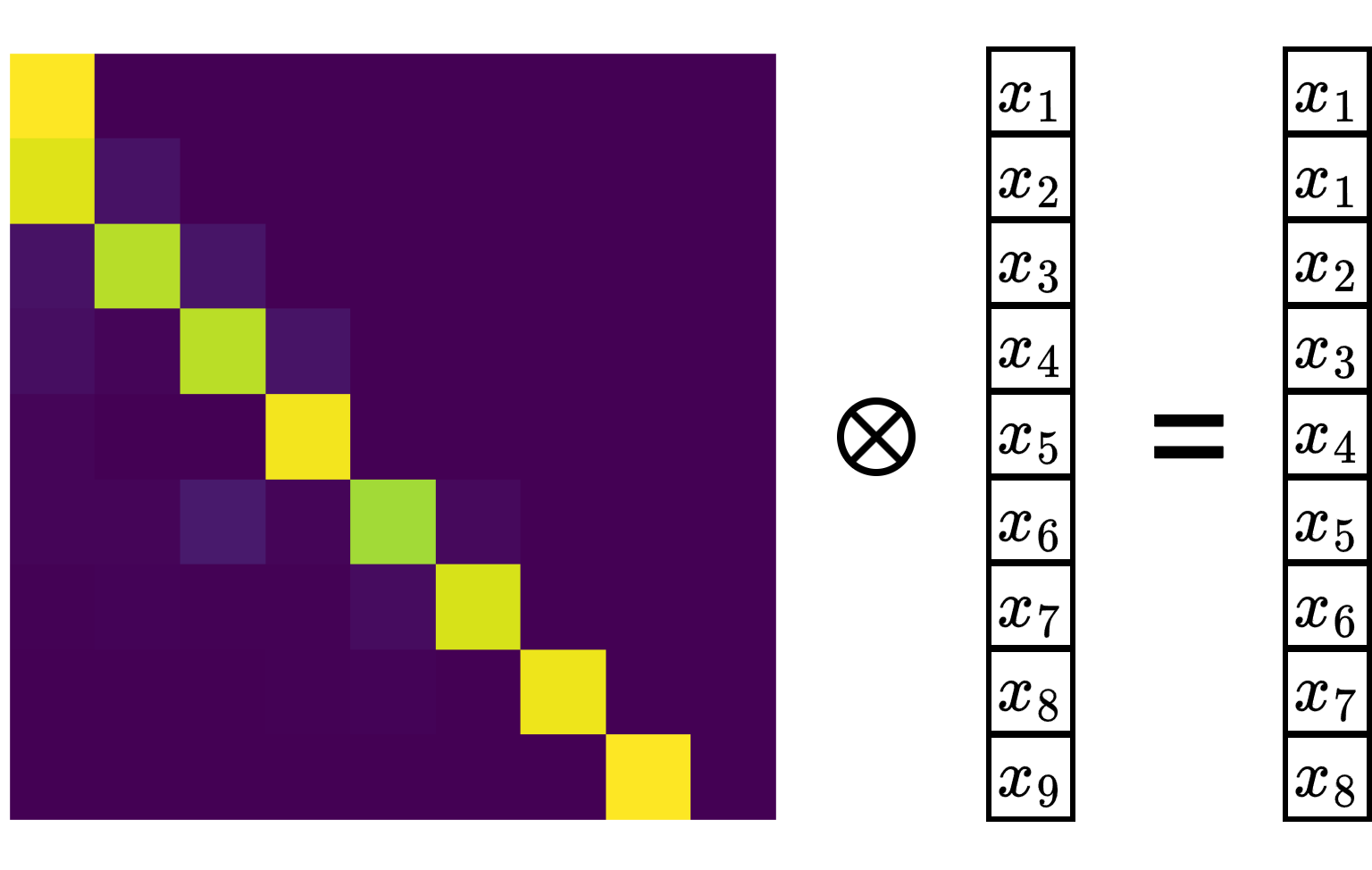}}
    \hspace{.5in} 
    \subfloat[second layer, broadcast operation]{\includegraphics[width=0.4\linewidth]{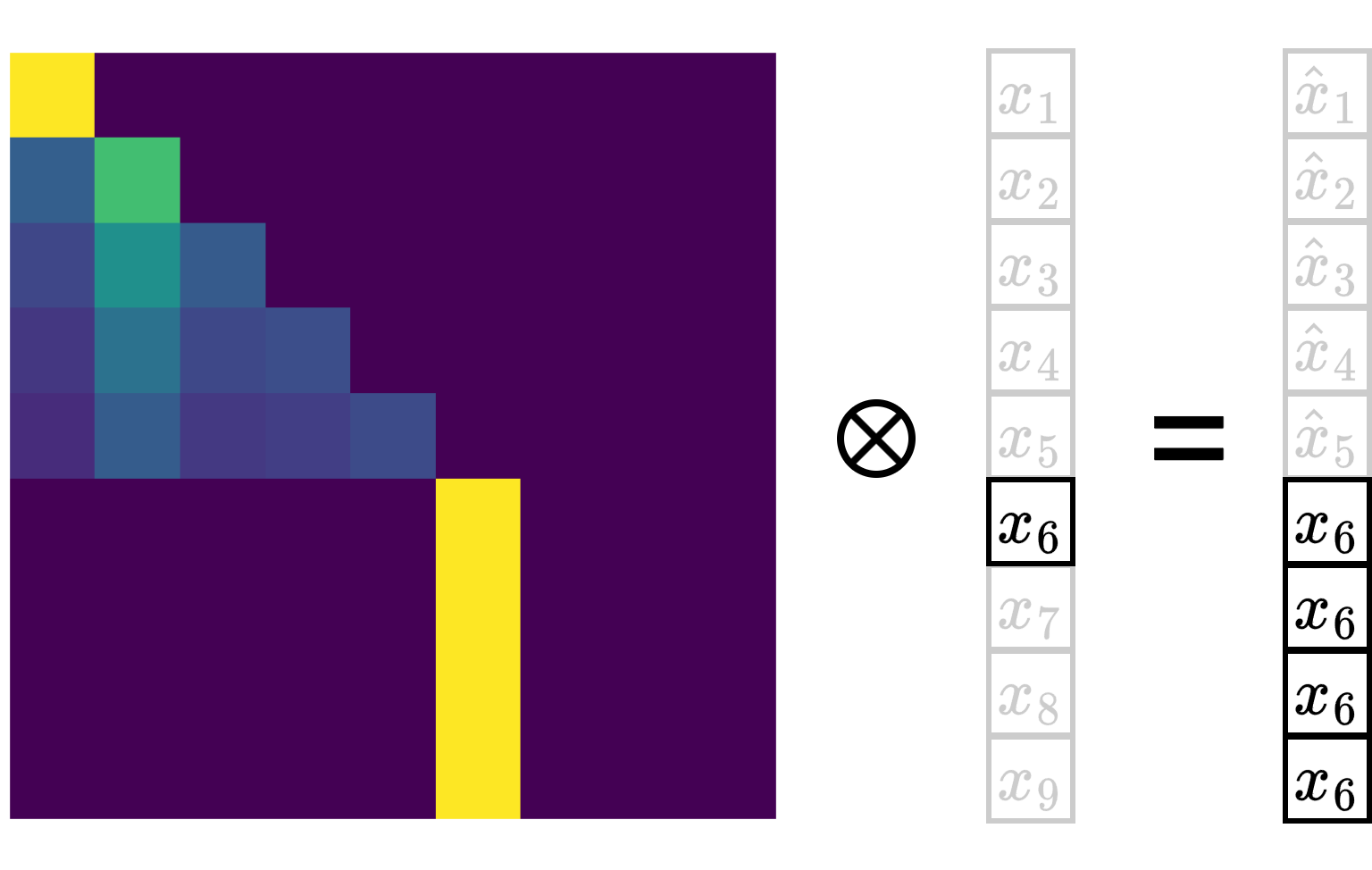}}
    \caption{Attention maps of the first and second layer.}
    \label{fig:two_layer_attn}
\end{figure}

In the task of identity learning, the attention maps of the two-layer model primarily serve two purposes: shift and broadcast. The first layer's attention map performs a shift operation, as illustrated in Fig. \ref{fig:two_layer_attn}(a), the sequence ($x_1$,$x_2$,$x_3$,$x_4$,$x_5$,$x_6$,$x_7$,$x_8$,$x_9$) is shifted to ($x_1$,$x_1$,$x_2$,$x_3$,$x_4$,$x_5$,$x_6$,$x_7$,$x_8$) through the first layer's attention map. Based on the residual connection, the shifted vector and the unshifted one is fused, 
% derived by the first layer's attention map, the ResNet within the attention block can fuse the information from the shifted and unshifted sequences, 
thereby establishing the integration of information between the anchor item and the key item. 
It is noteworthy that the operation is a one-position shift, due to the fact that the key item is designed to be the next token of the anchor. We will discuss more on how the data structure affects the shift operation.
% this attention structure is not influenced by the anchor position in a single input data but is determined by the data structure in the training dataset.

Once the information of the anchor and the key item is fused, the subsequent fully-connected structure will render the correct label based on this fused information.

The second layer's attention map executes a broadcast operation. As shown in Fig. \ref{fig:two_layer_attn}(b), the sequence ($x_1$,$x_2$,$x_3$,$x_4$,$x_5$,$x_6$,$x_7$,$x_8$,$x_9$) is transferred to ($\hat{x_1}$,$\hat{x_2}$,$\hat{x_3}$,$\hat{x_4}$,$\hat{x_5}$,$x_6$,$x_6$,$x_6$,$x_6$), where $\hat{x_i}$ is the linear combination of $\{x_1, \ldots, x_i\}$ but is irrelevant to obtaining the correct label. This broadcast operation transfers the correct label to the last position, namely the output position. The first layer's attention map is mainly based on the position embedding and the second layer's attention map is mainly based on the positions of anchors in the input sequence.

This example illustrates two important operations that attention structure provides in language models, i.e., shift and broadcast. We also find that these two operations are also common in LLMs (see experiments in later sections). Therefore, the anchor function can serve as a type of benchmark function for understanding transformer-based language models.

To realize the attention maps with shift and broadcast operations, one possible way is to tune vector orientations in high-dimensional space. Two random vectors in a high-dimensional space are orthogonal with high probability. For the shift attention, the training needs to tune the $i$-th row of matrix $Q$ to parallel with the $(i-1)$-th row of matrix $K$. For the broadcast one, the training needs to identify the anchor token. The fully-connected network non-linearly transforms all tokens to parallel ones (called non-interested tokens for convenience), except for the interested one that has combined the anchor token ``3'' and the key token ``$x$''.  Rows of non-interested tokens in the matrix $Q$ (or $K$) are parallel and the interested one is orthogonal to others. For non-interested token positions, the direction of each row of $Q$ is parallel but opposite to the corresponding one of $K$, while for the interested position, parallel with the same direction. Then, the element in attention that does not contain the interested position will be very negative, leading to the broadcast operation.

Please refer to the appendix for a detailed analysis of the mechanisms of the two-layer transformer.

\subsection{Data structure's impact on the shift attention map}

The attention map in the aforementioned model is heavily influenced by structural features inherent in the training data, particularly the adjacency of key items with anchors in the identity learning task. In this section, we investigate the influence of data structures on the attention map. We generate data for the identity learning task using two anchors, namely ``3'' and ``4''. For the data corresponding to anchor ``3'', the key item is the digit following ``3'', and the task is represented as $(\ldots, 3,x, \ldots) \rightarrow x$. For the data corresponding to ``4'', the key item is the digit located two positions after ``4'', and the task is represented as $(\ldots, 4,*,x, \ldots) \rightarrow x$. This investigation is conducted within a transformer model comprising two layers, each containing two heads.

\begin{figure}
    \centering
    \subfloat[1st layer, 1st head]{\includegraphics[width=0.22\linewidth]{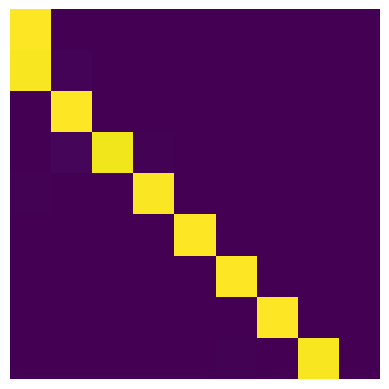}}
    \hspace{.15in} 
    \subfloat[1st layer, 2nd head]{\includegraphics[width=0.22\linewidth]{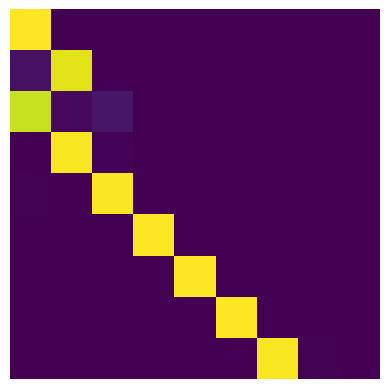}}
        \hspace{.15in} 
    \subfloat[2nd layer, anchor ``3'']{\includegraphics[width=0.22\linewidth]{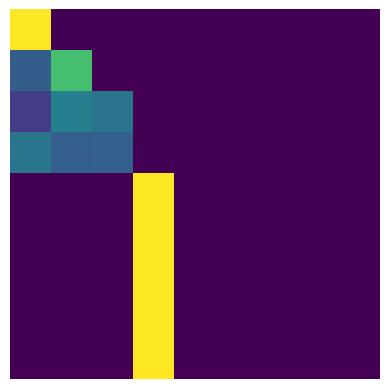}}
        \hspace{.15in} 
    \subfloat[2nd layer, anchor ``4'']{\includegraphics[width=0.22\linewidth]{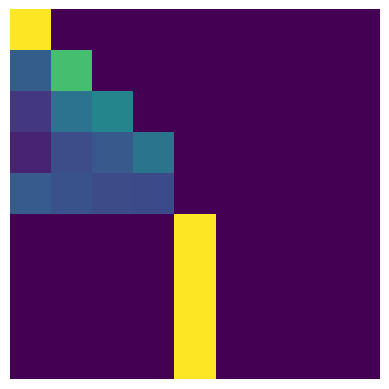}}
    \caption{(a, b) Attention maps of the two heads of the first layer. (c,d) Attention maps of two heads of the second layer for the input sequence with anchor ``3'' and ``4'', respectively.}
    \label{fig:two_layer_two_head}
\end{figure}

Fig. \ref{fig:two_layer_two_head}(a, b) illustrates the attention maps of the two heads in the first layer of the model. Both heads perform shift operations, with the first head, shown in Fig. \ref{fig:two_layer_two_head}(a), shifting by one position, aligning with the task corresponding to the anchor ``3'', while the other head, shown in Fig. \ref{fig:two_layer_two_head}(b), shifts by two positions, aligning with the anchor ``4''. This phenomenon is caused by the characteristics of the training data and the positional embedding obtained during training. During the testing phase, regardless of the input test sequence, these two attention maps will remain stable.

% The model activates distinct heads for the two different data structures, indicating diverse operational modes of the heads in the transformer model. 

For the attention maps of different heads in the second layer, they perform broadcast operations. Fig. \ref{fig:two_layer_two_head}(c, d) illustrates the attention map of the second layer head, taking ``3'' and ``4'' as anchors, respectively (only the first head is shown, and the other head is similar). When the positions of two anchors are the same, the positions of the key items corresponding to the two anchors are different. Therefore, the positions performing the broadcast operation for the attention map of anchor ``4'' are one position later than anchor ``3''.

\subsection{Ablation Experiment}

In the following, we empirically study how much we can simplify the decoder-only network structure, followed by analyzing a simplified model's mechanism in the identity learning task, which may be more accessible for future theoretical study.

A common block of a transformer network is shown in Fig.~\ref{fig:ablation experiment}(a). The detailed operation of each component is illustrated in Appendix~\ref{notation}. To find a model that can accomplish the anchor function but contains as few components as possible, we conduct ablation experiments of different components for the identity learning task with modular-residue data separation. 

Empirically, we find that the one-layer decode-only structure can not generalize at all for the identity learning task, while the two-layer one generalizes well. The reason why the one-layer structure fails is that each token is solely processed. Therefore, the fusion of different tokens cannot be done via the one-layer structure. If we manually shift vector $V$ as the shift operation, then, the network can easily fit the identity learning task.

We then conduct ablation experiments on various components of a two-layer model: ``res'' indicates the residual connection, ``linear'' indicates the linear transformation after attention operation, ``LN'' indicates the layer normalization, ``FNN'' indicates the feedforward neural network, ``projection'' is a linear transformation of the output of the decoder before the final softmax, ``mask'' indicates that each token can only see its previous tokens but not subsequent ones. It is worth mentioning that when we remove certain modules, the dimensions of some matrices within the network need to be adjusted accordingly to ensure that the model is well-posed. For example, when we delete the linear operation, we need the row vector dimension $d_v$ of $V^{(l)}$ to be the same as the row vector dimension $d_m$ of the hidden state; when we delete the projection layer, the hidden state row vector dimension $d_m$ needs to be the same as the dictionary size $d$. Beyond that, we adjust $d_m$, $d_k$, or $d$ so that the model used in each experiment has roughly the same amount of hyperparameters.

    % 消融实验
    \begin{figure}[ht]\centering
        \subfloat[]{\includegraphics[width=0.3\textwidth]{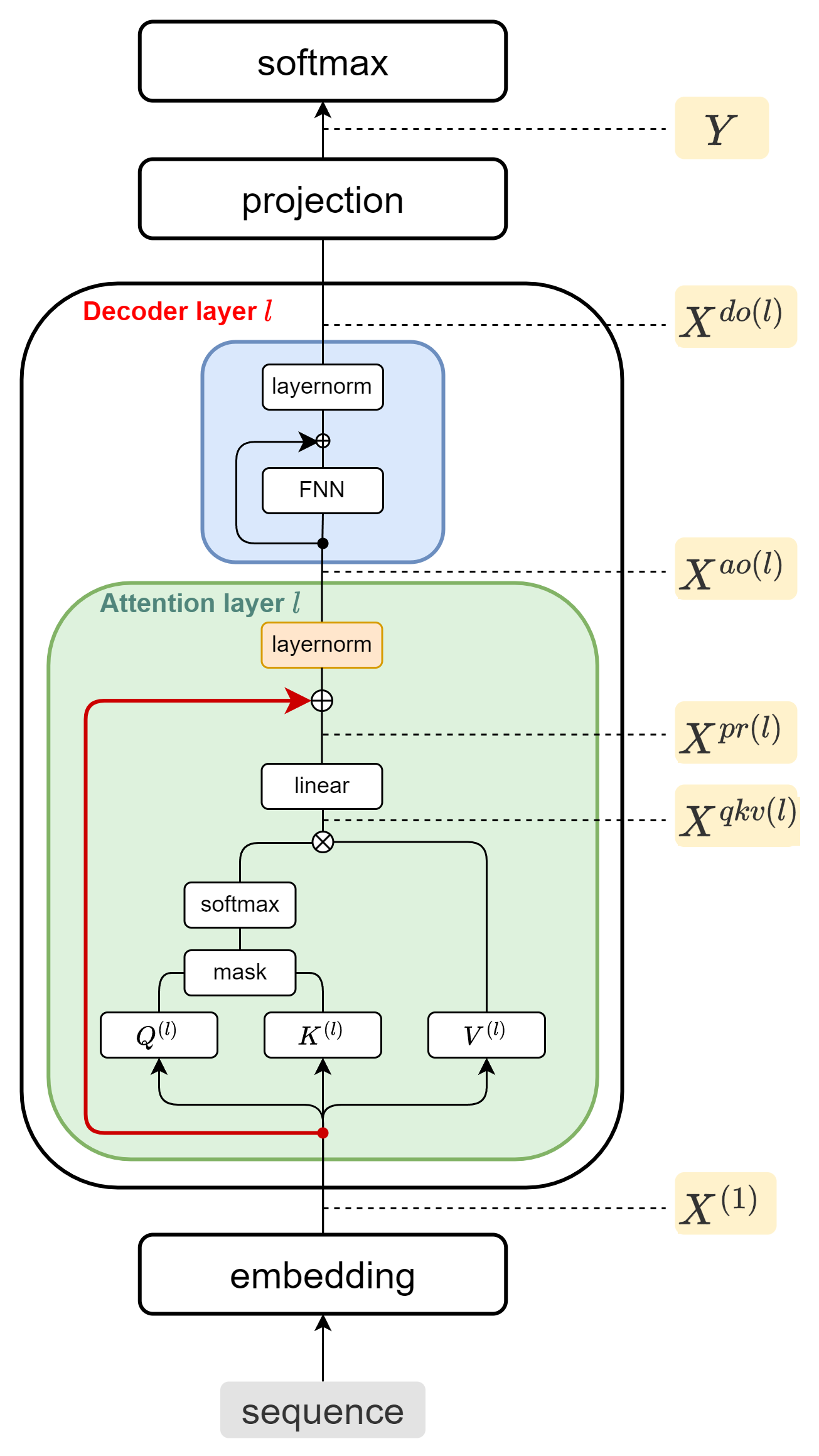}}
        \subfloat[]{\includegraphics[width=0.65\textwidth]{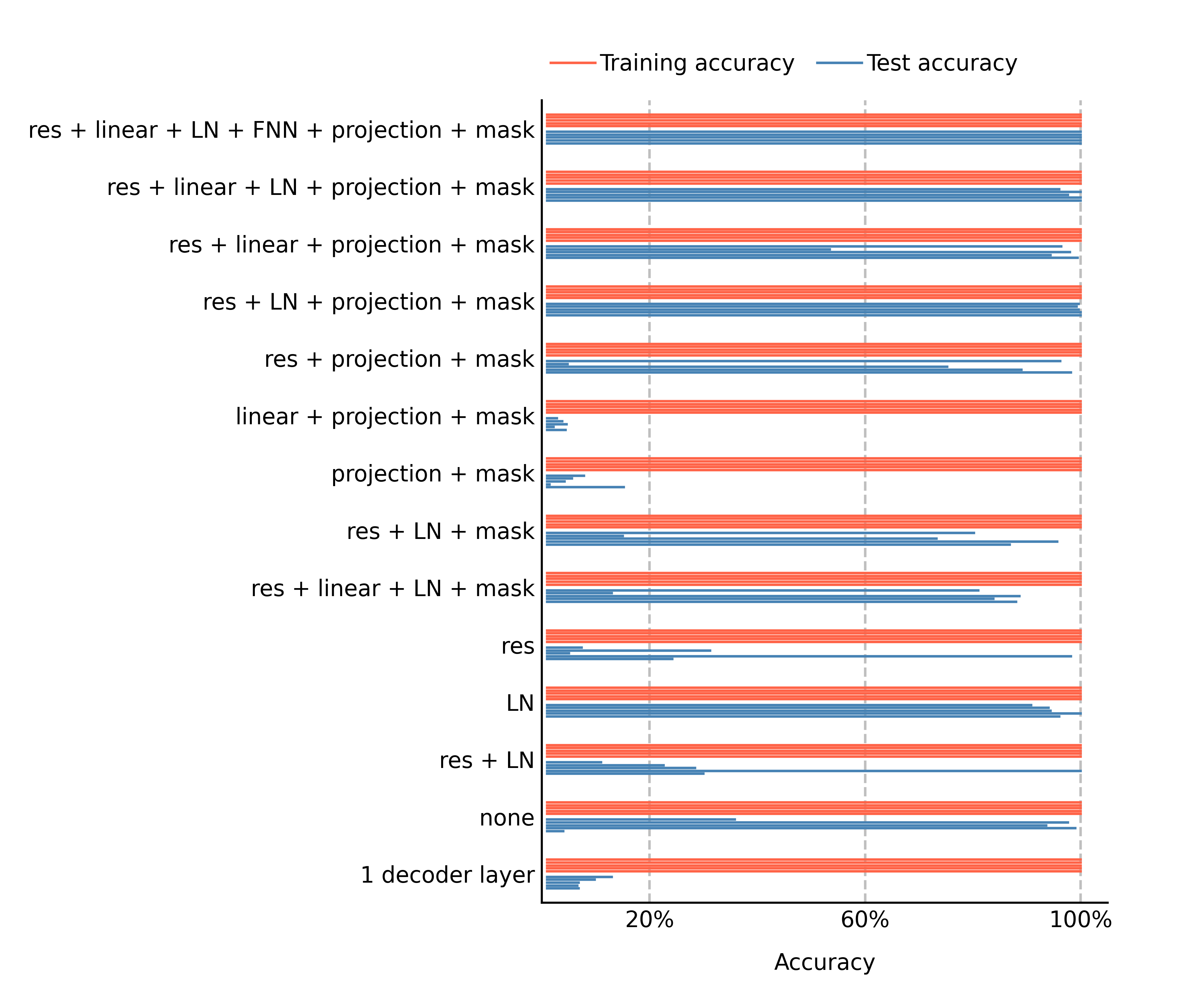}}
        \caption{(a) Transformer structure used in experiments. (b) Ablation experiments. For the first 3 and last 2 experiments, the row vector dimension of $X^{(l)}$ is 400, and the row vector dimension of $Q^{(l)}$, $K^{(l)}$, $V^{(l)}$ is 64. For others, all of those dimensions are 201.
        % $v(2,\ldots,n,1)$ means moving the first raw of $V^{(l)}$ to the last raw. 
        For each experiment setting, we run the experiment 5 times with different seeds for 4000 epochs to eliminate the impact of randomness.}
        \label{fig:ablation experiment}
    \end{figure}

For each experiment setting, we run the experiment five times with different random seeds.
As shown in Fig.~\ref{fig:ablation experiment}(b), the first experiment contains all regular components. For all five random seeds, the training accuracy and the test accuracy achieve $100\%$. Therefore, the five mini-bars together look like a large bar. In the second experiment, the FNN is removed. The training accuracy can always achieve $100\%$, but the test accuracy has slight variation across five trials. 

It is interesting to see that the effect of different components is highly nonlinear. For example, the experiment of ``linear+projection+mask'' generalizes very badly for all five trials, but when all examined components are removed, the network can generalize better, as shown by the experiment of ``none'', which can be $100\%$ although with large fluctuation. These experiments suggest there are many different mechanisms to learn an anchor function.

\subsection{A mechanism of a two-layer simplified model}

    % \begin{figure}
    %     \centering
    %     \includegraphics[width=1\linewidth]{pic/2layer_3x_to_x.png}
    %     \caption{Enter Caption}
    %     \label{fig:enter-label}
    % \end{figure}

    In this section, we employ a simplified two-layer model, which removes structures including masking, layer normalization, residual connection, linear transformation, and fully-connected networks, i.e., ``none'' in Fig. \ref{fig:ablation experiment}, to investigate the mechanisms of the identity learning task. We begin by providing the definition of the simplified model.

    Let $X^{(1)} \in \mathbb{R}^{n \times d_m}$ represent the input sequence after word embedding and position embedding, i.e., $X^{(1)}=X^{\rm em}+X^{\rm pos}$. We choose $d_m$ to be equal to the dictionary size $d$ to simplify the linear operator in the transformer and delete the projection layer. The final output is calculated as follows:
    
    % \[Q^{(1)} = X^{(1)}W^{Q(1)}, \quad K^{(1)} = X^{(1)}W^{K(1)}, \quad V^{(1)} = X^{(1)}W^{V(1)}.\]
    % \[\mathrm{Attn}^{(1)} = \text{softmax}\left(\frac{Q^{(1)}K^{(1)T}}{\sqrt{d}}\right), \quad X^{(2)} = \mathrm{Attn}^{(1)}V^{(1)}\]
    % \[Q^{(2)} = X^{(2)}W^{Q(2)}, \quad K^{(2)} = X^{(2)}W^{K(2)}, \quad V^{(2)} = X^{(2)}W^{V(2)}.\]
    % \[\mathrm{Attn}^{(2)} = \text{softmax}\left(\frac{Q^{(2)}K^{(2)T}}{\sqrt{d}}\right), \quad X^{(3)} = \mathrm{Attn}^{(2)}V^{(2)}\]

    \begin{align*}
    \mathrm{Attn}^{(1)} &= \text{softmax}\left(\frac{X^{(1)}W^{Q(1)}W^{K(1)T}X^{(1)T}}{\sqrt{d}}\right), \quad X^{(2)} = \mathrm{Attn}^{(1)}X^{(1)}W^{V(1)},\\
    \mathrm{Attn}^{(2)} &= \text{softmax}\left(\frac{X^{(2)}W^{Q(2)}W^{K(2)T}X^{(2)T}}{\sqrt{d}}\right), \quad X^{(\mathrm{out})} = \mathrm{Attn}^{(2)}X^{(2)}W^{V(2)}.
    \end{align*}
    
    It can be expressed simply as
    
    \[X^{\mathrm{(out)}} = \mathrm{Attn}^{(2)}\mathrm{Attn}^{(1)}X^{(1)}W^{V(1)}W^{V(2)}.\]

    % For the $l$-th layer, the matrices $Q^{(l)}, K^{(l)}, V^{(l)}$ of the attention mechanism are defined as:
    
    % \[Q^{(l)} = X^{(l)}W^{Q(l)}, \quad K^{(l)} = X^{(l)}W^{K(l)}, \quad V^{(l)} = X^{(l)}W^{V(l)}.\]
    
    % The attention matrix $\mathrm{Attn}^{(l)}(X^{(l)})$ for the $l$-th layer is computed as:
    
    % \[
    % \mathrm{Attn}^{(l)} = \text{softmax}\left(\frac{Q^{(l)}K^{(l)T}}{\sqrt{d}}\right).
    % \]
    
    % The input of the $l$-th layer is defined recursively as:

    % \[X^{(l)} = \mathrm{Attn}^{(l-1)}V^{(l-1)}.\]

    % The final output of the model is given by:
    
    % \[X^{\mathrm{output}} = X^{(3)} = \mathrm{Attn}^{(2)}\mathrm{Attn}^{(1)}X^{(1)}W^{V(1)}W^{V(2)}.\]

    % It is noteworthy that $\text{Attn}^{(l)}$ incorporates information from $X^{(l)}$. Therefore, the aforementioned formula is not a straightforward linear transformation. The mechanistic explanation of the model is illustrated in Fig.~\ref{fig:3x_to_x_simplified}, wherein the first attention map shifts the information of the key item to the position of anchor item ``3'' and duplicates the information from anchor item to other locations beyond anchor item. Therefore, there are only two types of matrix row vectors input to the second layer: vectors with $3$ information and vectors with $x$ information. This is also more convenient for the construction of the $\mathrm{Attn}_2$ matrix. $\mathrm{Attn}_2$ implements copying a vector with $x$ information to other locations. So in fact, the role of the two attention matrix is to move the information of $x$ to the last row, and $W^{V(1)}$, $W^{V(2)}$ acts as a classifier.

    \begin{figure}[h]
      \centering
      \includegraphics[width=0.9\textwidth]{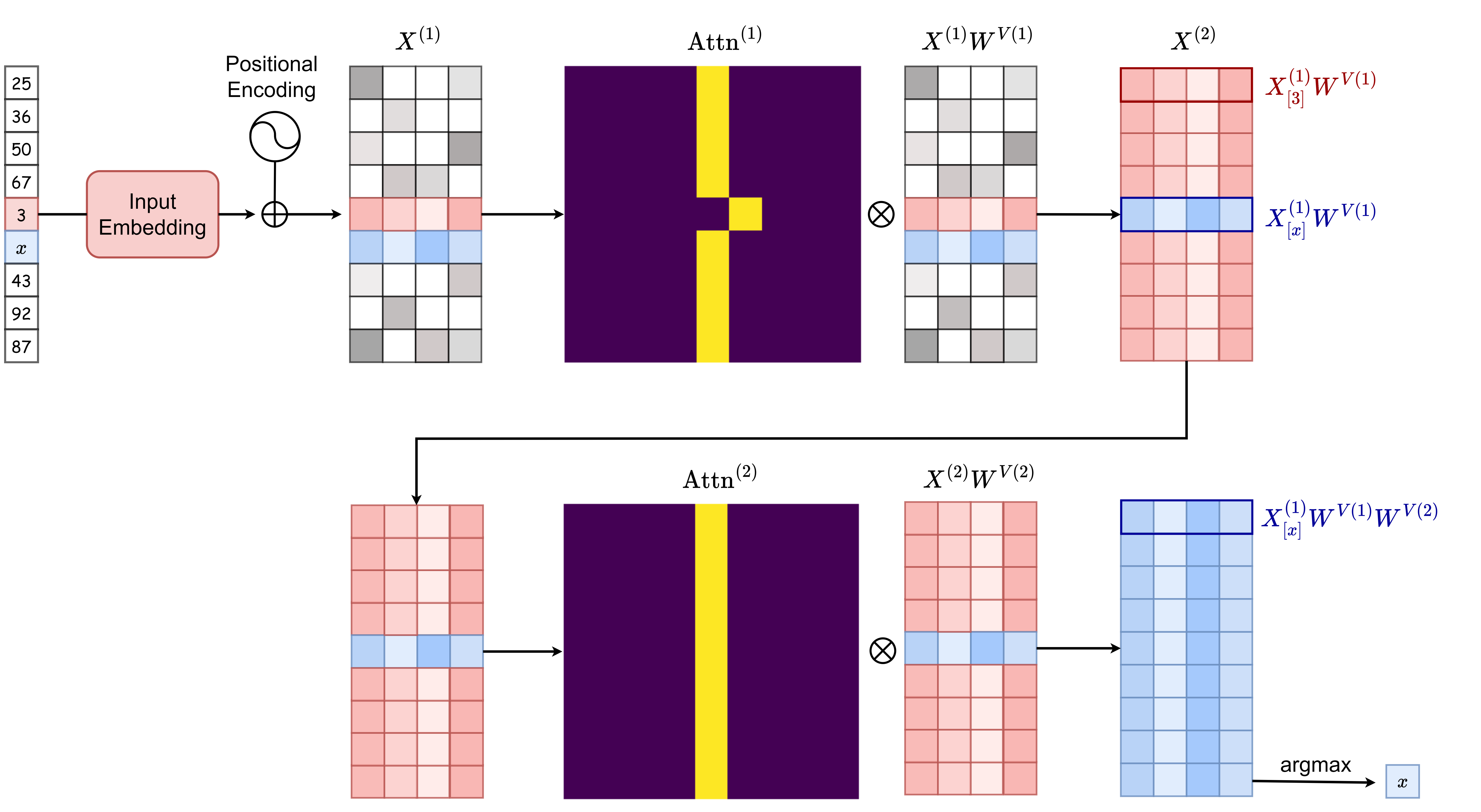}
      \caption{The mechanism of the 2-layer simplified transformer model for the identity learning task. The first attention matrix makes the output vector become $X^{(1)}_{[3]} W^{v(1)}$ or $X^{(1)}_{[x]} W^{v(1)}$, where $[3]$ and $[x]$ means the position of ``3'' and ``$x$'' in the input sequence. The second attention matrix makes all of the output vectors become $X^{(1)}_{[x]} W^{v(1)}W^{v(2)}$. $W^{v(1)}W^{v(2)}$ plays the role of classification.}
      \label{fig:3x_to_x_simplified}
    \end{figure}

    It is noteworthy that $\text{Attn}^{(l)}$ incorporates information from $X^{(l)}$. Therefore, the aforementioned formula does not represent a straightforward linear transformation. 
    The mechanistic explanation of the model is depicted in Fig.~\ref{fig:3x_to_x_simplified}, where the first-layer attention map shifts the information of the key item ``$x$'' to the position of anchor item ``3'' and broadcasts the information of the anchor item to other locations beyond it. Consequently, there are only two types of matrix row vectors input to the second layer: vectors with the anchor item information and vectors with the key item information. This structure also facilitates the construction of the second-layer attention map $\mathrm{Attn}^{(2)}$. The attention map $\mathrm{Attn}^{(2)}$ performs a broadcast operation, i.e., broadcast the vector with the key item information to all locations. 
    % Consequently, the role of the two attention matrices is to relocate the information of the key item to the last row. The matrices $W^{V(1)}$ and $W^{V(2)}$ serve the function of projection matrices, transforming the information of the key item. 
    The final prediction is obtained through the function argmax$(\cdot)$. 

    In the above-simplified model, $\mathrm{Attn}^{(1)}$ plays an important role in information transmission, which can be regarded as the combination of shift attention and broadcast attention.

\subsection{Shift and broadcast in Llama2-6B}
\begin{figure}
    \centering
    \subfloat[the 23rd head of the second layer]{\includegraphics[height=0.35\linewidth]{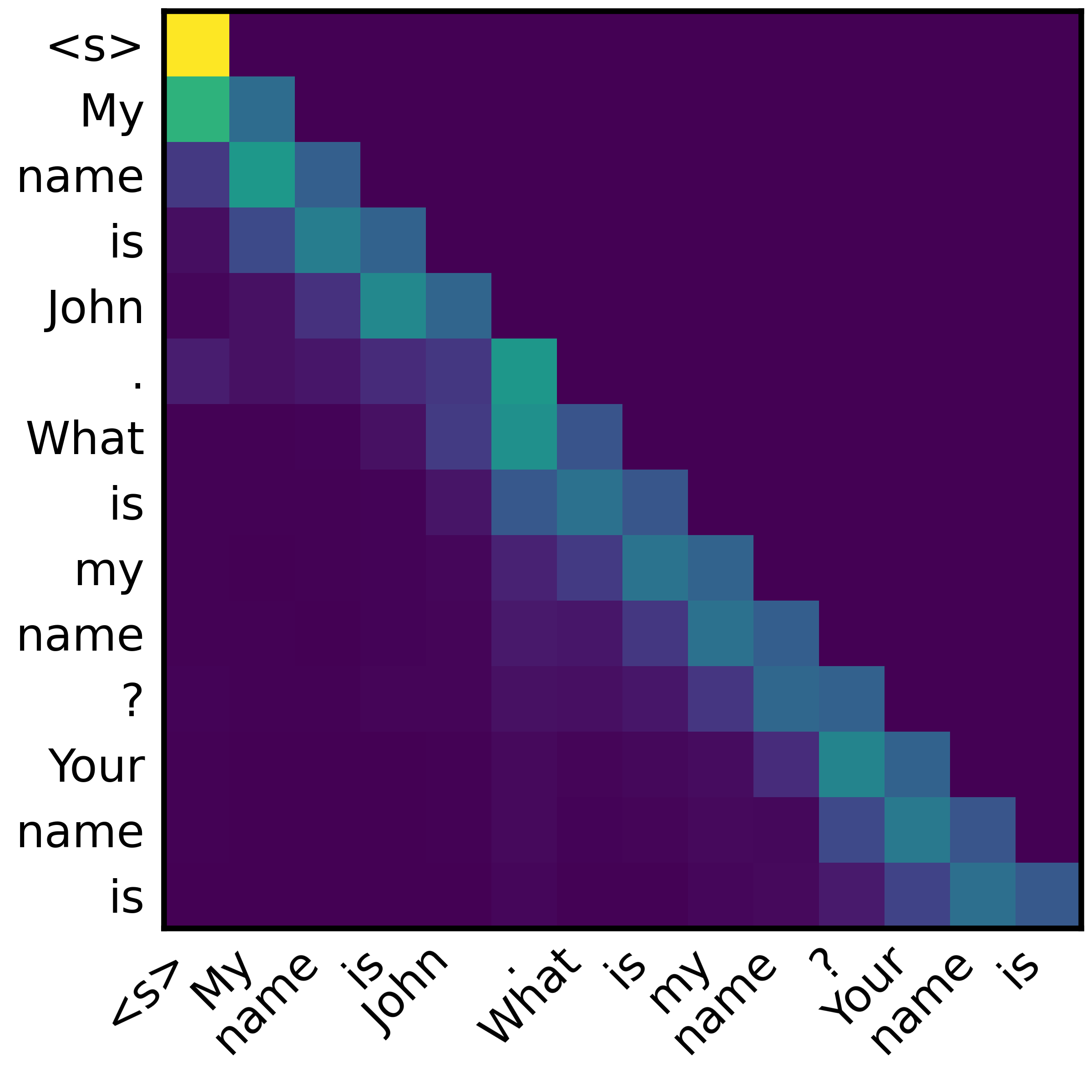}}
    \hspace{.4in} 
    \subfloat[the 17th head of the 29th layer]{\includegraphics[height=0.35\linewidth]{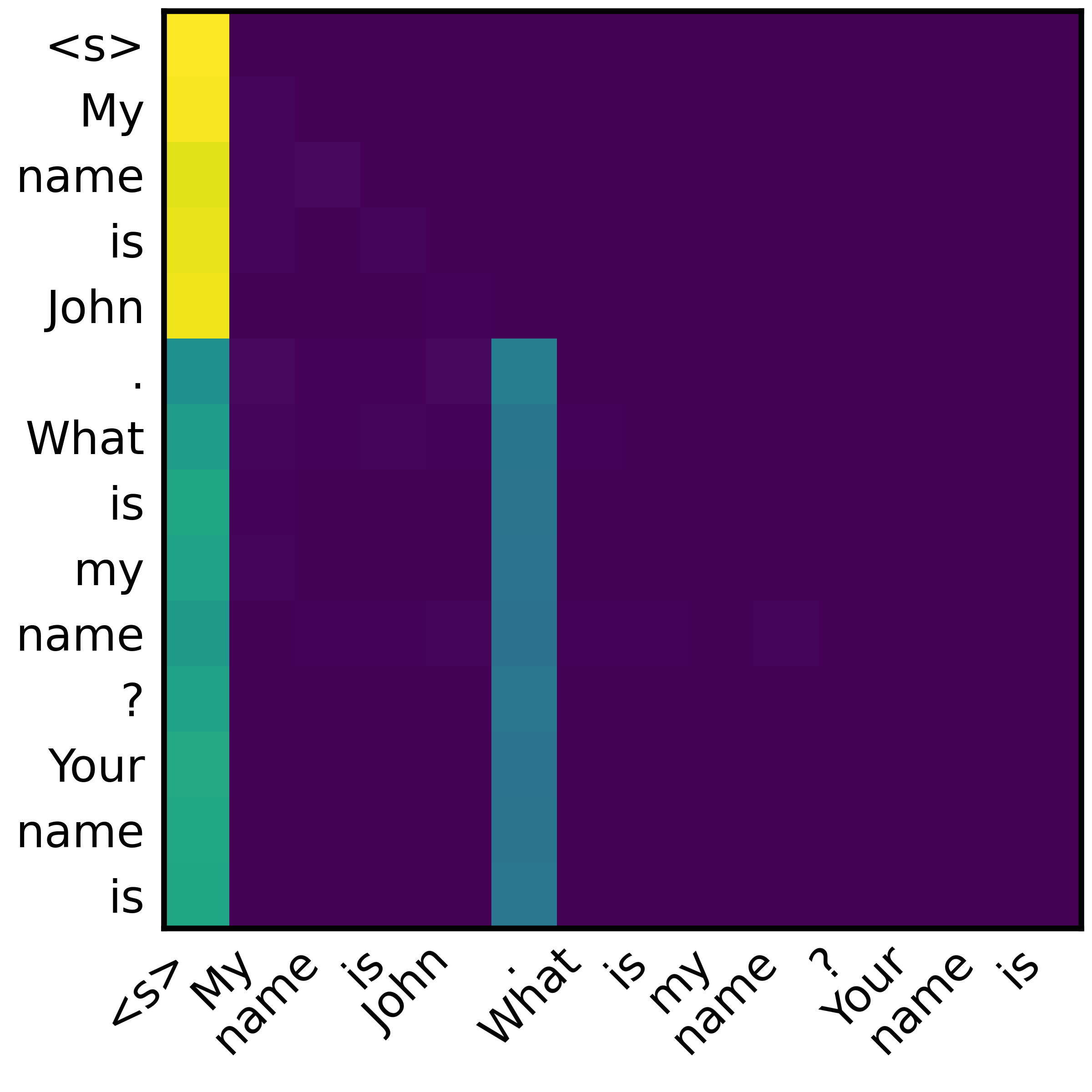}}
    % \hspace{.3in} 
    % \subfloat[the cosine similarity]{\includegraphics[height=0.26\linewidth]{pic/Llama_output_cos_sim.png}}
    \caption{Attention examples of Llama2-6B.}
    \label{fig:Llama2-6B}
\end{figure}

We find that the operations of shift and broadcast by attention, which are important in learning anchor functions, are also very common in LLMs. We use Llama2-6B \cite{touvron2023llama2} as an example to demonstrate these two operations.
Similar to the identity learning task,  
% Due to variations in training data complexity and differences in training strategies, investigating the generality of shift and duplicate mechanisms in pre-training large language models is worthwhile. In this section, we explore the prevalence of shift and duplicate operations in Llama2-6B \cite{touvron2023llama2}. According to the original intention of constructing the identity learning task, 
we employ ``My name is John. What is my name? Your name is'' as the input for Llama2-6B and observe the attention maps at different layers. Fig. \ref{fig:Llama2-6B}(a, b) illustrates two heads in Llama2-6B. The left one, a head in the first layer, exhibits noticeable shift operations, and the second one, a head in the twenty-eighth layer, demonstrates clear broadcast operations. Multiple attention maps for various heads are supplied in the appendix. Both operations are rather common.

It is noteworthy that shallow-layer attentions tend to favor shift operations, while deep-layer attentions tend to favor broadcast operations. The full-size attention map can be found in the appendix~\ref{sec: Attention maps of Llama2-6B}. Additionally, the highly similar broadcast attention patterns in deep layers explain the phenomenon where performance remains almost unaffected upon the removal of several consecutive layers that are near the final output.

% Although in Llama2-6B, the outputs of different items are different, there is a high cosine similarity between different outputs, so the broadcast operation is also established. As shown in Fig. \ref{fig:Llama2-6B}(c), we show the cosine similarity between model outputs corresponding to different items in the sequence. The corresponding output dimension of an item is 32,000 dimensions, and the cosine similarity between two random vectors of the same dimension is much smaller than the cosine similarity between the outputs of two items.

% <s> My name is John. What is my name? Your name is
% Unterscheidung name is J and I is your name? \n name is John

% In Llama2-6B, despite the distinct outputs for different items, there exists a notable cosine similarity between various outputs, thereby justifying the broadcast operation. As depicted in Fig. \ref{fig:Llama2-6B}(c), we present the absolute value of cosine similarity among model outputs corresponding to different items in the above sequence. The output dimension for each item is 32,000 dimensions, and the cosine similarity between two randomly generated vectors of the same dimension is significantly smaller than the cosine similarity between the outputs of two distinct items.

\subsection{Discussion of shift and broadcast operations}
Obviously, the attention structure can perform various operations in various learning tasks. What kind of operations the attention structure shows depends on the task and also the input sequence. For the identity learning anchor function, the distance between the anchor and the key item is fixed. Then, the attention structure learns this position relation and performs a shift operation. Such a shift can always fuse the information of the anchor and the key item for any input sequence.  

The broadcast operation can simplify the network, such as ignoring irrelevant tokens, providing working memory for intermediate results, making extra deep layers negligible, etc. From the perspective of complexity, broadcast operations enable neural networks to accomplish tasks using a function with low complexity.

The shift and broadcast are only two basic operations among many others. For example, the induction head correlates the current token with similar ones in the previous context \cite{elhage2021transformer,olsson2022context}. The future study of various anchor functions will reveal more and more basic operations of attention structure as well as other components.

\section{Existing theoretical understanding for anchor function}
In this section, we utilize some existing theoretical understanding for learning anchor functions, including frequency principle \cite{xu2019training,rahaman2018spectral,xu2019frequency,basri2019convergence,cao2019towards,zhang2021linear,luo2019theory,xu2022overview}  and condensation \cite{luo2021phase,zhou2021towards,zhou2022empirical,zhou2023understanding,chen2023phase}.

\subsection{Frequency principle}
The frequency principle indicates an implicit spectral bias during the learning process, i.e., neural networks often learn from low- to high-frequency. 

Frequency is used to indicate how much change of the output is affected by the change of the input. A larger change in the output corresponds to a higher frequency. We design two tasks that have different frequencies. In each task, the anchor set $A=\{3,4\}$, and in each input sequence, there exists only one anchor. Denote the key item following the anchor as $x$.

For task one, the output is always the item $x$. For example, the output of (12,33,14,3,42,22,32,20,28) is 42, and the output of (12,33,14,3,42,22,32,20,28) is also 42.

For task two, for anchor ``3'', the output is $x$; for anchor ``4'', the output is $x+1$. For example, the output of (12,33,14,3,42,22,32,20,28) is 42, and the output of (12,33,14,3,42,22,32,20,28) is 43.

\begin{figure}
    \centering
    \includegraphics[width=0.8\linewidth]{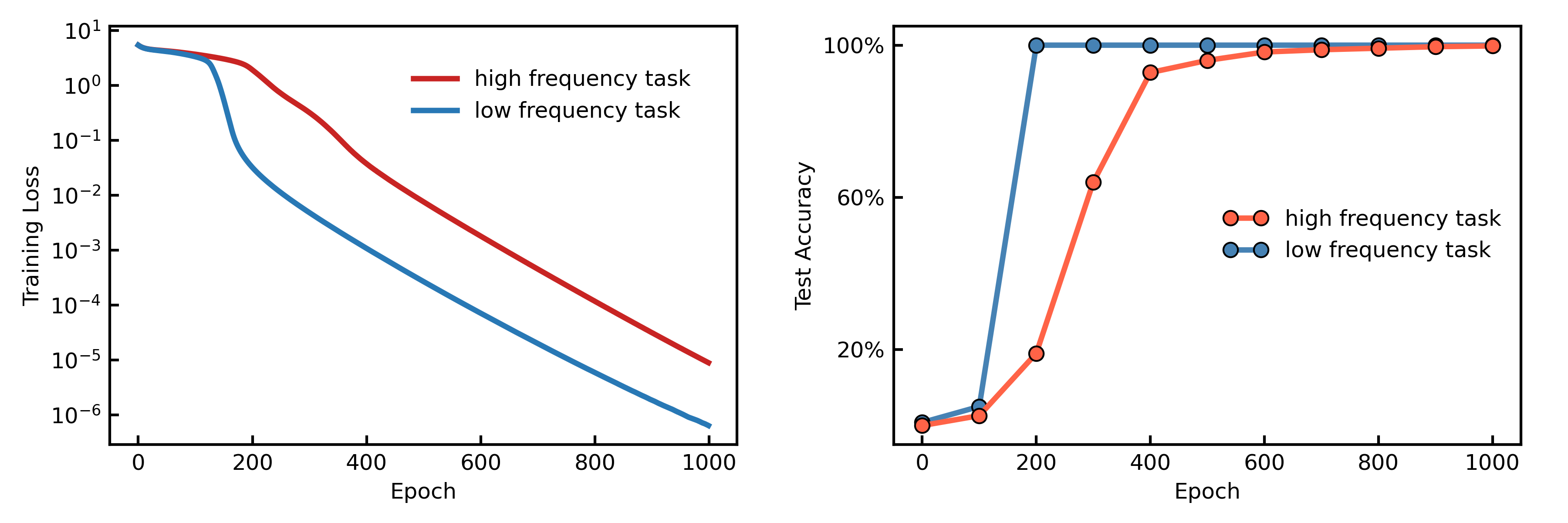}
    \caption{Training loss and test accuracy for two frequency tasks.}
    \label{fig:frequency}
\end{figure}

Obviously, task two has a higher frequency. According to the frequency principle, the learning of task two should be slower. As shown in Fig. \ref{fig:frequency}, for both the training loss and the test accuracy, the fitting speed of the low-frequency task is faster than that of the high-frequency task.

\subsection{Condensation}
Previous works have found that, as the initialization of network weights becomes smaller, the neurons in the same layer would tend to be more condensed during the training, i.e., weights of neurons in the same layer tend to be more similar. Condensation implies that neural networks tend to fit the data with a function with complexity as small as possible. A key reason underlying the condensation is that the neuron in the same layer has similar training dynamics.

\begin{figure}
    \centering
    \includegraphics[width=0.9\linewidth]{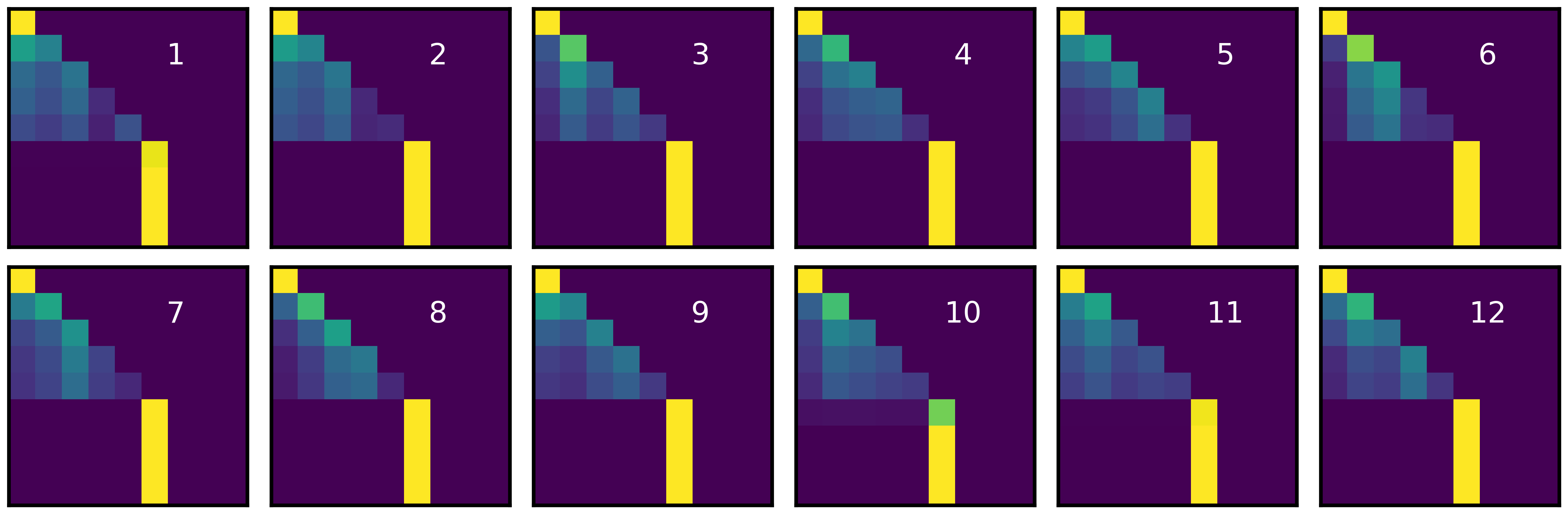}
    \caption{The attention map of each head in the second layer of the 4-layer 12-head model.}
    \label{fig:condensation}
\end{figure}

For a transformer network with multiple heads in the same layer, the heads also have similar dynamics. We should expect that heads in the same layer can have many similarities during or after the training. As shown in Fig. \ref{fig:condensation}, we show the attention map of each head in the second layer of a 4-layer 12-head model. The attention maps between different heads are highly similar. They all perform the broadcast operation.

\section{Future problems}
The anchor function has opened up a new avenue for studying language models. The study of the anchor function in the main text, along with the detailed analysis in the Appendix, has opened up a new avenue for studying language models. There are many more questions worth further exploration, such as some questions listed in Table \ref{tab:more questions}. An exciting thing is that readers can easily cook up more language tasks with anchor functions and propose a series of interesting questions for further exploration. With the empirical discovery based on the anchor function, a series of theoretical should also be anticipated.

\begin{longtable}
{>{\bfseries\arraybackslash}m{0.2\textwidth}|m{0.72\textwidth}} 
% \label{tab:more questions}\\
\caption{Example questions for further exploration based on anchor function.}\label{tab:more questions} \\
\hline
\rowcolor{gray!50}
\textbf{Category} & \textbf{Question} \\ \hline
\endfirsthead % 从第二页开始的表头

\caption{Classification of Neural Network Mechanism Questions (Continued)} \\
\hline
\rowcolor{gray!50}
\textbf{Category} & \textbf{Question} \\ \hline
\endhead % 所有其他页的表头

\rowcolors{2}{gray!25}{white} % 从第二行开始交替使用浅灰色和白色背景
% 表格内容
\multirow{4}{0.2\textwidth}{Structure characteristic} 
 & \cellcolor{gray!20}Why transformer network is better than the fully-connected network on anchor functions? \\
 & Why transformer network can generalize better than LSTM network on anchor functions with small data size? \\
 & \cellcolor{gray!20}What is the effect of each structure component? such as attention, layer normalization, softmax, skip connection, and fully-connected one? \\ 
 & Why the following can happen: when some components are removed, the generalization degrades, but when some more components are removed, the generalization improves.\\ \hline
 
\multirow{2}{0.2\textwidth}{Approximation} 
 & \cellcolor{gray!20}What is the benefit of multiple heads in approximation? \\
 & What is the benefit of multiple layers in approximation? \\ \hline
 
\multirow{15}{0.2\textwidth}{Generalization \newline\& \newline Optimization}
 & \cellcolor{gray!20}Why transformer networks can generalize over unseen composite tasks? \\
 & How does the transformer network learn synonym anchors? \\ 
% \multirow{13}{0.3\textwidth}{Optimization} 
 & \cellcolor{gray!20}What is the benefit of multiple heads in training? \\
 & What is the benefit of multiple layers in training? \\
 & \cellcolor{gray!20}How does the landscape differ over the increasing number of heads? \\
 & How does the landscape differ over the increasing number of layers? \\
 & \cellcolor{gray!20}Why does the emergence of generalization over unseen composite tasks later than that over unseen data of seen tasks? \\
 & Why learning rate is so sensitive?  \\
 & \cellcolor{gray!20}Why do loss spikes sometimes help but not always? \\
 & Why do many heads learn the same attention, is it similar to the condensation phenomenon? \\
 & \cellcolor{gray!20}Why forward recitation task is easier than the backward one?\\
 & Is there a learning order for different layers, and why?\\
 & \cellcolor{gray!20}How does the hidden space dimension affect learning?\\
 & Why the pattern of the output sequence is simpler when the data size is larger?\\
 & \cellcolor{gray!20}Explore more basic operations of the attention structure in various tasks. \\ \hline
 \multirow{1}{0.2\textwidth}{More} 
 & Build more benchmark functions. \\ \hline
\end{longtable}

\bibliographystyle{alpha}
\bibliography{sample}

\newpage

\appendix

\section{Basic settings}
    
    \noindent\textbf{Transformer} We use a 4-layer decoder-only transformer network to learn various tasks. Each decoder layer has 4 heads. The optimizer chosen is AdamW. The batch size is set to 100. The initial learning rate is 2e-5. In the first 400 epochs, the warmup learning rate update strategy is used to gradually increase the learning rate to 2e-4, and then the cosine annealing strategy is used to decay the learning rate back to 2e-5. A total of 4000 epochs are trained.

    \noindent\textbf{LSTM}  We use a 4-layer LSTM to learn various tasks. The hidden size of the model is 4. The optimizer chosen is AdamW. The batch size is set to 10. The initial learning rate is 2e-5. In the first 400 epochs, the warmup learning rate update strategy is used to gradually increase the learning rate to 2e-4, and then the cosine annealing strategy is used to decay the learning rate back to 2e-5. A total of 4000 epochs are trained.

    \noindent\textbf{DNN} We use a 4-layer DNN to learn various tasks. The hidden size of the model is 4. The optimizer chosen is AdamW. The batch size is set to 10. The initial learning rate is 2e-5. In the first 400 epochs, the warmup learning rate update strategy is used to gradually increase the learning rate to 2e-4, and then the cosine annealing strategy is used to decay the learning rate back to 2e-5. A total of 4000 epochs are trained.

\section{Two-layer transformer}

\begin{figure}
    \centering
    \includegraphics[width=1\linewidth]{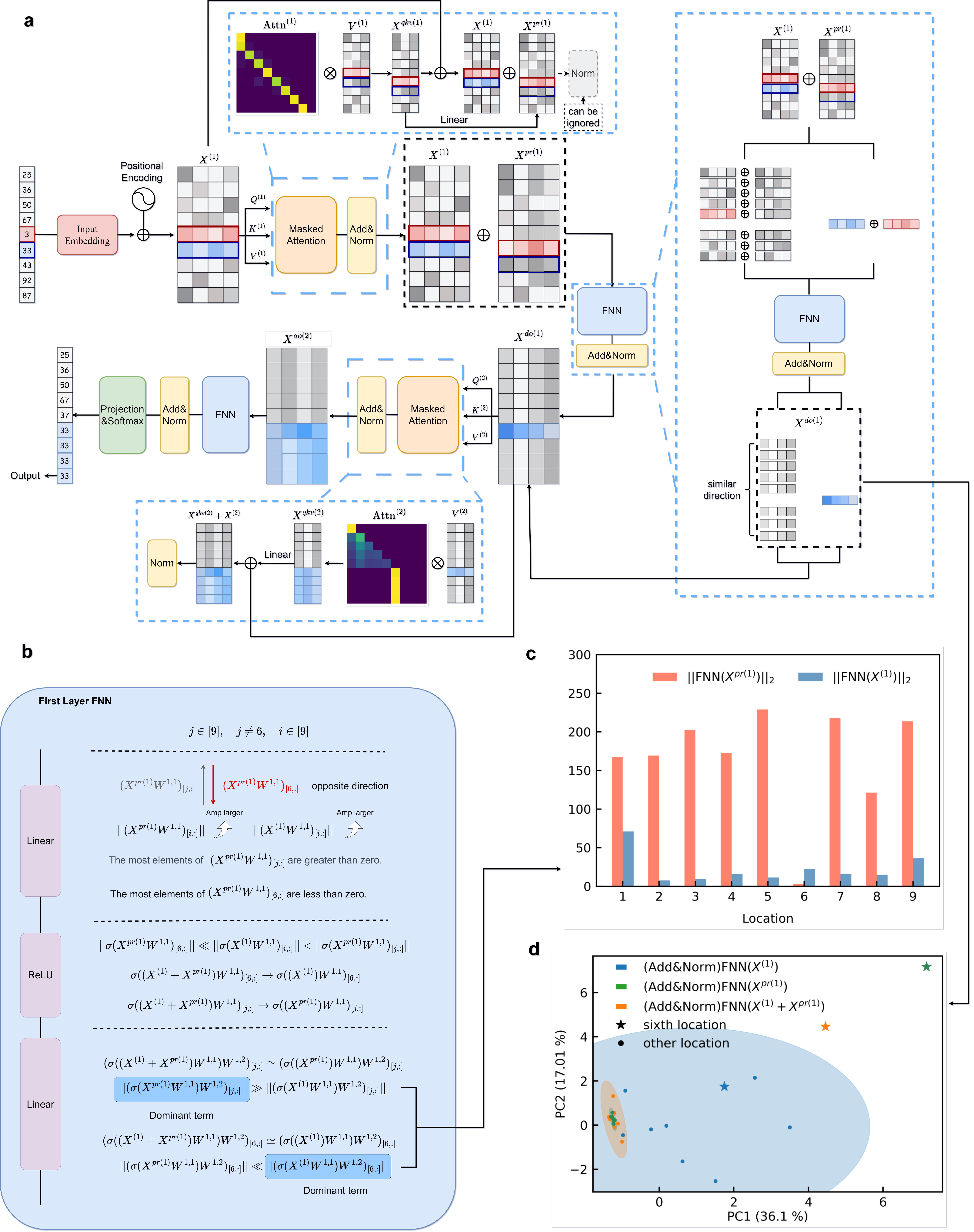}
    \caption{Detailed mechanism of a two-layer transformer. (a) Flow chart of two-layer transformer. (b) The detailed internal mechanism of the first layer FNN. (c) The norm of $\mathrm{FNN}(X^{(1)})$ and $\mathrm{FNN}(X^{\mathrm{pr}(1)})$ with respect to different tokens. (d) PCA Analysis of three kinds of FNN outputs, i.e., $\text{(Add\& Norm)FNN}(X^{(1)})$, $\text{(Add\& Norm)FNN}(X^{\text{pr(1)}})$ and $\text{(Add\& Norm)FNN}(X^{(1)}+X^{\text{pr(1)}})$. ``(Add\& Norm)'' indicates that output undergoes residual connection and layer normalization.  }
    \label{fig:detailed_machanism}
\end{figure}

In this section, we study the mechanism of two-layer transformers to achieve the identity task. We first give a detailed notation of the transformer structure, and then select three modules that have a greater contribution to this task, namely the first-layer attention module, the first-layer FNN module and the second-layer attention module to analyze its mechanism, shown in Fig. \ref{fig:detailed_machanism}.

\subsection{Transformer architecture details}\label{notation}
In this section, we give the notation of the transformer in modules to facilitate subsequent analysis of its mechanism.

\subsubsection{Input representation}

The input sequence is represented as a one-hot vector $X^{\mathrm{in}} \in \mathbb{R}^{n \times d}$, where $n$ is the sequence length, $d$ is the dictionary size, and $X^{\mathrm{in}}$ is the one-hot vector.

After embedding, the input sequence becomes:

\[
X^{\mathrm{em}} = X^{\mathrm{in}}W^{\mathrm{em}} \in \mathbb{R}^{n \times d_m}, \quad W^{\mathrm{em}} \in \mathbb{R}^{d \times d_m}.
\]

The positional vector is denoted as:

\[
X^{\mathrm{pos}} \in \mathbb{R}^{n \times d_m}.
\]

The input to the first transformer block is:

\[
X^{(1)} = X^{\mathrm{em}} + X^{\mathrm{pos}}.
\]

\subsubsection{Attention mechanism}

For the $l$-th layer, the $Q, K, V$ matrices of the attention mechanism are defined as functions of the input $X^{(l)}\in \mathbb{R}^{n \times d_{m}}$:

\[
Q^{(l)}(X^{(l)}) = X^{(l)}W^{Q(l)},  \quad W^{Q(l)} \in \mathbb{R}^{d_m \times d_k}
\]

\[
K^{(l)}(X^{(l)}) = X^{(l)}W^{K(l)},  \quad W^{K(l)} \in \mathbb{R}^{d_m \times d_k}
\]

\[
V^{(l)}(X^{(l)}) = X^{(l)}W^{V(l)},  \quad W^{V(l)} \in \mathbb{R}^{d_m \times d_k}
\]

The attention matrix $\mathrm{Attn}^{(l)}(X^{(l)})$ for the $l$-th layer is computed as:

\[
\mathrm{Attn}^{(l)}(X^{(l)}) = \text{softmax}\left(\frac{Q^{(l)}(X^{(l)})K^{(l)}(X^{(l)})^T}{\sqrt{d_k}}\right)\in\mathbb{R}^{n \times n}, 
\]
where the function $\text{softmax}(\cdot)$ is defined as:

\[ \text{Softmax}(\vx)_i = \frac{e^{x_i}}{\sum_{j=1}^{k} e^{x_j}}, \]

The output of the attention mechanism, denoted as $X^{\mathrm{qkv}(l)}$, is given by:

\[
X^{\mathrm{qkv}(l)} = \mathrm{Attn}^{(l)}(X^{(l)}) V^{(l)}(X^{(l)})\in \mathbb{R}^{n \times d_k}
\]

Finally, the output after position-wise feedforward processing is obtained using residual connection and layer normalization:

\[
X^{\mathrm{pr}(l)} = X^{\mathrm{qkv}(l)}W^{\mathrm{attn}, l}, \quad W^{\mathrm{attn},l} \in \mathbb{R}^{d_k \times d_m}, \quad X^{\mathrm{pr}(l)} \in \mathbb{R}^{n \times d_m}
\]

The output of the $l$-th layer, $X^{\mathrm{ao}(l)}$, is computed as:

\[
X^{\mathrm{ao}(l)} = \text{LayerNorm}(X^{(l)} + X^{\mathrm{pr}(l)}).
\]

\subsubsection{Feedforward neural network (FNN)}

The $l$-th layer Feedforward Neural Network (FNN) is expressed as:

\[
X^{\mathrm{do}(l)} := \text{FNN}(X^{\mathrm{ao}(l)}) = \text{LayerNorm}\left(X^{\mathrm{ao}(l)} + \sigma(X^{\mathrm{ao}(l)}W^{l, 1})W^{l, 2}\right),
\]

where $\sigma(\cdot) = \text{ReLU}(\cdot)$, and $W^{l, 1} \in \mathbb{R}^{d_m \times 3d_m}$, $W^{l, 2} \in \mathbb{R}^{3d_m \times d_m}$.

\subsubsection{Projection layer}

The projection layer is defined as:

\[
Y = X^{\mathrm{do}(L)}W^{\mathrm{proj}} + b^{\mathrm{proj}}, \quad W^{\mathrm{proj}} \in \mathbb{R}^{d_m \times N}, \quad b^{\mathrm{proj}} \in \mathbb{R}^{N}, \quad Y \in \mathbb{R}^{n \times N},
\]

where $n = 9$, $d_{m} = 400$, $L = 2$, $d_k = 64$. The output is obtained by taking the argmax of the softmax:

\[
\text{Output} = \text{argmax}(\text{softmax}(Y)).
\]

Here, the anchor ``3'' is positioned at the 5-th position, and the target position ``x'' is at the 6-th position in the sequence. For clarity, vector positions in the sequence may be referenced in subsequent discussions.

\subsection{Analysis of attention matrix $\mathrm{Attn}^{(1)}$}

The structure of the attention matrix $\mathrm{Attn}^{(1)}$ is mainly determined by the positional encoding vector $X^{\mathrm{pos}}$, shown in Fig.~\ref{fig:attn1}(a, b, c). To gain insights into the structure of the attention matrix, we examine the cosine similarity between $Q^{(1)}(X^{\mathrm{pos}})$ and $K^{(1)}(X^{\mathrm{pos}})$, as shown in the following Fig.~\ref{fig:attn1}(d), and find that the direction of $Q^{(1)}(X^{\mathrm{pos}})_i$ and $K^{(1)}(X^{\mathrm{pos}})_{i+1}$ is almost the same.  It is noteworthy that in the first layer's attention mechanism, the prediction results remain correct if we artificially replace the attention matrix with:

\begin{figure}[ht]\centering
    \subfloat[$\text{Attn}^{(1)}(X^{(1)})$]{\includegraphics[width=0.22\textwidth]{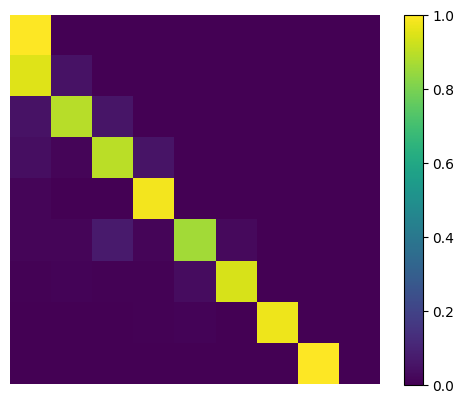}}
    \subfloat[$\text{Attn}^{(1)}(X^{em})$]{\includegraphics[width=0.22\textwidth]{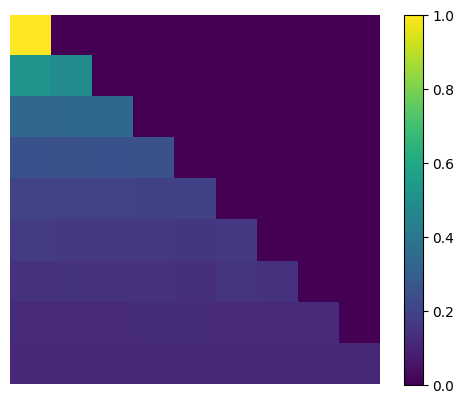}}
    \subfloat[$\text{Attn}^{(1)}(X^{\mathrm{pos}})$]{\includegraphics[width=0.22\textwidth]{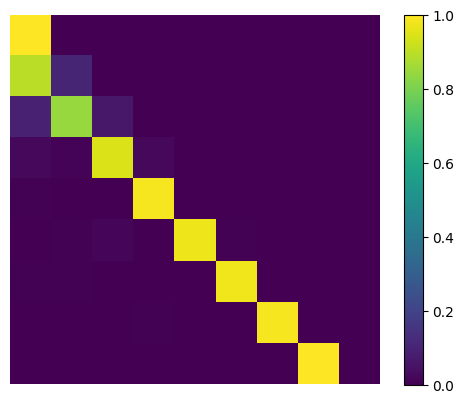}}
    \subfloat[cosine similarity]{\includegraphics[width=0.23\textwidth]{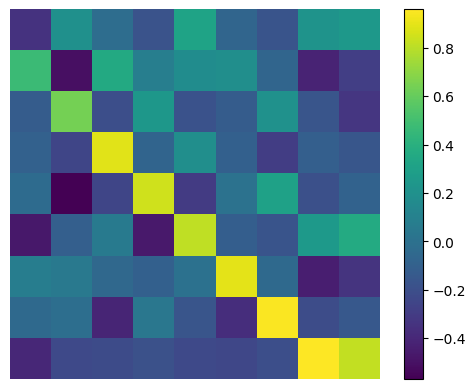}}
    \caption{(a, b, c) Attention output of the first layer with different input $X^{(1)}$, $X^{em}$ and $X^{\mathrm{pos}}$. (d) The cosine similarity between $Q^{(1)}(X^{\mathrm{pos}})$ and $K^{(1)}(X^{\mathrm{pos}})$.}
    \label{fig:attn1}
\end{figure}

% Add matrices and details here, or continue with the next section.

% To gain insights into the structure of the attention matrix, we examine the cosine similarity between $Q^{(1)}(X^{\mathrm{pos}})$ and $K^{(1)}(X^{\mathrm{pos}})$, as shown in the following figure. The analysis aims to provide an intuitive understanding of how $W^{Q(1)}$ and $W^{K(1)}$ map the positional encoding vectors $X^{\mathrm{pos}}_{[i, :]}$ and $X^{\mathrm{pos}}_{[i+1, :]}$ to vectors with similar directions, resulting in higher element-wise products.

% It is noteworthy that in the first layer's attention mechanism, if we artificially replace the attention matrix with:

\[
\text{Attn}^{(1)}(X^{(1)}) = \begin{bmatrix} 0\\1&0\\0&1&0\\\vdots&\vdots&\vdots&\\0&0&0&\ldots&1&0&&\\0&0&0&\ldots&0&1&0&\end{bmatrix}.
\]

% the prediction results remain correct. 
This suggests that the fine structure of attention in the first layer does not significantly affect the outcomes. Additionally, the layer normalization following the attention mechanism does not impact the results. Therefore, the main purpose of $\mathrm{Attn}^{(1)}$ is to complete the shift of the vector and combine it with the subsequent residual connection to fuse the key item information with the anchor information.

% This observation forms the basis for separating the two components of the residual connection in our Feedforward Neural Network (FNN) layer.

% Insert figure and description of the cosine similarity analysis here.

\subsection{Analysis of feedforward neural network (FNN)}

The FNN tends to align the vectors excluding the sixth position, i.e., $X^{\mathrm{do}(1)}_i$, $i \neq 6$, in parallel while making the vector at the sixth position, $X^{\mathrm{do}(1)}_6$, as orthogonal as possible to the other vectors. 

This alignment and orthogonality are primarily influenced by the output $X^{\mathrm{pr}(1)}$ from the first layer's attention mechanism. In other words, the ResNet operation applied to $X^{(1)}$ in the first layer of FNN alters the direction of the input vector $X^{(1)}$ to some extent, resulting in pairwise parallelism and orthogonality in $X^{\mathrm{do}(1)}$. It is worth noting that this similarity and orthogonality are not immediately apparent in $X^{\mathrm{ao}(l)}$, even though $X^{\mathrm{ao}(l)}$ already incorporates information from $X^{\mathrm{pr}(1)}$. These properties become more pronounced after passing through the FNN. 

The following is the experimentally verification of these observations. Experimental results suggest that the layer normalization of the first layer's attention has minimal impact on the results, thus we assume for convenience that 
$$X^{\mathrm{ao}(1)} := \text{LayerNorm}(X^{(1)} + X^{\mathrm{pr}(1)}) = X^{(1)} + X^{\mathrm{pr}(1)}.$$
We examine the cosine similarity between vectors at different positions and stages of the FNN. The FNN is divided into six stages: Input (the FNN's input vector), First linear output, ReLU output, Second linear output, ResNet output, and Layernorm output.

% $X^{(1)}:$

% \begin{figure}[h]
%   \centering
%   \includegraphics[width=0.6\textwidth]{pic/two-layer_network/Untitled9.png}
%   \caption{Cosine Similarity Analysis for $X^{(1)}$ in different FNN stages.}
%   \label{fig:cosine_similarity_X1}
% \end{figure}

% $X^{\mathrm{pr}(1)}:$

\begin{figure}[h]
  \centering
\includegraphics[width=0.6\textwidth]{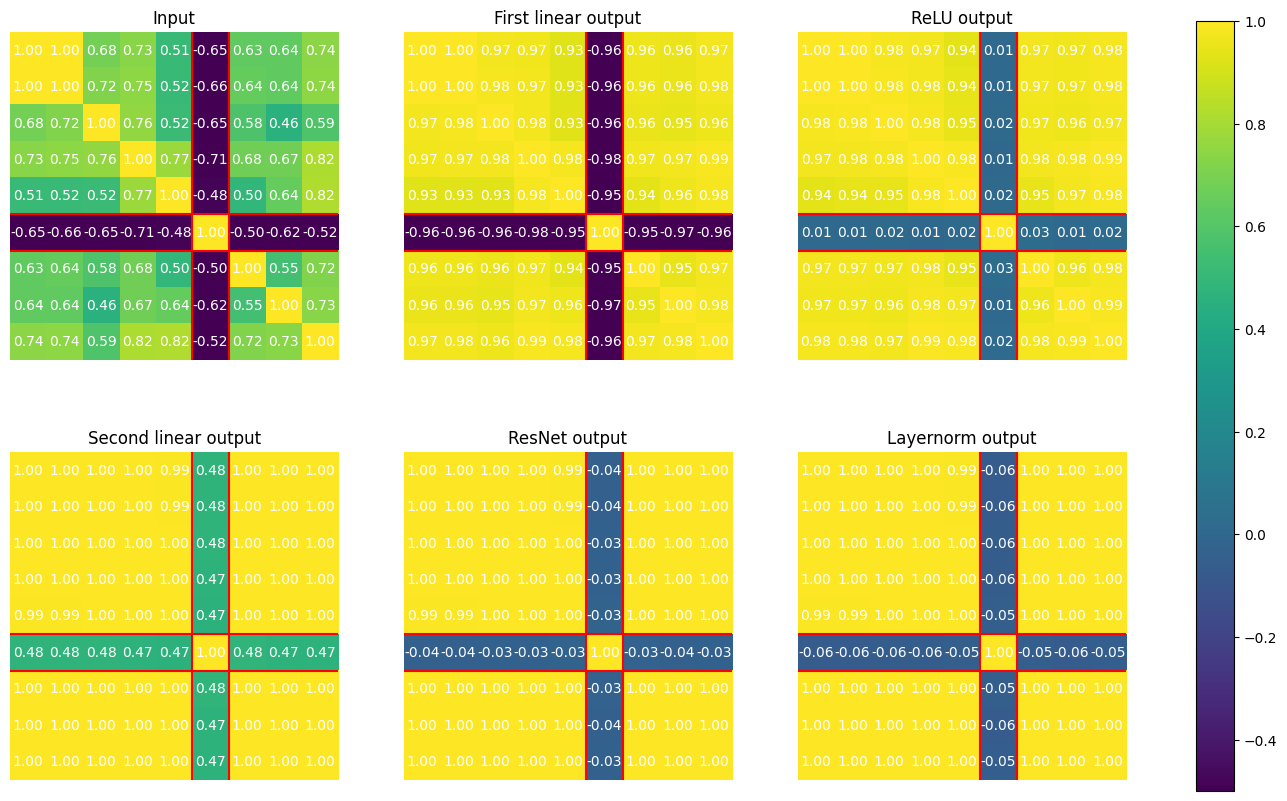}
  \caption{Cosine Similarity Analysis for the output in different first-layer FNN stages with input $X^{\mathrm{pr}(1)}$. Each subfigure represents in turn: the input vector, the output of the first linear layer, the output after the activation function, the output of the second linear layer, the output after residual connection, and the output after layer normalization.}
  \label{fig:cosine_similarity_Xpr1}
\end{figure}

% $X^{(1)} + X^{\mathrm{pr}(1)}:$

\begin{figure}[h]
  \centering
\includegraphics[width=0.6\textwidth]{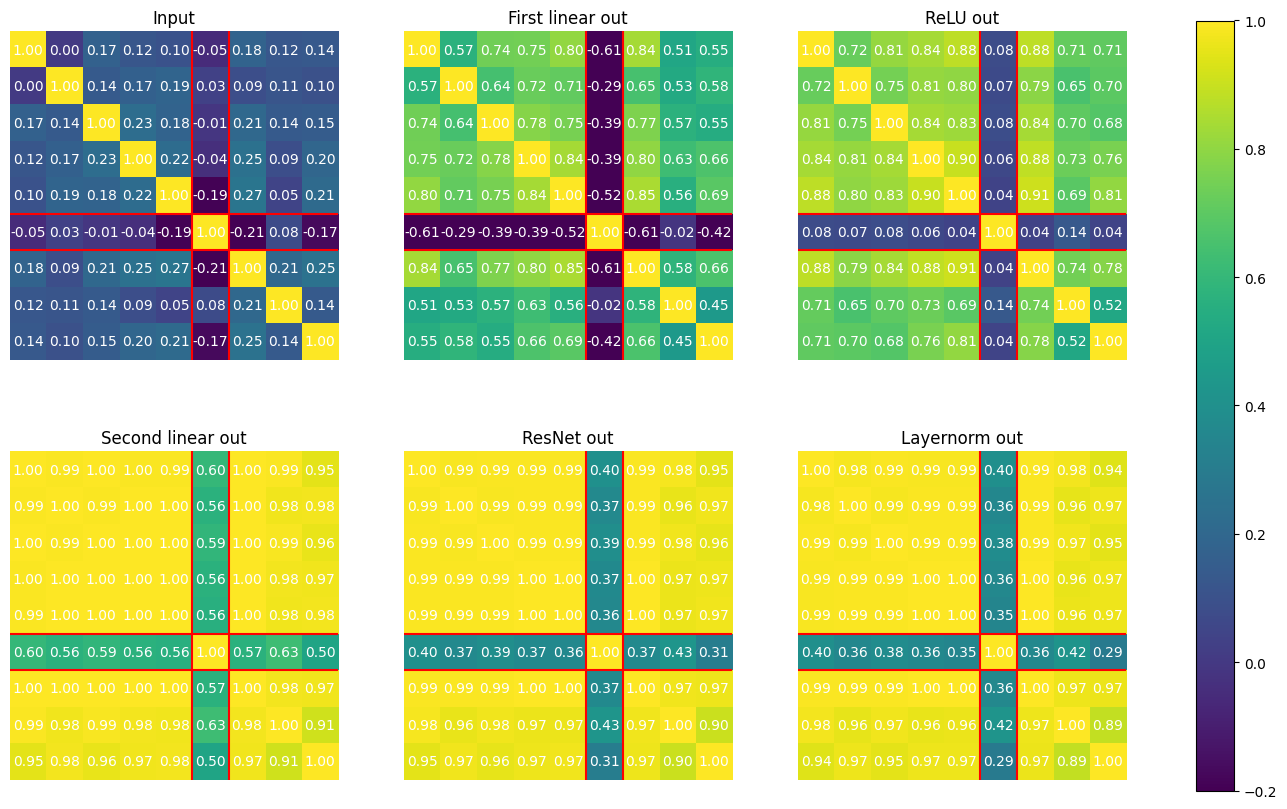}
  \caption{Cosine Similarity Analysis for the output in different first-layer FNN stages with input $X^{(1)} + X^{\mathrm{pr}(1)}$. Each subfigure represents in turn: the input vector, the output of the first linear layer, the output after the activation function, the output of the second linear layer, the output after residual connection, and the output after layer normalization.}
  \label{fig:cosine_similarity_X1_Xpr1}
\end{figure}

\begin{figure}[h]
  \centering
\includegraphics[width=0.6\textwidth]{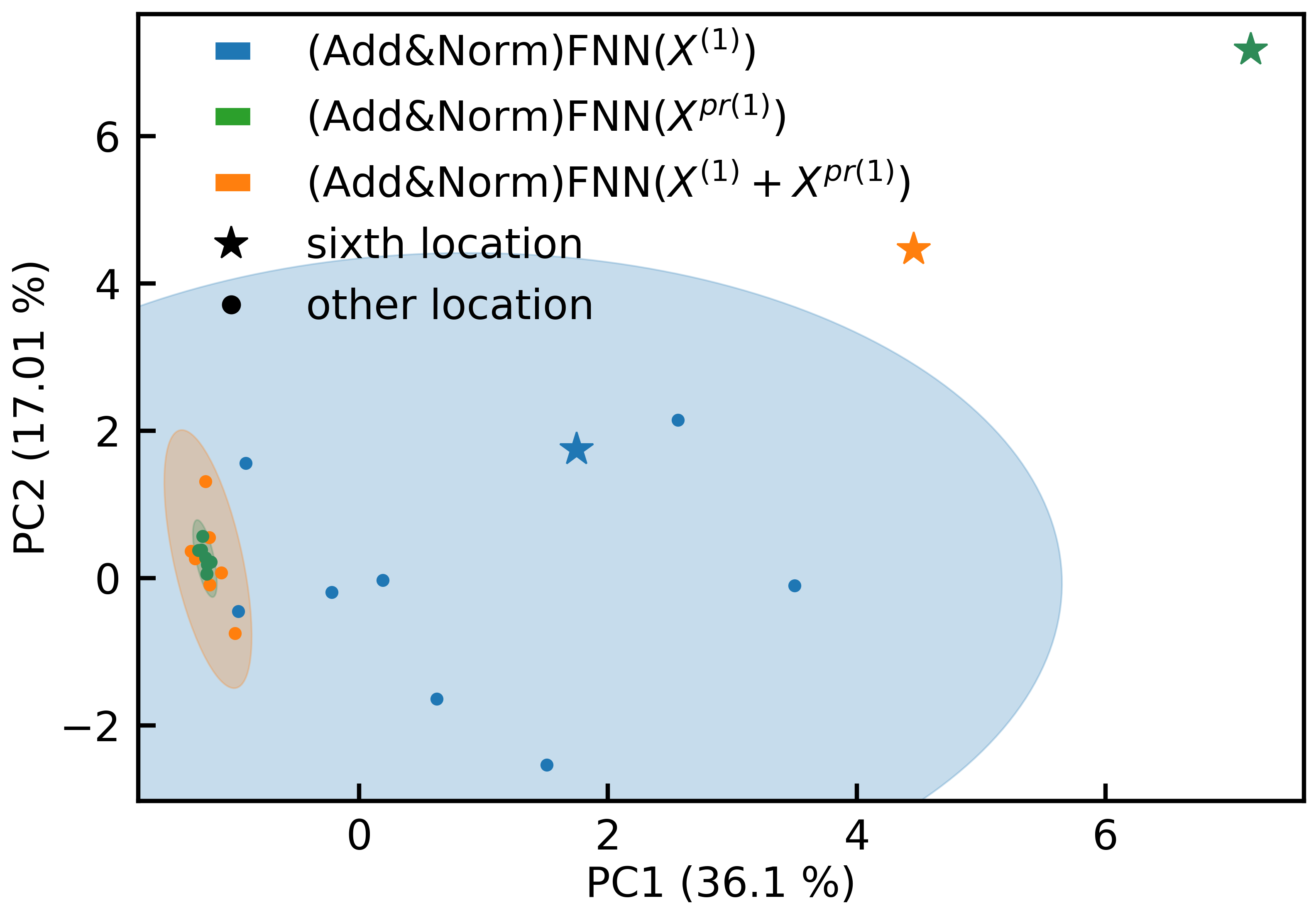}
  \caption{PCA Analysis of three kinds of FNN outputs, i.e., $\text{(Add\& Norm)FNN}(X^{(1)})$, $\text{(Add\& Norm)FNN}(X^{\text{pr(1)}})$ and $\text{(Add\& Norm)FNN}(X^{(1)}+X^{\text{pr(1)}})$. ``(Add\& Norm)'' indicates that output undergoes residual connection and layer normalization. Asterisks represent quantities corresponding to the sixth position digit, while circles represent quantities at other positions. The blue asterisk and circles are almost coincident, indicating minimal distributional differences. The orange asterisk, influenced by the addition of $X^{\mathrm{pr}(1)}$, causes the circles to become relatively concentrated (closer to the green circles), while the asterisk moves farther away (closer to the green asterisk).}
  \label{fig:pca_analysis}
\end{figure}

As shown in Fig.~\ref{fig:cosine_similarity_Xpr1}, the FNN tends to align the vectors (excluding the sixth position) and orthogonalize the vector at the sixth position. And the parallelism of these vectors is also reflected in $\mathrm{FNN}(X^{(1)}+X^{\mathrm{pr}(1)})$ (see Fig.~\ref{fig:cosine_similarity_X1_Xpr1}). This phenomenon can also be further characterized using PCA. As shown in Fig.~\ref{fig:pca_analysis}, asterisks represent quantities corresponding to the sixth position digit, and circles represent quantities at other positions. The close alignment of the blue asterisk and circles suggests minimal distributional differences. Conversely, the orange asterisk, influenced by the addition of $X^{\mathrm{pr}(1)}$, causes the circles to become relatively concentrated (closer to the green circles), while the asterisk moves farther away (closer to the green asterisk).

The conversion of vector directions through the FNN layer is mainly to facilitate the construction of the second layer attention matrix. In the next subsection, we delve deeper into the generation of the second-layer attention matrix.

% The FNN layer induces a parallel configuration for vectors excluding the sixth position and imparts a distinctive orientation to the vector at the sixth position. This transformation is primarily undertaken to facilitate the construction of the attention matrix in the second layer. Subsequently, we delve into the generation of the attention matrix in the second layer.

\subsection{Analysis of attention matrix $\mathrm{Attn}^{(2)}$}

% We begin by presenting the attention mechanism without undergoing masking and softmax (left figure):

\begin{figure}[h]
  \centering
  \subfloat[]{\includegraphics[width=0.5\textwidth]{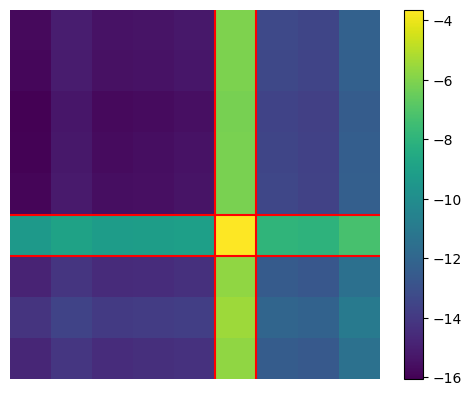}}
  \subfloat[]{\includegraphics[width=0.5\textwidth]{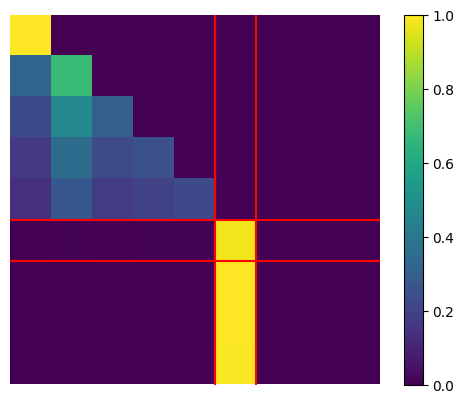}}
  \caption{Analysis of the second-layer attention mechanism before and after masking and softmax. (a) The mechanism before applying masking and softmax. (b) The mechanism with broadcast functionality after masking and softmax.}
  \label{fig:attention_analysis}
\end{figure}

Fig.~\ref{fig:attention_analysis} shows the $\mathrm{Attn}^{(2)}$ matrix before and after mask and softmax operations. We only need to figure out the cause of the attention matrix before these two operations.

\begin{figure}[h]
  \centering
  \includegraphics[width=0.5\textwidth]{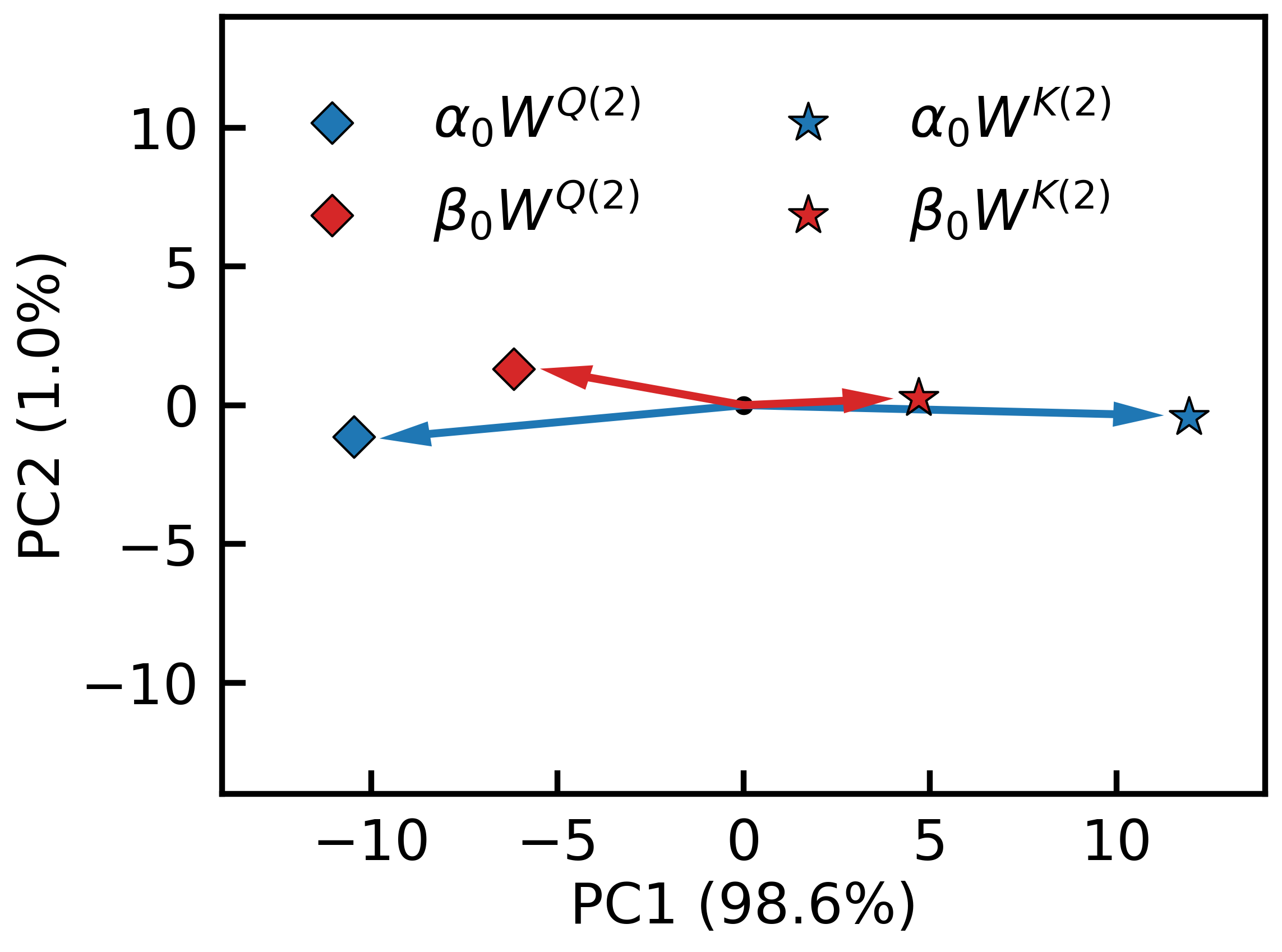}
  \caption{The PCA of the first and sixth row vector of $Q$ and $K$ in the second layer. 
  % $\alpha_0W^{Q(2)}$ represented by the first row vector of $Q$. $\beta_0W^{Q(2)}$ is represented by the sixth row vector of $Q$. $\alpha_0W^{K(2)}$ is represented by the first row vector of $K$. $\beta_0W^{K(2)}$ is represented by the sixth row vector of $K$.
  $\alpha_0W^{Q(2)}$, $\beta_0W^{Q(2)}$, $\alpha_0W^{K(2)}$, and $\beta_0W^{K(2)}$ represent the first raw vector of $Q^{(2)}$, the sixth raw vector of $Q^{(2)}$, the first raw vector of $K^{(2)}$, and the sixth raw vector of $K^{(2)}$, respectively.
  }
  \label{fig:QK2 matrix}
\end{figure}

We mentioned earlier that after FNN, the row vector of $X^{(2)}$ has only 2 directions. We denote the vector of the position outside the key item as $\alpha_0$, and the vector of the position of the key item as $\beta_0$. Therefore, as shown in Fig.~\ref{fig:QK2 matrix}, the row vectors of the matrix $Q^{(2)}$ and $K^{(2)}$ in the second layer also have approximately only two directions. And the two directions of matrix $Q^{(2)}$ and $K^{(2)}$ have a roughly opposite relationship. Therefore, matrix $Q^{(2)}$ and $K^{(2)}$ can be approximately expressed as:

$$Q^{(2)}=(\alpha^T,\ldots,\alpha^T,\beta^T,\alpha^T,\ldots,\alpha^T)^T, K^{(2)}=(-\alpha^T,\ldots,-\alpha^T,-\beta^T,-\alpha^T,\ldots,-\alpha^T)^T.$$

Therefore, we have:

    % \begin{equation}
    %     Q^{(2)}K^{(2)T} = 
    %     \left(
    %     \begin{array}{lllllll}
    %         -\alpha\alpha^T & \ldots & -\alpha\alpha^T & -\alpha\beta^T & -\alpha\alpha^T & \ldots & -\alpha\alpha^T\\
    %         \vdots          & \ddots & \vdots          & -\alpha\beta^T & \vdots          & \ddots & \vdots         \\
    %         -\alpha\alpha^T & \ldots & -\alpha\alpha^T & -\alpha\beta^T & -\alpha\alpha^T & \ldots & -\alpha\alpha^T\\
    %         -\beta\alpha^T  & \ldots & -\beta\alpha^T  & -\beta\beta^T  & -\beta\alpha^T  & \ldots & -\beta\alpha^T \\
    %         -\alpha\alpha^T & \ldots & -\alpha\alpha^T & -\alpha\beta^T & -\alpha\alpha^T & \ldots & -\alpha\alpha^T\\
    %         \vdots          & \ddots & \vdots          & -\alpha\beta^T & \vdots          & \ddots & \vdots         \\
    %         -\alpha\alpha^T & \ldots & -\alpha\alpha^T & -\alpha\beta^T & -\alpha\alpha^T & \ldots & -\alpha\alpha^T\\
    %     \end{array}
    %     \right)
    % \end{equation}

    \begin{equation}
        Q^{(2)}K^{(2)T} = -
        \left(
        \begin{array}{lllllll}
            \alpha\alpha^T & \ldots & \alpha\alpha^T & \alpha\beta^T & \alpha\alpha^T & \ldots & \alpha\alpha^T\\
            \vdots         & \ddots & \vdots         & \alpha\beta^T & \vdots         & \ddots & \vdots        \\
            \alpha\alpha^T & \ldots & \alpha\alpha^T & \alpha\beta^T & \alpha\alpha^T & \ldots & \alpha\alpha^T\\
            \beta\alpha^T  & \ldots & \beta\alpha^T  & \beta\beta^T  & \beta\alpha^T  & \ldots & \beta\alpha^T \\
            \alpha\alpha^T & \ldots & \alpha\alpha^T & \alpha\beta^T & \alpha\alpha^T & \ldots & \alpha\alpha^T\\
            \vdots         & \ddots & \vdots         & \alpha\beta^T & \vdots         & \ddots & \vdots        \\
            \alpha\alpha^T & \ldots & \alpha\alpha^T & \alpha\beta^T & \alpha\alpha^T & \ldots & \alpha\alpha^T\\
        \end{array}
        \right).
    \end{equation}

    To realize the matrix in Fig.~\ref{fig:attention_analysis}(a), just need $\alpha\beta^T>0$ and $\norm{\alpha}_2>\norm{\beta}_2$.

\subsection{Interpretation of attention mechanism results}

% Concluding the analysis, the orange color represents the values of Q, while the blue color represents the values of K.

% Finally, let's investigate if there is any special significance in the direction of $\bar{\text{FNN}(X^{\mathrm{pr}(1)})}$. Given that $X^{\mathrm{ao}(2)}=\mathrm{LayerNorm}(X^{(2)}+X^{\mathrm{pr}(2)})$ and experimentally $X^{\mathrm{ao}(2)}_9$ provides the desired output, and $X^{(2)}_9$ exhibits similarity to the direction of $\bar{\text{FNN}(X^{\mathrm{pr}(1)})}$, we hypothesize that $\bar{\text{FNN}(X^{\mathrm{pr}(1)})}$ likely has an orthogonal relationship with the projection layer. This implies that the addition of $X^{(2)}$ can be canceled out by the projection layer.

% Let's examine the output of $\bar{\text{FNN}(X^{\mathrm{pr}(1)})}$ directly passing through the projection layer. Fig.~\ref{fig:projection_output}(a) shows only the projection output for values ranging from 20 to 100, while Fig.~\ref{fig:projection_output}(b) displays the projection output for all values. The result indicate that the values from 20 to 100 are nearly identical, and the remaining values are also quite similar. This is primarily because there is no need to fit values outside the 20-100 range, leading to a tendency to assign a larger negative value to prevent the output of these numbers. Moreover, the outputs for values from 20 to 100 are consistent, and due to the mean subtraction operation of softmax, $\bar{\text{FNN}(X^{\mathrm{pr}(1)})}$ has minimal impact on the output after passing through the projection layer.

\begin{figure}[h]
  \centering
  \includegraphics[width=0.8\textwidth]{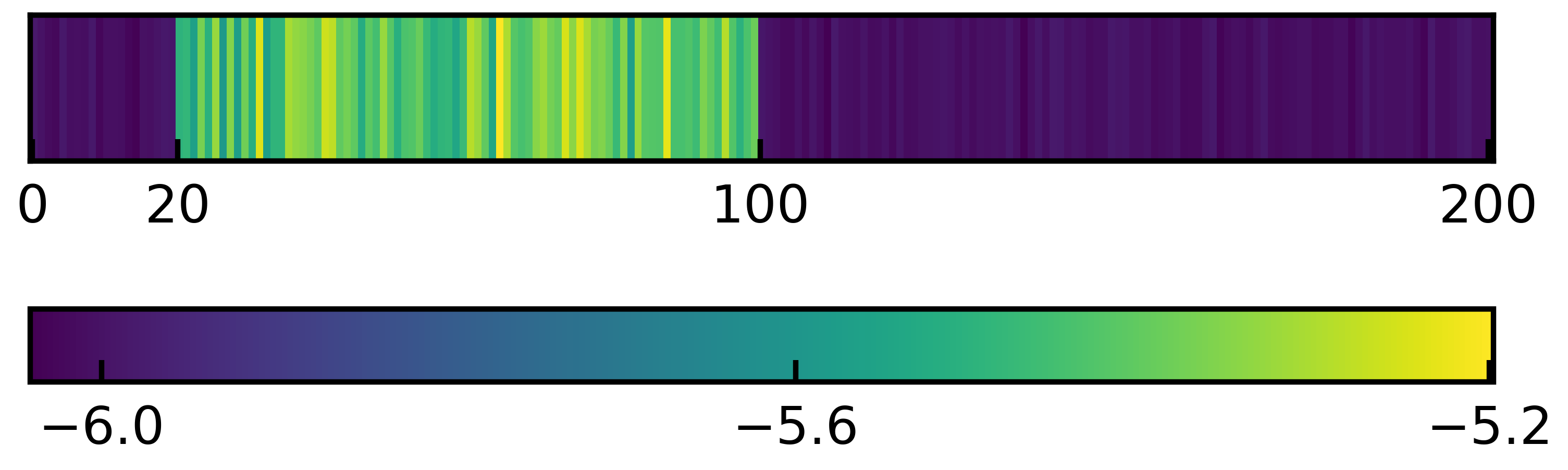}
  \caption{Projection output of $\alpha_0$ passing through the projection layer. $\alpha_0$ makes the model tend to output numbers in the range 20
  -100. But it has little impact on the specific output value.}
  \label{fig:projection_output}
\end{figure}

After the second layer of the decoder block, the last row vector of hidden state $X^{(3)}$ is 

$$X^{(3)}_9=\mathrm{Layernorm}(\alpha_0+\beta_0W^{V(2)}),$$
where $\alpha_0$ is the  vector of positions other than the position of the key item and $\beta_0$ is the vector of the key item position. $\alpha_0$ has no information about the coupling of anchor and key item, so its response to the projection layer should not have an impact on the final result. As can be seen from Fig.~\ref{fig:projection_output}, $\alpha_0W^{\mathrm{proj}}+b^{\mathrm{proj}}$ makes the probability of outputting 20-100 greater than other tokens, but the probability of outputting 20-100 is approximately the same.

% \section{Loss and accuracy}

\section{Attention maps of Llama2-6B}\label{sec: Attention maps of Llama2-6B}

\begin{figure}
    \centering
    \includegraphics[width=\linewidth]{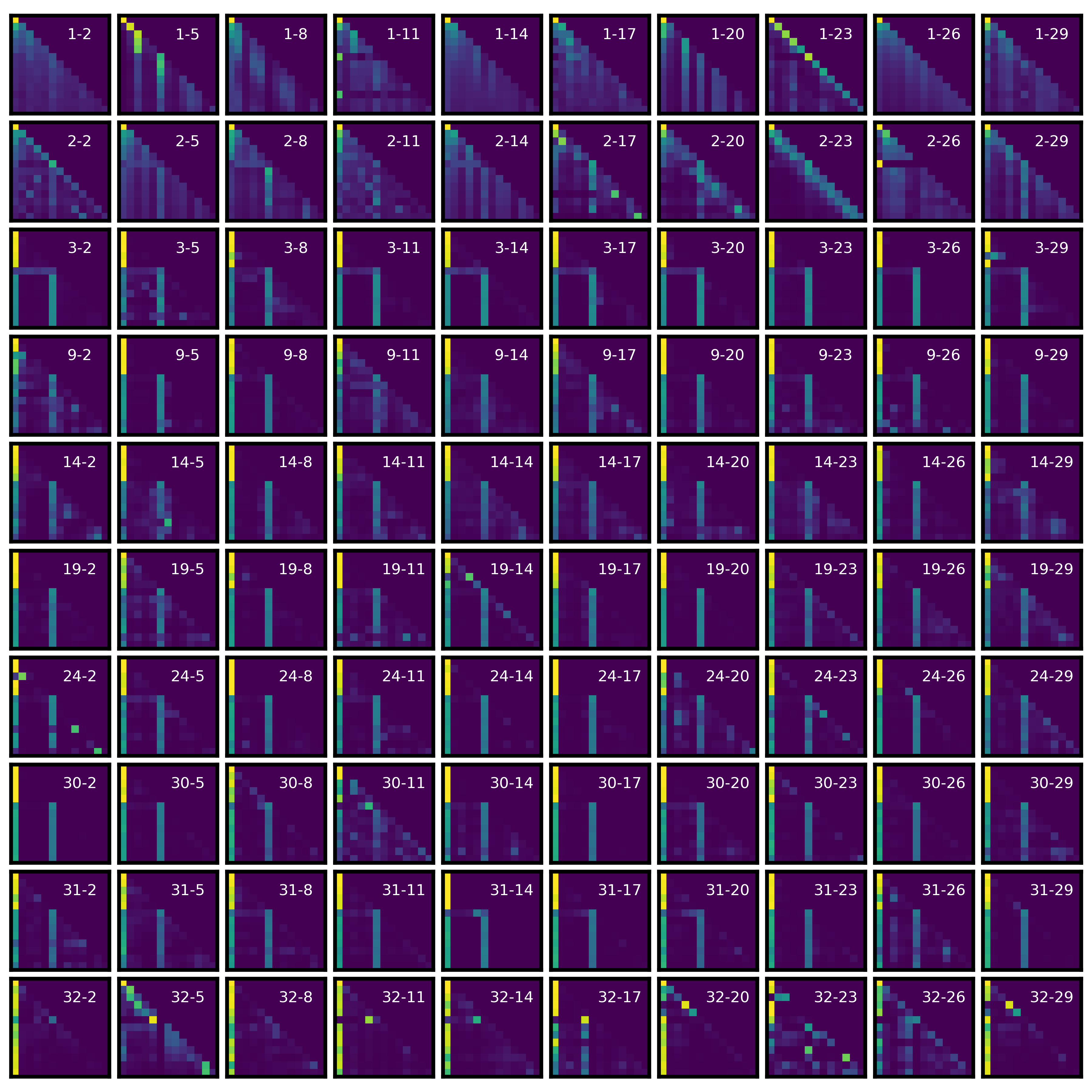}

    \caption{Attention maps of Llama2-6B. From top to bottom, from left to right are the attention matrices of different layers and heads. In the first few layers, attention mostly shows the pattern of shift. In the later layers, attention mostly shows the pattern of the broadcast.}
    \label{fig:Llama_attention}
\end{figure}

\end{document}